\documentclass[acmtocl]{acmtrans2m}

\usepackage{amsmath}
\usepackage{amssymb}
\usepackage{epsfig}
\usepackage{latexsym}
\usepackage{graphicx}
\usepackage{url}

\newtheorem{textoftheorem}{Theorem}[section]
\newtheorem{textofdefinition}{Definition}
\newtheorem{textofexample}{Example}
\newtheorem{textofcorollary}{Corollary}
\newtheorem{textoflemma}{Lemma}

\newcommand\ignore[1]{}
\newcommand{\imp}{\supset}
\let \yeq = \simeq
\def\nyeq{\mathop{\not\simeq}}
\def\der#1{\mathop{\vdash}_{#1}}
\newcommand{\irule}[2]{\begin{array}{c}{#1}\\\hline{#2}\end{array}}
\newcommand{\srule}[2]{\begin{array}{c}{#1}\\\hline\hline{#2}\end{array}}
\def\sub{\stackrel{\scriptscriptstyle \bullet}{}\!\!\!>}
\def\subeq{\stackrel{\scriptscriptstyle \bullet}{}\!\!\!\geq}
\def\cfredd{\mathop{\rightarrow}^*}
\def\cfress{\mathop{\leftarrow}^*}
\def\jjoin{\cfredd\circ\cfress}
\newcommand{\st}{\mathop:\,}
\newcommand{\Th}[1]{\mathit{Th}\,#1}
\def\IN{\hbox{\rm I\hbox{\hskip -2pt N}}}
\newcommand{\unfold}{\!\!\downarrow}
\def\sunif{\mathop{=}^?}
\newcommand{\eprover}{\mbox{E}}
\newcommand{\harvey}{\mbox{haRVey}}
\newcommand{\SP} {\ensuremath{\cal SP}}

\DeclareMathOperator{\STORECOMM}{{\tt STORECOMM}}
\DeclareMathOperator{\STORECOMMINVALID}{{\tt STORECOMM\_INVALID}}
\DeclareMathOperator{\SWAP}{{\tt SWAP}}
\DeclareMathOperator{\SWAPINVALID}{{\tt SWAP\_INVALID}}
\DeclareMathOperator{\STOREINV}{{\tt STOREINV}}
\DeclareMathOperator{\STOREINVINVALID}{{\tt STOREINV\_INVALID}}
\DeclareMathOperator{\multiswap}{multiswap}
\DeclareMathOperator{\IOS}{{\tt IOS}}

\DeclareMathOperator{\QUEUE}{{\tt QUEUE}}
\DeclareMathOperator{\CIRCULARQUEUE}{{\tt CIRCULAR\_QUEUE}}

\DeclareMathOperator{\select}{select}
\DeclareMathOperator{\store}{store}
\DeclareMathOperator{\rselect}{rselect}
\DeclareMathOperator{\rstore}{rstore}
\DeclareMathOperator{\s}{s}
\DeclareMathOperator{\p}{p}
\DeclareMathOperator{\reset}{reset}
\DeclareMathOperator{\enqueue}{enqueue}
\DeclareMathOperator{\dequeue}{dequeue}
\DeclareMathOperator{\first}{first}
\DeclareMathOperator{\last}{last}
\DeclareMathOperator{\cons}{cons}
\DeclareMathOperator{\car}{car}
\DeclareMathOperator{\cdr}{cdr}
\DeclareMathOperator{\nil}{nil}
\DeclareMathOperator{\atom}{atom}

\title{New results on rewrite-based satisfiability procedures}

\author{
ALESSANDRO ARMANDO,
DIST, Universit\`a degli Studi di Genova
\and
MARIA PAOLA BONACINA,
Dip. di Informatica, Universit\`a degli Studi di Verona
\and
SILVIO RANISE,
LORIA \& INRIA-Lorraine
\and
STEPHAN SCHULZ,
Dip. di Informatica, Universit\`a degli Studi di Verona
}

\begin{abstract}
Program analysis and verification require decision procedures to reason on
theories of data structures.
Many problems can be reduced to the {\em satisfiability}
of sets of {\em ground literals} in theory $T$.
If a sound and complete inference system for first-order logic
is guaranteed to {\em terminate} on {\em $T$-satisfiability problems},
any theorem-proving strategy with that system
and a fair search plan is a {\em $T$-satisfiability procedure}.
We prove termination of a rewrite-based first-order engine
on the theories of {\em records},
{\em integer offsets}, {\em integer offsets modulo} and {\em lists}.
We give a {\em modularity theorem} stating sufficient conditions for
termination on a {\em combinations of theories}, given termination on each.
The above theories, as well as others, satisfy these conditions.
We introduce several sets of benchmarks on these theories and their combinations,
including both {\em parametric} synthetic benchmarks to test {\em scalability},
and real-world problems to test performances on huge sets of literals.
We compare the rewrite-based theorem prover \eprover{}
with the validity checkers CVC and CVC~Lite.
Contrary to the folklore that a general-purpose prover
cannot compete with reasoners with built-in theories,
the experiments are overall favorable to the theorem prover,
showing that not only the rewriting approach is elegant and conceptually simple,
but has important practical implications.
\end{abstract}




\begin{document}

\begin{bottomstuff}
Research supported in part by MIUR grant no.~2003-097383.
First author's address:
Viale Causa 13, 16145 Genova, Italy,
{\tt armando@dist.unige.it}.
Second and fourth authors' address:
Strada Le Grazie 15, 37134 Verona, Italy,
{\tt mariapaola.bonacina@univr.it},
{\tt schulz@eprover.org}.
Third author's address:
615 Rue Du Jardin Botanique, B.P. 101,
54600 Villers-l\`es-Nancy, France,
{\tt silvio.ranise@loria.fr}.
\end{bottomstuff}

\maketitle

\section{Introduction}

Decision procedures for satisfiability in theories of
data types, such as {\em arrays}, {\em lists} and {\em records},
are at the core of many state-of-the-art verification tools
(e.g., PVS \cite{PVS}, ACL2 \cite{ACL2}, Simplify \cite{simplify},
CVC \cite{BarrettEtAl1}, ICS \cite{ICS04},
CVC~Lite \cite{CVCLite}, Zap \cite{Musuvathi}, MathSAT \cite{mathsat1},
Yices \cite{DM06} and Barcelogic \cite{NieuOli06}).
The design, proof of correctness, and implementation of
satisfiability procedures\footnote{In the literature on decision
procedures, a ``satisfiability procedure'' is a decision procedure
for ``satisfiability problems'' that are sets of ground literals.}
present several issues,
that have brought them to the forefront of research in automated
reasoning applied to verification.

First, most verification problems involve more than one theory,
so that one needs procedures for {\em combinations of theories},
such as those pioneered by \cite{NO79} and \cite{ShostakCC2}.
Combination is complicated: for instance,
understanding, formalizing and proving correct Shostak's method
required much work (e.g., \cite{CyrLinSha,ShaonSho,BDS02a,Ga02,RaRiTran:ICTAC04}).
The need for combination of theories means that decision procedures
ought to be easy to modify, extend, integrate into,
or at least interface, with other decision procedures or more general systems.
Second, satisfiability procedures need to be proved correct and complete:
a key part is to show that whenever the algorithm reports {\em satisfiable},
a model of the input does exist.
Model-construction arguments for concrete procedures are specialized for those,
so that each new procedure requires a new proof.
Frameworks that offer a higher level of abstraction (e.g., \cite{BacTiwVig,Ga02})
often focus on combining the quantifier-free
theory of equality\footnote{Also known as
{\em EUF} for Equality with Un-interpreted Function symbols.
In the literature on decision procedures,
most authors use ``interpreted'' and ``un-interpreted''
to distinguish between those symbols whose interpretation is restricted
to the models of a given theory and those whose interpretation is unrestricted.
In the literature on rewriting,
it is more traditional to use ``definite'' in place of ``interpreted''
and ``free'' in place of ``un-interpreted'', as done in \cite{Ga02}.},
with at most one additional theory,
while problems from applications, and existing systems, combine many.
Third, although systems begin to offer some support for adding theories,
developers usually have to write a large amount of new code for each procedure,
with little software reuse and high risk of errors.

If one could use first-order theorem-proving strategies,
combination would become conceptually much simpler,
because combining theories would amount to giving as input to the strategy
the union of the presentations of the theories.
No {\em ad hoc} correctness and completeness proofs would be needed,
because a sound and complete theorem-proving strategy
is a {\em semi-decision procedure} for unsatisfiability.
Existing first-order provers,
that embody the results of years of research
on data structures and algorithms for deduction,
could be applied, or at least their code could be reused,
offering a higher degree of assurance about soundness and completeness
of the procedure.
Furthermore, theorem-proving strategies support {\em proof generation}
and {\em model generation},
that are two more desiderata of satisfiability procedures
(e.g., \cite{Necula,flea,Musuvathi}),
in a theory-independent way.
Indeed, if the input is unsatisfiable,
the strategy generates a proof with no additional effort.
If it is satisfiable,
the strategy generates a saturated set, that, if finite,
may form a basis for model generation \cite{CLPbook}.

The crux is {\em termination}: in order to have a decision procedure,
one needs to prove that a complete theorem-proving strategy is bound
to terminate on satisfiability problems in the theories of interest.
Results of this nature were obtained in \cite{ArRaRu2}:
a refutationally complete {\em rewrite-based inference system},
named $\SP$ (from superposition),
was shown to generate finitely many clauses on satisfiability problems in
the theories of {\em non-empty lists},
{\em arrays with or without extensionality},
{\em encryption},
{\em finite sets with extensionality},
{\em homomorphism},
and the combination of lists and arrays.
This work was extended in \cite{LM02:lics},
by using a meta-saturation procedure
to add complexity characterizations\footnote{Meta-saturation as in \cite{LM02:lics}
was later corrected in \cite{LM02:licsFix}.}.
Since the inference system $\SP$ reduces to ground completion
on a set of ground equalities and inequalities,
it terminates and represents a decision procedure also for
the quantifier-free theory of equality\footnote{That ground completion
can be used to compute congruence closure has been known since \cite{L75:ut}.}.

These termination results suggest that, at least in principle,
rewrite-based theorem provers might be used
``off the shelf'' as validity checkers.
The common expectation, however, is that validity checkers with built-in
theories will be much faster than theorem provers that take
theory presentations as input.
In this paper,
we bring evidence that using rewrite-based theorem provers can be a pratical option.
Our contributions include:
\begin{itemize}
\item New termination results,
showing that $\SP$ generates finitely many clauses from satisfiability problems
in the theories of more data structures,
{\em records with or without extensionality}
and {\em possibly empty lists},
and in two fragments of integer arithmetic, the theories of
{\em integer offsets} and {\em integer offsets modulo};
\item A general {\em modularity theorem},
that states sufficient conditions for $\SP$ to terminate on satisfiability problems
in a union of theories, if it terminates on the satisfiability problems of each theory
taken separately;
\item A report on experiments where six sets of parametric synthetic benchmarks
were given to the rewrite-based theorem prover \eprover{} \cite{JE,E081},
the \emph{Cooperating Validity Checker} CVC \cite{CVC}
and its successor CVC~Lite \cite{CVCLite}:
contrary to expectation,
the general first-order prover with the theory presentations in input
was, overall, comparable with the validity checkers with built-in
theories, and in some cases even outperformed them.
\end{itemize}

Among the termination results,
the one for the theory of {\em integer offsets} is perhaps the most surprising,
because the axiomatization is infinite.
All the theories considered in this paper (i.e.,
{\em records with or without extensionality},
{\em lists},
{\em integer offsets},
{\em integer offsets modulo},
{\em arrays with or without extensionality}
and the quantifier-free theory of equality)
satisfy the hypotheses of the modularity theorem,
so that a fair $\SP$-strategy is a satisfiability procedure
for {\em any} of their combinations.
This shows the flexibility of the rewrite-based approach.

For the experiments,
we chose a state-of-the-art theorem prover that implements $\SP$,
and two systems that combine decision procedures with built-in theories
{\em \`a la} Nelson-Oppen.
At the time of these experiments,
CVC and CVC~Lite were the only state-of-the-art tools implementing a correct
and complete procedure for arrays with extensionality\footnote{Neither
Simplify nor ICS are complete in this regard: cf. Section 5 in
\cite{simplify} and \cite{Ruess}, respectively.},
namely that of \cite{StuBaDiLe}.
We worked with {\em parametric} synthetic benchmarks,
because they allow one to assess the \textit{scalability} of systems
by a sort of \textit{experimental asymptotic analysis}.
Three sets of benchmarks involve the theory of {\em arrays with extensionality},
one combines the theory of {\em arrays} with that of {\em integer offsets},
one is about {\em queues},
and one is about {\em circular queues}.
In order to complete our appraisal,
we tested \eprover{} on sets of literals extracted from real-world
problems of the UCLID suite \cite{SS:CAV-2004},
and found it solves them extremely fast.
The selection of problems emphasizes {\em the combination of theories},
because it is relevant in practice.
The synthetic benchmarks on {\em queues} feature the theories of
{\em records}, {\em arrays} and {\em integer offsets},
because a queue can be modelled as a record,
that unites a partially filled array
with two indices that represent {\em head} and {\em tail}.
Similarly,
the benchmarks on {\em circular queues}
involve the theories of {\em records}, {\em arrays} and {\em integer offsets modulo},
because a circular queue of length $k$ is a queue
whose indices take integer values modulo $k$.
The UCLID problems combine the {\em theory of integer offsets} and
the {\em quantifier-free theory of equality}.

\subsection{Previous work}

Most termination results for theorem-proving
methods are based on identifying generic syntactic constraints
that the input must satisfy to induce termination
(e.g., \cite{FLHT01,CLPbook} for two overviews).
Our results are different,
because they apply to specific theories,
and in this respect they can be considered of a more semantic nature.
There are a few other recent works that experiment with the application
of first-order theorem provers to decidable theories of data structures.
A proof of correctness of a basic Unix-style file system implementation
was obtained in \cite{AZKR}, by having a proof checker invoke the SPASS
\cite{WABCEKTT} and Vampire \cite{RV:AICOM-2002} provers
for non-inductive reasoning on {\em lists} and {\em arrays},
on the basis of their first-order presentations.
The \harvey{} system \cite{haRVey-sefm03} is a verification
tool based on the rewriting approach that we propound in this paper.
It integrates the \eprover{} prover with a SAT solver,
based on ordered binary decision diagrams,
to implement decision procedures for a few theories.
Experiments with \harvey{} offered additional evidence
of the effectiveness of the rewriting approach \cite{RaDe03}.

The collection of theories considered here is different
from that treated in \cite{ArRaRu2}.
Lists {\em \`a la Shostak} (with $\cons$, $\car$, $\cdr$ and three axioms)
and lists {\em \`a la Nelson-Oppen} (with $\cons$, $\car$, $\cdr$, $\atom$ and four axioms)
were covered in \cite{ArRaRu2}.
Both axiomatize {\em non-empty} lists,
since there is no symbol such as {\em nil} to represent the empty list.
Here we consider a different presentation,
with $\cons$, $\car$, $\cdr$, $\nil$ and six axioms,
that allows for empty lists.
In an approach where the axioms are given in input to a theorem prover,
a different presentation represents a different problem,
because termination on satisfiability problems including a presentation
does not imply termination on satisfiability problems including another presentation.
To wit, the finite saturated sets generated by $\SP$ are different
(cf. Lemma~\ref{lemma:caseAnalysis:list} in this article
with Lemmata 4.1 and 5.1 in \cite{ArRaRu2}).
Application of a rewrite-based engine to
the theories of records, integer offsets and integer offsets modulo
is studied here for the first time.
Although the presentation of the theory of records resembles
that of arrays, the treatment of extensionality is different
and the generated saturated sets are very different
(cf. Lemma~\ref{lemma:caseAnalysis:record}
and~\ref{lemma:caseAnalysis:array} in this article).
The only overlap with \cite{ArRaRu2} is represented by the theory of arrays,
for which we redo only the case analysis of generated clauses,
because that reported in \cite{ArRaRu2} is incomplete
(cf. Lemma~\ref{lemma:caseAnalysis:array} in this article
with Lemma 7.2 in \cite{ArRaRu2}).
A short version and an extended abstract of this article
were presented in \cite{rewsatproc} and \cite{engines}, respectively.
Very preliminary experiments with a few of the synthetic
benchmarks were reported in \cite{array}.

\section{Background}

We employ the basic notions from logic usually assumed in theorem proving.
For notation,
the symbol $\yeq$ denotes equality;\footnote{The notation $\yeq$ is standard
for unordered pair, so that $l\yeq r$ stands for $l\yeq r$ or $r\yeq l$.}
$\bowtie$ stands for either $\yeq$ or $\nyeq$;
$=$ denotes identity;
$l, r, u, t$ are terms;
$v, w, x, y, z$ are variables;
other lower-case Latin letters are constant or function symbols based on arity;
$L$ is a literal;
$C$ and $D$ denote clauses, that is, multisets of literals interpreted as disjunctions;
$\varphi$ is a formula;
and $\sigma$ is used for substitutions.
More notation will be introduced as needed.

A theory is presented by a set of sentences,
called its \emph{presentation} or \emph{axiomatization}.
Given a presentation ${\cal T}$,
the \emph{theory} $\Th {\cal T}$ is the set of all its logical consequences,
or theorems:
$\Th {\cal T} = \{\varphi\ \vert\ {\cal T}\models\varphi\}$.
Thus, a theory is a deductively-closed presentation.
An \emph{equational theory} is a theory presented
by a set of universally quantified equations.
A \emph{Horn clause} is a clause with at most one positive literal,
and a \emph{definite Horn clause}, or non-negative Horn clause,
is a clause with at most and at least one positive literal.
A \emph{Horn theory} is presented
by a set of non-negative Horn clauses, and
a \emph{Horn equational theory} is Horn theory
where the only predicate is equality.
From a model-theoretic point of view,
the term \emph{theory} refers to the family of {\em models} of ${\cal T}$,
or ${\cal T}$-models.
A model is called \textit{trivial} if its domain has only one element.
It is customary to ascribe to a model the cardinality of its domain,
so that a model is said to be \textit{finite} or \textit{infinite}
if its domain is.

By ${\cal T}$-atom, ${\cal T}$-literal, ${\cal T}$-clause,
${\cal T}$-sentence and ${\cal T}$-formula,
we mean an atom, a literal, a clause, a sentence and a formula, respectively,
on ${\cal T}$'s signature,
omitting the ${\cal T}$ when it is clear from context.
Equality is the only predicate,
so that all ${\cal T}$-atoms are ${\cal T}$-equations.
The problem of {\em ${\cal T}$-satisfiability},
or, equivalently, {\em satisfiability modulo ${\cal T}$},
is the problem of deciding whether a set $S$ of ground ${\cal T}$-literals
is satisfiable in ${\cal T}$, or has a ${\cal T}$-model.
The more general {\em ${\cal T}$-decision problem} consists of deciding
whether a set $S$ of quantifier-free ${\cal T}$-formul{\ae}
is satisfiable in ${\cal T}$.
In principle, the ${\cal T}$-decision problem can be reduced
to the ${\cal T}$-satisfiability problem via reduction of every
quantifier-free ${\cal T}$-formula to disjunctive normal form.
However, this is not pratical in general.
In this paper, we are concerned only with
{\em ${\cal T}$-satisfiability}.\footnote{We discuss existing approaches
and future directions for the ${\cal T}$-decision problem in Section~\ref{disc}.}
${\cal T}$-satisfiability is important,
because many problems reduce to ${\cal T}$-satisfiability:
the {\em word problem}, or the problem of
deciding whether ${\cal T}\models\forall\bar x\ l\yeq r$,
where $l\yeq r$ is a ${\cal T}$-equation,
the {\em uniform word problem}, or the problem of
deciding whether ${\cal T}\models\forall\bar x\ C$,
where $C$ is a Horn ${\cal T}$-clause, and
the {\em clausal validity problem}, or the problem of
deciding whether ${\cal T}\models\forall\bar x\ C$,
where $C$ is a ${\cal T}$-clause,
all reduce, through skolemization,
to deciding the unsatisfiability of a set of ground literals,
since all variables are universally quantified.

The traditional approach to ${\cal T}$-satisfiability
is that of {\em ``little'' engines of proofs} (e.g., \cite{Shankar}),
which consists of building each theory into a de\-di\-ca\-ted inference engine.
Since the theory is built into the engine,
the input of the procedure consists of $S$ only.
The most basic example is that of congruence closure algorithms
for satisfiability of sets of ground equalities and inequalities
(e.g., \cite{ShostakCC,NO80,DST,BacTiwVig})\footnote{Unknown to most,
the conference version of \cite{DST} appeared in \cite{DSS}
with a different set of authors.}.
Theories are built into the congruence closure algorithm
by generating the necessary instances of the axioms
(see \cite{NO80} for non-empty lists)
or by adding pre-processing with respect to the axioms
and suitable case analyses
(see \cite{StuBaDiLe} for arrays with extensionality).
Theories are combined by using the method of \cite{NO79}.
Two properties relevant to this method are \textit{convexity}
and \textit{stable infiniteness}:

\begin{textofdefinition}\label{convex}
A theory $\Th {\cal T}$ is {\em convex},
if for any conjunction $H$ of ${\cal T}$-atoms
and for ${\cal T}$-atoms $P_i$, $1\le i\le n$,
${\cal T}\models H\imp\bigvee_{i=1}^n P_i$
implies that there exists a $j$, $1\le j\le n$,
such that ${\cal T}\models H\imp P_j$.
\end{textofdefinition}

In other words, if $\bigvee_{i=1}^n P_i$ is true in all models of ${\cal T}\cup H$,
there exists a $P_j$ that is true in all models of ${\cal T}\cup H$.
This excludes the situation where all models of ${\cal T}\cup H$ satisfy some $P_j$,
but no $P_j$ is satisfied by all.
Since Horn theories are those theories
whose models are closed under intersection --
a fact due to Alfred Horn \cite[Lemma 7]{Horn} --
it follows that Horn theories, hence equational theories, are convex.
The method of Nelson-Oppen
without case analysis (also known as ``branching'' or ``splitting'')
is complete for combinations where all theories are convex
(e.g., \cite{NO79,BDS02a,Ga02}).
The method of Nelson-Oppen with case analysis is complete
for combinations where all involved theories are stably infinite \cite{TinelliHarandi}:

\begin{textofdefinition}\label{stably:infinite}
A theory $\Th {\cal T}$ is {\em stably infinite},
if for any quantifier-free ${\cal T}$-formula $\varphi$,
$\varphi$ has a ${\cal T}$-model if and only if it has an
infinite ${\cal T}$-model.
\end{textofdefinition}

When combining the quantifier-free theory of equality with only one other theory,
the requirement of stable infiniteness can be dropped \cite{Ga02}.
For first-order logic,
compactness implies that if a set of formul\ae\ has models
with domains of arbitrarily large finite cardinality,
then it has models with infinite domains (e.g., \cite{vanDalen} for a proof).
Thus, for ${\cal T}$ a first-order presentation,
and $\varphi$ a quantifier-free ${\cal T}$-formula,
if $\varphi$ has arbitrarily large finite ${\cal T}$-models,
it has infinite ${\cal T}$-models; or, equivalently,
if it has no infinite ${\cal T}$-model,
there is a finite bound on the size of its ${\cal T}$-models.\footnote{A
proof of this consequence of compactness in the context of decision procedures
appears in \cite{GanRuSha}, where $\varphi$ is assumed to have been reduced to disjunctive normal form, so that the proof is done for a set of literals.}
Using this property, one proves (cf. Theorem 4 in \cite{BDS02a}):

\begin{textoftheorem}(Barrett, Dill and Stump 2002)\label{convex:stablyInfinite}
Every convex first-order theory with no trivial models is stably-infinite.
\end{textoftheorem}

\noindent
Thus, stable infiniteness is a weaker property characterizing the theories
that can be combined according to the Nelson-Oppen scheme.

If a decision procedure with a built-in theory is a {\em little engine of proof},
an inference system for full first-order logic with equality can be considered a
{\em ``big'' engine of proof} (e.g., \cite{Stickel}).
One such engine is the {\em rewrite-based inference system $\SP$},
whose {\em expansion} and {\em contraction} inference rules are listed in Figures~\ref{tab:sp-er} and~\ref{tab:sp-cr}, respectively.
Expansion rules {\em add} what is below the inference line to the clause set that
contains what is above the inference line.
Contraction rules {\em remove} what is above the double inference line and add
what is below the double inference line.
Combinations of these inference rules or variants thereof
form the core of most theorem provers for first-order logic with equality,
such as Otter \cite{otter3}, SPASS \cite{WABCEKTT}, Vampire \cite{RV:AICOM-2002},
and \eprover{} \cite{JE}, to name a few.
Formulations with different terminologies
(e.g., left and right superposition in place of paramodulation
and superposition) appear in the vast literature on the subject
(e.g., \cite{Plaistedeqsurvey,taxonomy,NRHB,DPHB} for surveys
where more references can be found).

A fundamental assumption of rewrite-based inference systems
is that the universe of terms, hence those of literals and clauses,
is ordered by a {\em well-founded ordering}.
$\SP$ features a {\em complete simplification ordering (CSO)} $\succ$ on terms,
extended to literals and clauses by multiset extension as usual.
A simplification ordering is {\em stable}
($l\succ r$ implies $l\sigma\succ r\sigma$ for all substitutions $\sigma$),
{\em monotonic}
($l\succ r$ implies $t[l]\succ t[r]$ for all $t$, where
the notation $t[l]$ represents a term where $l$ appears as subterm in context $t$),
and has the {\em subterm property}
(i.e., it contains the {\em subterm ordering} $\rhd$: $l\rhd r$ implies $l\succ r$).
An ordering with these properties is well-founded.
A CSO is also total on ground terms.
The most commonly used CSO's are instances of the {\em recursive path ordering (RPO)}
and the {\em Knuth-Bendix ordering (KBO)}.
An RPO is based on a {\em precedence}
(i.e., a partial ordering on the signature)
and the attribution of a {\em status} to each symbol in the signature
(either lexicographic or multiset status).
If all symbols have lexicographic status,
the ordering is called {\em lexicographic (recursive) path ordering (LPO)}.
A KBO is based on a {\em precedence}
and the attribution of a {\em weight} to each symbol.
All instances of RPO and KBO are simplification orderings.
All instances of KBO and LPO based on a total precedence are CSO's.
Definitions, results and references on
orderings for rewrite-based inference can be found in \cite{DPHB}.

\begin{figure}[tbh]
\begin{center}
\begin{tabular}{cc}
{\small \textsl{Superposition}} &
$\irule{C\vee l[u^\prime]\yeq r ~~~ D\vee u\yeq t}
{(C\vee D\vee  l[t]\yeq r)\sigma}
\begin{array}{cccc}
(i), & (ii), & (iii), & (iv)
\end{array}$
\\\hline
{\small \textsl{Paramodulation}} &
$\irule{C\vee l[u^\prime]\nyeq r ~~~ D\vee u\yeq t}
{(C\vee D\vee  l[t]\nyeq r)\sigma}
\begin{array}{cccc}
(i), & (ii), & (iii), & (iv)
\end{array}$
\\\hline
{\small \textsl{Reflection}} &
$\irule{C\vee u^\prime \nyeq u}{C\sigma}
\begin{array}{c}
\forall L\in C : (u^\prime \yeq u)\sigma \not\prec L\sigma 
\end{array}$
\\\hline
{\small \textsl{Equational Factoring}}  &
$\irule{C\vee u\yeq t \vee u^\prime\yeq t^\prime}
{(C\vee t\nyeq t^\prime \vee u\yeq t^\prime)\sigma}
\begin{array}{c}
(i),\ \ \forall L\in\{u^\prime\yeq t^\prime\}\cup C : (u\yeq t)\sigma\not\prec L\sigma
\end{array}$
\\\hline
\end{tabular}
\vskip 6pt
\begin{minipage}{\textwidth}
{\small where $\sigma$ is the most general unifier (mgu) of $u$ and $u^\prime$,
$u^\prime$ is not a variable in \textsl{Superposition} and
\textsl{Pa\-ra\-mo\-du\-la\-tion},
and the following abbreviations hold:
\begin{description}
\item[\emph{(i)}] is $u\sigma \not\preceq t\sigma$,
\item[\emph{(ii)}] is $\forall L\in D : (u\yeq t)\sigma\not\preceq L\sigma$,
\item[\emph{(iii)}] is $l[u^\prime]\sigma \not \preceq r\sigma$, and
\item[\emph{(iv)}] is $\forall L\in C : (l[u^\prime]\bowtie r)\sigma\not\preceq L\sigma$.
\end{description}
}
\end{minipage}
\end{center}
\caption{{\small Expansion inference rules of $\SP$.}}
\label{tab:sp-er} 
\end{figure}

\begin{figure}[tbh]
\begin{center}
\begin{tabular}{ccc}
{\small \textsl{Strict Subsumption}} &
$\srule{C ~~~~ D}{C}$ & $D\sub C$\\
\hline 
{\small \textsl{Simplification}} &
$\srule{C[u] ~~~~ l\yeq r}{C[r\sigma], ~~~ l\yeq r}$ &
$\begin{array}{c}
u=l\sigma,\ \ \ l\sigma\succ r\sigma,\ \ \ C[u]\succ (l\yeq r)\sigma
\end{array}$\\
\hline 
{\small \textsl{Deletion}} &
$\srule{C\vee t\yeq t}{}$ & \\
\hline
\end{tabular}
\vskip 6pt
\begin{minipage}{\textwidth}
{\small where $D\sub C$ if $D\subeq C$ and $C\not\subeq D$;
and $D\subeq C$ if $C\sigma\subseteq D$ (as multisets) for some substitution $\sigma$.
In practice, theorem provers such as \eprover{} apply also subsumption of variants:
if $D\subeq C$ and $C\subeq D$, the oldest clause is retained.}
\end{minipage}
\end{center}
\caption{{\small Contraction inference rules of $\SP$.}}
\label{tab:sp-cr}
\end{figure}

A well-founded ordering $\succ$ provides the basis for a notion of {\em redundancy}:
a ground clause $C$ is redundant in $S$
if for ground instances $\{D_1,\ldots D_k\}$ of clauses in $S$ it is
$\{D_1,\ldots D_k\}\models C$ and $\{D_1,\ldots D_k\}\prec \{C\}$;
a clause is redundant if all its ground instances are.
An inference is redundant if it uses or generates a redundant clause
and a set of clauses is {\em saturated} if all expansion inferences
in the set are redundant.
In $\SP$, clauses deleted by contraction are redundant
and expansion inferences that do not respect the ordering constraints are
redundant.

Let $\SP_\succ$ be $\SP$ with CSO $\succ$.
An {\em $\SP_\succ$-derivation} is a sequence of sets of clauses
$$S_0\der{\SP_\succ} S_1\der{\SP_\succ} \ldots S_i\der{\SP_\succ} \ldots$$
where at each step an $\SP_\succ$-inference is applied.
A derivation is characterized by its {\em limit},
defined as the set of {\em persistent clauses}
$$S_\infty = \bigcup_{j\ge 0} \bigcap_{i\ge j} S_i.$$
A derivation is {\em fair} if all expansion inferences become redundant eventually,
and a fair derivation generates a {\em saturated} limit.

Since inference systems are non-deterministic,
a {\em theorem-proving strategy} is obtained by adding a {\em search plan},
that drives rule application.
A search plan is fair if it only generates fair derivations.
An {\em $\SP_\succ$-strategy} is a theorem-proving strategy with inference
system $\SP_\succ$.
If the inference system is {\em refutationally complete}
and the search plan is {\em fair},
the theorem-proving strategy is {\em complete}:
$S_\infty$ is saturated and the empty clause $\Box$ is in $S_\infty$
if and only if $S_0$ is unsatisfiable.
A proof of the refutational completeness of $\SP$ can be found in \cite{NRHB}
and definitions and references for redundancy, saturation and fairness
in (e.g., \cite{TCScompletion,NRHB,TCLcanonicity}).

For additional notations and conventions used in the paper,
$Var(t)$ denotes the set of variables occurring in $t$;
the {\em depth} of a term $t$ is written $depth(t)$,
and $depth(t) = 0$, if $t$ is either a constant or a variable,
$depth(t) = 1 + max\{depth(t_i)\st 1\le i\le n\}$,
if $t$ is a compound term $f(t_1,\ldots, t_n)$.
A term is {\em flat} if its depth is $0$ or $1$.
For a literal, $depth(l\bowtie r) = depth(l) + depth(r)$.
A positive literal is {\em flat} if its depth is $0$ or $1$.
A negative literal is {\em flat} if its depth is $0$.
Let $\Gamma=\langle D, J\rangle$ be the interpretation
with domain $D$ and interpretation function $J$.
Since our usage of interpretations is fairly limited,
we use $\Gamma$ without specifying $D$ or $J$ whenever possible.
Lower case letters surmounted by a hat, such as $\hat d$ and $\hat e$,
denote elements of the domain $D$.
As usual, $[t]_\Gamma$ denotes the interpretation of term $t$ in $\Gamma$.
Generalizing this notation,
if $c$ is a constant symbol and $f$ a function symbol,
we use $[c]_\Gamma$ in place of $J(c)$ for the interpretation of $c$ in $\Gamma$
and $[f]_\Gamma$ in place of $J(f)$ for the interpretation of $f$ in $\Gamma$.
Small capital letters, such as \textsc{s}, denote sorts.
If there are many sorts,
$D$ is replaced by a tuple of sets, one per sort,
and $[\textsc{s}]_\Gamma$ denotes the one corresponding to \textsc{s} in $\Gamma$.

\section{Rewrite-based satisfiability procedures}\label{decidabletheories}

The rewriting approach to ${\cal T}$-satisfiability aims at applying
an inference system such as $\SP$ to clause sets $S_0 = \mathcal{T}\cup S$,
where $\mathcal{T}$ is a presentation of a theory and
$S$ a set of ground ${\cal T}$-literals.
This is achieved through the following phases, that, together, define
a {\em rewrite-based methodology for satisfiability procedures}:
\begin{enumerate}
\item\textit{${\cal T}$-reduction:}
specific inferences, depending on ${\cal T}$,
are applied to the problem to remove
certain literals or symbols and obtain an
equisatisfiable {\em ${\cal T}$-reduced} problem.
\item\textit{Flattening:} all ground literals are transformed into
flat literals, or flattened,
by introducing new constants and new equations,
yielding an equisatisfiable ${\cal T}$-reduced \emph{flat} problem.
For example, a literal $store(a_1,i_1,v_1) \yeq store(a_2,i_2,v_2)$
is replaced by the literals
$store(a_1,i_1,v_1) \yeq c_1$, $store(a_2,i_2,v_2) \yeq c_2$ and $c_1 \yeq c_2$.
Depending on ${\cal T}$,
flattening may precede or follow ${\cal T}$-reduction.
\item\textit{Ordering selection and termination:}
$\SP_\succ$ is shown to generate finitely many clauses
when applied fairly to a ${\cal T}$-reduced flat problem.
Such a result may depend on simple properties of the ordering $\succ$:
an ordering that satisfies them is termed {\em ${\cal T}$-good},
and an $\SP_\succ$-strategy is \emph{${\cal T}$-good} if $\succ$ is.
It follows that a fair ${\cal T}$-good $\SP_\succ$-strategy is guaranteed to terminate
on a ${\cal T}$-reduced flat problem.
The ${\cal T}$-goodness requirement may be vacuous,
meaning that any CSO is ${\cal T}$-good.
\end{enumerate}

This methodology can be \emph{fully automated},
except for the proof of termination and the definition of ${\cal T}$-goodness:
indeed, ${\cal T}$-reduction is made of mechanical inferences,
flattening is a mechanical operation,
and contemporary theorem provers feature mechanisms to generate automatically
orderings for given signatures and with given properties.

Let ${\cal E}$ denote the empty presentation, that is,
the presentation of the \textit{quantifier-free theory of equality}.
If ${\cal T}$ is $\mathcal{E}$,
$S$ is a set of ground equational literals built from free function
and constant symbols,
and $\mathcal{SP}_\succ$ reduces to ground completion,
which is guaranteed to terminate, with no need of flattening,
${\cal T}$-reduction or ${\cal T}$-goodness.
Therefore, any fair $\SP_\succ$-strategy is a satisfiability
procedure for the quantifier-free theory of equality.
In the rest of this section we apply the rewrite-based methodology
to several theories. For each theory, the signature contains the
function symbols indicated and a finite set of constant symbols.
A supply of countably many new constant symbols is assumed to be
available for flattening.

\subsection{The theory of records}\label{sec:records}

Records aggregate attribute-value pairs.
Let $\textit{Id}=\{id_1,\ldots, id_n\}$ be a set of attribute identifiers and
$\textsc{t}_1,\ldots, \textsc{t}_n$ be $n$ sorts.
Then, $\textsc{rec}(id_1:\textsc{t}_1,\ldots,id_n:\textsc{t}_n)$,
abbreviated $\textsc{rec}$, is the sort of records
that associate a value of sort $\textsc{t}_i$ to the attribute
identifier $id_i$, for $1\le i\le n$.
The signature of the \emph{theory of records}
has a pair of function symbols
$\rselect_i:\textsc{rec}\rightarrow\textsc{t}_i$ and
$\rstore_i:\textsc{rec}\times\textsc{t}_i\rightarrow\textsc{rec}$
for each $i$, $1\le i\le n$.
The presentation, named
$\mathcal{R}(id_1:\textsc{t}_1,\ldots,id_n:\textsc{t}_n)$,
or $\mathcal{R}$ for short,
is given by the following axioms,
where $x$ is a variable of sort $\textsc{rec}$
and $v$ is a variable of sort $\textsc{t}_i$:
\begin{eqnarray}
\label{r1}
\forall x, v.\ \rselect_i(\rstore_i(x,v)) \yeq v & \text{for all $i$, $1\le i\le n$}\\
\label{r2}
\forall x, v.\ \rselect_j(\rstore_i(x, v)) \yeq \rselect_j(x) & \text{for all $i,j$,
$1\le i\neq j\le n$}
\end{eqnarray}

For the \emph{theory of records with extensionality}, the presentation,
named $\mathcal{R}^e$, includes also the following axiom,
that states that two records are equal if all their fields are:
\begin{eqnarray}
\label{r:ext}
  \forall x, y.\ (\bigwedge_{i=1}^n \rselect_i(x) \yeq \rselect_i(y) \imp x \yeq y)
\end{eqnarray}
where $x$ and $y$ are variables of sort $\textsc{rec}$.
${\cal R}$ and $\mathcal{R}^e$ are Horn theories,
and therefore they are convex.
We begin with ${\cal R}$-reduction,
that allows us to reduce $\mathcal{R}^e$-satisfiability to ${\cal R}$-satisfiability:

\begin{textofdefinition}\label{reduction:record}
A set of ground ${\cal R}$-literals is {\em ${\cal R}$-reduced}
if it contains no literal $l\nyeq r$,
where $l$ and $r$ are terms of sort \textsc{rec}.
\end{textofdefinition}

Given a set of ground ${\cal R}$-literals $S$
and a literal $L=l\nyeq r\in S$,
such that $l$ and $r$ are terms of sort \textsc{rec},
{\em ${\cal R}$-reduction} first
replaces $L$ by the clause
$$C_L = \bigvee_{i=1}^n\rselect_i(l)\nyeq\rselect_i(r)$$
that is the resolvent of $L$ and the clausal form of~(\ref{r:ext}).
Thus, if $S = S_1\uplus S_2$,
where $S_2$ contains the literals $l\nyeq r$ with $l$ and $r$ of sort \textsc{rec}
and $S_1$ all the other literals,
$S$ is replaced by $S_1\cup \{C_L \st L\in S_2\}$.
Then, this set of clauses is reduced into disjunctive normal form,
yielding a disjunction of ${\cal R}$-reduced sets of literals.
Let $Red_{{\cal R}}(S)$ denote the class of ${\cal R}$-reduced sets thus obtained.

\begin{textoflemma}\label{lemma:reduction:record}
Given a set of ground ${\cal R}$-literals $S$,
$\mathcal{R}^e\cup S$ is satisfiable if and only if
$\mathcal{R}\cup Q$ is, for some $Q\in Red_{{\cal R}}(S)$.
\end{textoflemma}
\emph{Proof:}\\
($\Leftarrow$)
Let $\Gamma$ be a many-sorted model of $\mathcal{R}\cup Q$.
The claim is that there exists an interpretation $\Gamma^\prime$
that satisfies $\mathcal{R}^e\cup S$.
The only non-trivial part is to show that
$\Gamma^\prime$ satisfies the extensionality axiom of $\mathcal{R}^e$,
because in order to satisfy extensionality
$\Gamma^\prime$ needs to interpret the equality predicate $\yeq$ also on records,
whereas $\Gamma$ does not.
To simplify notation, let $\sim_\Gamma$ stand for $[\yeq]_\Gamma$
and $\sim_{\Gamma^\prime}$ stand for $[\yeq]_{\Gamma^\prime}$.
Then let $\Gamma^\prime$ be the interpretation that is identical to $\Gamma$,
except that $\sim_{\Gamma^\prime}$ is defined as follows:
\begin{itemize}
\item for all $\hat a,\hat b\in [\textsc{rec}]_{\Gamma}$,
      $\hat a\sim_{\Gamma^\prime} \hat b$ if and only if
      $[\rselect_i]_{\Gamma}(\hat a) \sim_\Gamma [\rselect_i]_{\Gamma}(\hat b)$
      for all $i$, $1\le i\le n$, and
\item for all $\hat a,\hat b\in [\textsc{t}_i]_{\Gamma}$,
      $\hat a\sim_{\Gamma^\prime} \hat b$ if and only if $\hat a \sim_\Gamma \hat b$,
      for all $i$, $1\le i\le n$.
\end{itemize}
The relation $\sim_{\Gamma^\prime}$ is clearly an equivalence.
To prove that it is a congruence,
we only need to show that if $\hat a\sim_{\Gamma^\prime} \hat b$, then
$[\rstore_i]_{\Gamma}(\hat a,\hat e)\sim [\rstore_i]_{\Gamma}(\hat b,\hat e)$
for all $i$, $1\le i\le n$ and
$\hat e\in [\textsc{t}_i]_{\Gamma}$.
By way of contradiction,
assume that $\hat a\sim_{\Gamma^\prime} \hat b$, but
$[\rstore_i]_{\Gamma}(\hat a,\hat e)\not\sim_{\Gamma^\prime}
[\rstore_i]_{\Gamma}(\hat b,\hat e)$
for some $i$, $1\le i\le n$, and
$\hat e\in [\textsc{t}_i]_{\Gamma}$.
In other words, by definition of $\sim_{\Gamma^\prime}$, it is 
$[\rselect_k]_{\Gamma}([\rstore_i]_{\Gamma}(\hat a,\hat e))\not\sim_\Gamma
[\rselect_k]_{\Gamma}([\rstore_i]_{\Gamma}(\hat b,\hat e))$
for some $k$, $1\le k\le n$.
There are two cases: either $k = i$ or $k\neq i$.
If $k = i$, then, since $\Gamma$, whence $\Gamma^\prime$,
is a model of axiom~(\ref{r1}),
it follows that $\hat e\not\sim_\Gamma \hat e$, a contradiction. 
If $k\neq i$, since $\Gamma$, whence $\Gamma^\prime$,
is a model of axiom~(\ref{r2}),
it follows that
$[\rselect_k]_{\Gamma}(\hat a)\not\sim_\Gamma [\rselect_k]_{\Gamma}(\hat b)$,
which contradicts the assumption $a\sim_{\Gamma^\prime} b$.
Thus, $\sim_{\Gamma^\prime}$ is well defined,
and $\Gamma^\prime$ is a model of $\mathcal{R}^e\cup S$.\\
($\Rightarrow$) This case is simple and is omitted for brevity.
\hfill$\Box$
\vskip 6pt

Termination depends on a case analysis
showing that only certain clauses can be generated,
and resting on a simple assumption on the CSO:

\begin{textofdefinition}\label{good:ordering:record}
A CSO $\succ$ is {\em ${\cal R}$-good} if
$t \succ c$ for all ground compound terms $t$ and constants $c$.
\end{textofdefinition}

\noindent
Most orderings can meet this requirement easily:
for instance, for RPO's, it is sufficient to assume a precedence where
all constant symbols are smaller than all function symbols.

\begin{textoflemma}\label{lemma:caseAnalysis:record}
All clauses in the limit $S_\infty$
of the derivation $S_0\der{\SP} S_1 \ldots S_i\der{\SP} \ldots$
generated by a fair ${\cal R}$-good $\SP_\succ$-strategy
from $S_0=\mathcal{R}\cup S$,
where $S$ is an $\mathcal{R}$-reduced set of ground flat $\mathcal{R}$-literals,
belong to one of the following classes,
where
$r,r^\prime$ are constants of sort $\textsc{rec}$,
and $e,e^\prime$ are constants of sort $\textsc{t}_i$
for some $i$, $1\le i\le n$:
\begin{enumerate}
\item[i)] the empty clause;
\item[ii)] the clauses in $\mathcal{R}$:
\begin{enumerate}
\item[ii.a)] $\rselect_i(\rstore_i(x,v)) \yeq v$, for all $i$, $1\le i\le n$
\item[ii.b)] $\rselect_j(\rstore_i(x, v)) \yeq \rselect_j(x)$,
for all $i,j$, $1\le i\neq j\le n$;
\end{enumerate}
\item[iii)] ground flat unit clauses of the form:
\begin{enumerate}
\item[iii.a)] $r\yeq r^\prime$,
\item[iii.b)] $e\yeq e^\prime$,
\item[iii.c)] $e\nyeq e^\prime$,
\item[iii.d)] $\rstore_i(r,e) \yeq r^\prime$, for some $i$, $1\le i\le n$,
\item[iii.e)] $\rselect_i(r) \yeq e$, for some $i$, $1\le i\le n$;
\end{enumerate}
\item[iv)] $\rselect_i(r) \yeq \rselect_i(r^\prime)$,
for some $i$, $1\le i\le n$.
\end{enumerate}
\end{textoflemma}
\emph{Proof:}
we recall that inequalities $r \nyeq r^\prime$
are not listed in \emph{(iii)}, because $S$ is $\mathcal{R}$-reduced.
All clauses in the classes above are unit clauses,
and therefore have a unique maximal literal.
Since $\succ$ is an ${\cal R}$-good CSO,
the left side of each literal is maximal
(for \emph{(iii.a)}, \emph{(iii.b)}, \emph{(iii.c)} and \emph{(iv)}
this can be assumed without loss of generality).
The proof is by induction on the index $i$ of the sequence $\{{S_i}\}_i$.
For $i=0$, all clauses in $S_0$ are in \emph{(ii)} or \emph{(iii)}.
For the inductive case,
we assume the claim is true for $i$ and we prove it for $i+1$.
Equational factoring applies to a clause with at least two positive literals,
and therefore does not apply to unit clauses.
Reflection may apply only to a clause in \emph{(iii.c)} to yield the empty clause.
For binary inferences, we consider each class in turn:
\begin{itemize}
\item \emph{Inferences within \emph{(ii)}}:
None applies.
\item \emph{Inferences within \emph{(iii)}}:
The only possible inferences produce
ground flat unit clauses in \emph{(iii)} or the empty clause.
\item \emph{Inferences between a clause in \emph{(iii)} and
a clause in \emph{(ii)}}:
A superposition of a clause in \emph{(iii.d)}
into one in \emph{(ii.a)}
yields a clause of the form $\rselect_i(r^\prime) \yeq e$ which is in \emph{(iii.e)}.
A superposition of a clause in \emph{(iii.d)}
into one in \emph{(ii.b)}
yields a clause of the form $\rselect_j(r^\prime) \yeq \rselect_j(r)$
which is in \emph{(iv)}.
No other inferences apply.
\item \emph{Inferences between a clause in \emph{(iv)} and
a clause in \emph{(ii)-(iv)}}:
the only applicable inferences are
simplifications between clauses in \emph{(iii.a)}, \emph{(iii.e)} and \emph{(iv)}
which yield clauses in \emph{(iii.e)} or \emph{(iv)}.
\hfill$\Box$
\end{itemize}
\vskip 6pt

Since only finitely many clauses of the kinds
enumerated in Lemma~\ref{lemma:caseAnalysis:record} can be built from a finite signature,
the saturated limit $S_\infty$ is finite,
and a fair derivation is bound to terminate:

\begin{textoflemma}\label{lemma:termination:record}
A fair ${\cal R}$-good $\SP_\succ$-strategy is guaranteed to terminate
when applied to $\mathcal{R}\cup S$,
where $S$ is an ${\cal R}$-reduced set of ground flat ${\cal R}$-literals.
\end{textoflemma}

\begin{textoftheorem}\label{theorem:record}
A fair ${\cal R}$-good $\SP_\succ$-strategy is a polynomial satisfiability
procedure for $\mathcal{R}$ and an exponential satisfiability procedure for $\mathcal{R}^e$.
\end{textoftheorem}
\emph{Proof:}
it follows from Lemmas~\ref{lemma:reduction:record} and~\ref{lemma:termination:record}
that a fair ${\cal R}$-good $\SP_\succ$-strategy is a satisfiability
procedure for $\mathcal{R}$ and $\mathcal{R}^e$.
For the complexity, let $m$ be the number of subterms
occurring in the input set of literals $S$.
Let $h$ be the number of literals of $S$ in the form $l\nyeq r$,
with $l$ and $r$ terms of sort \textsc{rec}.
$Red_{{\cal R}}(S)$ contains $n^h$ sets,
where $n$ is a constant,
the number of field identifiers in the presentation.
Since $h$ is $O(m)$, $Red_{{\cal R}}(S)$ contains $O(n^m)$ sets.
Since flattening is a $O(m)$ operation,
after flattening there are $O(n^m)$ sets,
each with $O(m)$ subterms.
For each set, inspection of the types of clauses and inferences allowed by
Lemma~\ref{lemma:termination:record} shows that the number
of generated clauses is $O(m^2)$. In other words,
the size of the set of clauses during the derivation is
bound by a constant $k$ which is $O(m^2)$.
Since each inference step takes polynomial time in $k$,
the procedure is polynomial for each set,
and therefore for $\mathcal{R}$,
and exponential for $\mathcal{R}^e$.
\hfill$\Box$
\vskip 6pt

\subsection{The theory of integer offsets}\label{sec:ios}

The \emph{theory of integer offsets} is a fragment of the theory of
the integers, which is applied in verification (e.g., \cite{uclid}).
Its signature does not assume sorts,
or assumes a single sort for the integers,
and has two unary function symbols $\s$ and $\p$, that represent the
successor and predecessor functions, respectively.
Its presentation, named ${\cal I}$, is given by the following infinite set
of sentences, as in, e.g., \cite{dpllt}:
\begin{eqnarray}
\label{eqn:sp}
  \forall x.\s(\p(x)) &\yeq& x \\
\label{eqn:ps}
  \forall x.\p(\s(x)) &\yeq& x \\
\label{eqn:acyc}
  \forall x.\s^i(x) &\nyeq& x~~~~~~~~~\text{ for $i>0$}
\end{eqnarray}
where $\s^1(x)=\s(x)$, $\s^{i+1}(x)=\s(\s^i(x))$ for $i\ge 1$,
and the sentences in (\ref{eqn:acyc}) are called \emph{acyclicity axioms}.
For convenience, let $Ac=\{\forall x.\s^i(x) \nyeq x\st i>0\}$ and
$Ac(n)=\{\forall x.\s^i(x) \nyeq x\st 0 < i\le n\}$.
Like the theory of records, ${\cal I}$ is also Horn,
and therefore convex.

\paragraph*{Remark 1}
Axiom~(\ref{eqn:ps}) implies that $\s$ is {\em injective}:
\begin{eqnarray}
  \label{inj}
  \forall x, y.\ \s(x) \yeq \s(y) \imp x \yeq y
\end{eqnarray}
Indeed, consider the set $\{\p(\s(x)) \yeq x, \s(a)\yeq\s(b), a\nyeq b\}$,
where $\{\s(a)\yeq\s(b), a\nyeq b\}$ is the clausal form of the negation
of~(\ref{inj}).
Superposition of $\s(a)\yeq\s(b)$ into $\p(\s(x)) \yeq x$ generates
$\p(\s(b)) \yeq a$. Superposition of $\p(\s(b)) \yeq a$ into $\p(\s(x)) \yeq x$
generates $a\yeq b$, that contradicts $a\nyeq b$.

\begin{textofdefinition}\label{def:reduction:ios}
A set of ground flat ${\cal I}$-literals is {\em ${\cal I}$-reduced}
if it does not contain occurrences of $\p$.
\end{textofdefinition}

Given a set $S$ of ground flat ${\cal I}$-literals,
the symbol $\p$ may appear only in literals of the form $\p(c)\yeq b$.
Negative ground flat literals have the form $c\nyeq b$,
and therefore do not contain $\p$.
\textit{${\cal I}$-reduction} consists of replacing
every equation $\p(c)\yeq b$ in $S$ by $c\yeq \s(b)$.
The resulting ${\cal I}$-reduced form of $S$ is denoted $Red_{{\cal I}}(S)$.
${\cal I}$-reduction reduces satisfiability with respect
to ${\cal I}$ to satisfiability with respect to $Ac$, so that
axioms (\ref{eqn:sp}) and (\ref{eqn:ps}) can be removed,
provided lemma (\ref{inj}) is added:

\begin{textoflemma}\label{lemma:reduction1:ios}
Let $S$ be a set of ground flat ${\cal I}$-literals.
$\mathcal{I}\cup S$ is satisfiable if and only if
$Ac\cup \{(\ref{inj})\}\cup Red_{{\cal I}}(S)$ is.
\end{textoflemma}
\emph{Proof}:\\
($\Rightarrow$)
It follows from Remark 1 and the observation that
$\mathcal{I}\cup S\models Red_{{\cal I}}(S)$,
since $c\yeq \s(b)$ is a logical
consequence of $\mathcal{I}$ and $\p(c)\yeq b$,
as it can be generated by a
superposition of $\p(c)\yeq b$ into axiom $\s(\p(x))\yeq x$.\\
($\Leftarrow$)
Let $\Gamma$ be a model of $Ac\cup \{(\ref{inj})\}\cup Red_{{\cal I}}(S)$
and let $D$ be its domain.
We build a model $\Gamma^\prime$ of $\mathcal{I}\cup S$.
$\Gamma^\prime$ interprets all constants in $S$ in the same way as $\Gamma$ does.
The crucial point is that
$\p$ does not occur in $Ac\cup \{(\ref{inj})\}\cup Red_{{\cal I}}(S)$,
so that $\Gamma$ does not interpret it,
whereas $\Gamma^\prime$ should.
Since not all elements of $D$ may have predecessor in $D$ itself,
the domain $D^\prime$ of $\Gamma^\prime$ will be a superset of $D$,
containing as many additional elements as are needed to interpret $\p$.
We construct recursively families of sets $\{D_i\}_i$ and functions $\{s_i\}_i$
and $\{p_i\}_i$ in such a way that, at the limit, all elements have predecessor:
\begin{itemize}
\item Base case: $i=0$.
Let $D_0 = D$, $s_0 = [s]_\Gamma$ and $p_0 = \emptyset$.
By establishing $s_0 = [s]_\Gamma$,
the interpretation of $\s$ on $D$ is preserved:
for all $\hat d\in D$, $[s]_\Gamma(\hat d) = [s]_{\Gamma^\prime}(\hat d)$.
We start by partitioning $D_0$ into the subset $E_0$ of elements that are successors
of some other element, and the subset $F_0$ of those that are not:
$E_0 = \{\hat e \st s_0(\hat d) = \hat e\ for\ some\ \hat d\in D_0\}$
and $F_0 = D_0\setminus E_0$.
For all $\hat e\in E_0$,
we define $p_1(\hat e) = \hat d$ such that $s_0(\hat d) = \hat e$:
such a $\hat d$ exists by definition of $E_0$
and it is unique because $\Gamma\models (\ref{inj})$.
Thus, $p_1$ is well-defined on $E_0$.
Next, we define $p_1$ on $F_0$.
Let $D^\prime_0$ be a set disjoint from $D_0$ and let $\eta_0\colon F_0\to D^\prime_0$
be a bijection: intuitively, $\eta_0$ maps each element of $F_0$ to its predecessor
that is missing in $D_0$.
Indeed, for all $\hat d\in F_0$,
we define $p_1(\hat d) = \eta_0(\hat d)$.
Then we define $s_1$:
for all $\hat e\in D_0$, $s_1(\hat e) = s_0(\hat e)$ and
for all $\hat e\in D^\prime_0$, $s_1(\hat e) = \hat d$,
where $\hat d$ is the element such that $\eta_0(\hat d) = \hat e$.
Establishing that $D_1 = D_0\uplus D^\prime_0$ closes the base case.
\item
Recursive case:
suppose that for $i\ge 1$,
we have a $D_{i-1}\subseteq D_i$, where we have defined $s_i$
and $p_i$ in such way that for all $\hat d\in D_{i-1}$,
there exists an $\hat e\in D_i$ such that
$s_i(\hat d) = \hat e$ and $p_i(\hat e) = \hat d$.
On the other hand,
there may be elements in $D_i$ that are not successors of any other element,
so that their predecessor is not defined.
Thus, let
$E_i = \{\hat e \st s_i(\hat d) = \hat e\ for\ some\ \hat d\in D_i\}$
and $F_i = D_i\setminus E_i$.
Let $D^\prime_i$ be a set of new elements and
$\eta_i\colon F_i\to D^\prime_i$ a bijection.
Then,
let $D_{i+1} = D_i\uplus D^\prime_i$.
For $p_{i+1}$,
for all $\hat e\in E_i$, $p_{i+1}(\hat e) = \hat d$ such that $s_i(\hat d) = \hat e$
and for all $\hat d\in F_i$, $p_{i+1}(\hat d) = \eta_i(\hat d)$.
For $s_{i+1}$,
for all $\hat e\in D_i$, $s_{i+1}(\hat e) = s_i(\hat e)$ and
for all $\hat e\in D^\prime_i$, $s_{i+1}(\hat e) = \hat d$,
where $\hat d$ is the element such that $\eta_i(\hat d) = \hat e$.
\end{itemize}
Then we define $D^\prime = \bigcup_i D_i$,
$[\s]_{\Gamma^\prime} = \bigcup_i s_i$ and $[\p]_{\Gamma^\prime} = \bigcup_i p_i$.
We show that $\Gamma^\prime\models \mathcal{I}\cup S$.
Axioms~(\ref{eqn:sp}) and~(\ref{eqn:ps}) are satisfied,
since, by construction, for all $\hat e\in D^\prime$,
$[p]_{\Gamma^\prime}(\hat e) = \hat d$
and $[s]_{\Gamma^\prime}(\hat d) = \hat e$.
The only equations in $S\setminus Red_{{\cal I}}(S)$
are the $\p(c)\yeq b$ for which $\s(b)\yeq c\in Red_{{\cal I}}(S)$.
Let $[b]_{\Gamma^\prime} = \hat d$ and $[c]_{\Gamma^\prime} = \hat e$:
since $\Gamma\models\s(b)\yeq c$, it follows that $\hat e\in E_0$,
and $\Gamma^\prime\models\p(c)\yeq b$.
\hfill$\Box$
\vskip 6pt

\begin{textofexample}
Let $S=\{\s(c)\yeq c^\prime\}$ and let $\Gamma$ be the model with
domain $\IN$, such that $c$ is interpreted as $0$ and $c^\prime$ as $1$.
Then $D_0 = \IN$, $E_0=\IN\setminus\{0\}$ and $F_0=\{0\}$.
At the first step of the construction,
we can take $D_0^\prime = \{-1\}$,
and have $p_1(0) = -1$ and $s_1(-1) = 0$.
Then $F_1=\{-1\}$, and we can take $D_1^\prime = \{-2\}$,
and so on.
At the limit, $D^\prime$ is the set $\mathbb{Z}$ of the integers.
\end{textofexample}

The next step is to bound the number of axioms in $Ac$ needed to
solve the problem.
The intuition is that the bound will be given by the number of
elements whose successor is determined by a constraint $\s(c)\yeq c^\prime$ in $S$.
Such a constraint establishes that, in any model $\Gamma$ of $S$,
the successor of $[c]_\Gamma$ must be $[c^\prime]_\Gamma$.
We call {\em $\s$-free} an element that is not thus constrained:

\begin{textofdefinition}\label{def:s-free:ios}
Let $S$ be a satisfiable ${\cal I}$-reduced set of ground flat ${\cal I}$-literals,
$\Gamma$ be a model of $S$ with domain $D$ and
$C_S$ be the set of constants
$C_S = \{c \st \s(c)\yeq c^\prime \in S\}$.
An element $\hat d\in D$ is {\em $\s$-free} in $S$ for $\Gamma$,
if for no $c\in C_S$,
it is the case that $[c]_\Gamma = \hat d$.
\end{textofdefinition}

We shall see that it is sufficient to consider $Ac(n)$,
where $n$ is the cardinality of $C_S$.

\begin{textofexample}
If $S = \{\s(c_1)\yeq c_2, \s(c_1)\yeq c_3, c_2\yeq c_3\}$,
then $C_S = \{c_1\}$ and $\vert C_S \vert = 1$.
In the worst case, however,
all occurrences of $\s$ apply to different constants,
so that $\vert C_S \vert$ is the number of occurrences of $\s$ in $S$.
\end{textofexample}

We begin with a notion of {\em $\s$-path} that mirrors
the paths in the graph $(D, [s]_\Gamma)$ defined by an interpretation $\Gamma$:

\begin{textofdefinition}\label{def:s-path:ios}
Let $S$ be a satisfiable ${\cal I}$-reduced set of ground flat ${\cal I}$-literals
and let $\Gamma$ be a model of $S$ with domain $D$.
For all $m\ge 2$,
a tuple $\langle\hat d_1,\s,\hat d_2,\s,\ldots\hat d_m,\s\rangle$
is an {\em $\s$-path} of length $m$ if
\begin{enumerate}
\item $\forall i,j$, $1\le i\ne j\le m$, $\hat d_i\ne \hat d_j$ and
\item $\forall i$, $1\le i < m$, $\hat d_{i+1} = [s]_\Gamma(\hat d_i)$.
\end{enumerate}
It is an {\em $\s$-cycle} if, additionally, $[s]_\Gamma(\hat d_m) = \hat d_1$.
\end{textofdefinition}

\noindent
It is clear to see that $\Gamma\models Ac(n)$ if and only if $\Gamma$
has no {\em $\s$-cycles} of length smaller or equal to $n$.

\begin{textoflemma}\label{lemma:prepReduction2:ios}
Let $S$ be an ${\cal I}$-reduced set of ground flat ${\cal I}$-literals
with $\vert C_S\vert = n$.
If there is an $\s$-path $p$ of length $m>n$ in a model $\Gamma$ of $S$,
then $p$ includes an element that is $\s$-free in $S$.
\end{textoflemma}
\emph{Proof}: by way of contradiction,
assume that no element in
$p=\langle\hat d_1,\s,\hat d_2,\s,\ldots\hat d_m,\s\rangle$
is $\s$-free in $S$.
By Definition~\ref{def:s-free:ios},
this means that for all $j$, $1\le j\le m$,
there is a constant $c_j\in C_S$ such that $[c_j]_\Gamma = \hat d_j$.
Since by Definition~\ref{def:s-path:ios} all elements in an $\s$-path are distinct,
it follows that $C_S$ should contain at least $m$ elements, or $n\ge m$,
which contradicts the hypothesis that $m>n$.
\hfill$\Box$
\vskip 6pt

\begin{textoflemma}\label{lemma:reduction2:ios}
Let $S$ be an ${\cal I}$-reduced set of ground flat ${\cal I}$-literals
with $\vert C_S\vert = l$.
For $n\ge l$,
if $\mathit{Ac}(n)\cup \{(\ref{inj})\}\cup S$ is satisfiable then
$\mathit{Ac}(n+1)\cup \{(\ref{inj})\}\cup S$ is.
\end{textoflemma}
\emph{Proof}:
let $\Gamma=\langle D, J\rangle$ be a model of
$\mathit{Ac}(n)\cup \{(\ref{inj})\}\cup S$.
$\Gamma$ has no $\s$-cycles of length smaller or equal to $n$.
We build a model $\Gamma^\prime$ with no $\s$-cycles of length smaller
or equal to $n+1$.
Let $P = \{p \st p\ is\ an \s\!-\!cycle\ of\ length\ n+1\}$.
If $P=\emptyset$, $\Gamma\models \mathit{Ac}(n+1)$
and $\Gamma^\prime$ is $\Gamma$ itself.
If $P\ne\emptyset$, there is some $p\in P$.
Since $n+1 > l$,
by Lemma~\ref{lemma:prepReduction2:ios},
there is some $\hat d$ in $p$ that is $\s$-free in $S$ for $\Gamma$.
Let $E_p = \{\hat e_j \st j\ge 0\}$ be a set disjoint from $D$,
and let $J_p$ be the interpretation function that is identical to $J$,
except that $J_p(\s)(\hat d)=\hat e_0$ and
$J_p(\s)(\hat e_j)=\hat e_{j+1}$
for all $j \ge 0$.
By extending $D$ into $D\cup E_p$ and extending $J$ into $J_p$,
we obtain a model where the $\s$-cycle $p$ has been broken.
By repeating this transformation for all $p\in P$,
we obtain the $\Gamma^\prime$ sought for.
Indeed,
$\Gamma^\prime\models \mathit{Ac}(n+1)$, because it has no $\s$-cycles
of length smaller or equal to $n+1$.
$\Gamma^\prime\models S$, because it interprets constants in the same way
as $\Gamma$, and for each $\s(c)\yeq e\in S$, $[c]_\Gamma$ is not $\s$-free in $S$,
which means $[\s(c)]_{\Gamma^\prime} = [\s(c)]_\Gamma$.
To see that $\Gamma^\prime\models (\ref{inj})$,
let $\hat d$ and $\hat d^\prime$ be two elements such that
$[\s]_{\Gamma^\prime}(\hat d) = [\s]_{\Gamma^\prime}(\hat d^\prime)$.
If $[\s]_{\Gamma^\prime}(\hat d)\in D$,
then $[\s]_{\Gamma^\prime}(\hat d) = [\s]_\Gamma(\hat d)$,
so that $\hat d = \hat d^\prime$,
because $\Gamma\models (\ref{inj})$.
If $[\s]_{\Gamma^\prime}(\hat d)\not\in D$,
then $[\s]_{\Gamma^\prime}(\hat d) = \hat e$ for some $\hat e$
introduced by the above construction,
so that $\hat d$ is the unique element whose successor is $\hat e$,
and $\hat d = \hat d^\prime$.
\hfill$\Box$
\vskip 6pt

By compactness, we have the following:

\begin{textofcorollary}\label{corollary:reduction2:ios}
Let $S$ be an ${\cal I}$-reduced set of ground flat ${\cal I}$-literals
with $\vert C_S \vert = n$.
$\mathit{Ac}\cup \{(\ref{inj})\}\cup S$ is satisfiable if and only if
$\mathit{Ac}(n)\cup \{(\ref{inj})\}\cup S$ is.
\end{textofcorollary}
\emph{Proof}:
the ``only if'' direction is trivial and for the ``if'' direction
induction using Lemma~\ref{lemma:reduction2:ios}
shows that for all $k \ge 0$,
if $\mathit{Ac}(n)\cup \{(\ref{inj})\}\cup S$ is satisfiable,
then so is $\mathit{Ac}(n+k)\cup \{(\ref{inj})\}\cup S$.
\hfill$\Box$
\vskip 6pt

\begin{textofdefinition}\label{good:ordering:ios}
A CSO $\succ$ is {\em ${\cal I}$-good} if
$t \succ c$ for all constants $c$ and
all terms $t$ whose root symbol is $\s$.
\end{textofdefinition}

\noindent
For instance, a precedence where all constant symbols are smaller than $\s$
will yield an ${\cal I}$-good RPO.

\begin{textoflemma}\label{lemma:caseAnalysis:ios}
All clauses in the limit $S_\infty$
of the derivation $S_0\der{\SP} S_1 \ldots S_i\der{\SP} \ldots$
generated by a fair ${\cal I}$-good $\SP_\succ$-strategy
from $S_0=\mathit{Ac}(n)\cup \{(\ref{inj})\}\cup S$,
where $S$ is an $\mathcal{I}$-reduced set of ground flat $\mathcal{I}$-literals
with $\vert C_S \vert = n$,
belong to one of the following classes,
where
$b_1, \ldots b_k$, $d_1, \ldots d_k$, $c$, $d$ and $e$ are constants ($k\ge 0$):
\begin{enumerate}
\item[i)] the empty clause;
\item[ii)] the clauses in $\mathit{Ac}(n)\cup \{(\ref{inj})\}$:
\begin{enumerate}
\item[ii.a)] $\s^i(x) \nyeq x$, for all $i$, $0 < i\le n$,
\item[ii.b)] $\s(x) \nyeq \s(y) \vee x \yeq y$;
\end{enumerate}
\item[iii)] ground flat unit clauses of the form:
\begin{enumerate}
\item[iii.a)] $c\yeq d$,
\item[iii.b)] $c\nyeq d$,
\item[iii.c)] $\s(c)\yeq d$;
\end{enumerate}
\item[iv)] other clauses of the following form:
\begin{enumerate}
\item[iv.a)] $\s(x)\nyeq d\vee x\yeq c\vee \bigvee_{i=1}^k d_i\nyeq b_i$,
\item[iv.b)] $c\yeq e\vee \bigvee_{i=1}^k d_i\nyeq b_i$,
\item[iv.c)] $\bigvee_{i=1}^k d_i\nyeq b_i$,
\item[iv.d)] $\s(c)\yeq e\vee \bigvee_{i=1}^k d_i\nyeq b_i$,
\item[iv.e)] $\s^j(c)\nyeq e\vee \bigvee_{i=1}^k d_i\nyeq b_i$, $1 \le j\le n-1$.
\end{enumerate}
\end{enumerate}
\end{textoflemma}
\emph{Proof:}
since $\succ$ is a CSO,
the first literal is the only maximal literal in \emph{(ii.b)}.
Since it is ${\cal I}$-good,
the first literal is the only maximal literal
in \emph{(iv.a)}, \emph{(iv.d)} and \emph{(iv.e)}.
For the same reason,
the left hand side is maximal in the maximal literals in
\emph{(iii.c)}, \emph{(iv.a)}, \emph{(iv.d)} and \emph{(iv.e)}.
The proof is by induction on the sequence $\{{S_i}\}_i$.
For the base case, all clauses in $S_0$ are in \emph{(ii)} or \emph{(iii)}.
For the inductive case,
we consider all possible inferences,
excluding upfront equational factoring,
which applies to a clause with at least two positive literals,
and therefore does not apply to Horn clauses.
\begin{itemize}
\item \emph{Inferences within \emph{(ii)}}:
Reflection applies to \emph{(ii.b)} to generate $x\yeq x$,
that gets deleted by deletion.
\item \emph{Inferences within \emph{(iii)}}:
The only possible inferences produce
ground flat unit clauses in \emph{(iii)} or the empty clause.
\item \emph{Inferences between a clause in \emph{(iii)} and
a clause in \emph{(ii)}}:
A paramodulation of an equality of kind \emph{(iii.c)}
into an inequality of type \emph{(ii.a)}
yields inequalities $\s^{i-1}(d) \nyeq c$, $1\le i\le n$,
that are in \emph{(iv.e)} with $k=0$ (for $i>1$) or \emph{(iii.b)} (for $i=1$).
A paramodulation of a \emph{(iii.c)} equality into \emph{(ii.b)}
yields $\s(x) \nyeq d\vee x\yeq c$
which is in \emph{(iv.a)} with $k=0$.
\item \emph{Inferences between a clause in \emph{(iv)} and a clause in \emph{(ii)}}:
A paramodulation of a clause in \emph{(iv.d)} into \emph{(ii.a)}
produces a clause in \emph{(iv.c)} or \emph{(iv.e)},
and a paramodulation of a clause in \emph{(iv.d)} into \emph{(ii.b)}
produces a clause in \emph{(iv.a)}.
\item \emph{Inferences between a clause in \emph{(iv)} and a clause in \emph{(iii)}}:
A simplification of a clause in \emph{(iv)} by an equality in
\emph{(iii.a)} or \emph{(iii.c)} generates another clause in \emph{(iv)}.
Paramodulating an equality of kind \emph{(iii.c)} into a \emph{(iv.a)} clause
yields a \emph{(iv.b)} clause.
Similarly, superposing a \emph{(iii.c)} unit with a \emph{(iv.d)} clause
gives a \emph{(iv.b)} clause.
The only possible remaining inferences between a clause in \emph{(iv)}
and one in \emph{(iii)} are paramodulations or superpositions of a \emph{(iv.b)} clause
into clauses in \emph{(iii)},
that add clauses in \emph{(iv.b)}, \emph{(iv.c)} and \emph{(iv.d)}.
\item \emph{Inferences within \emph{(iv)}}:
Reflection applies to clauses in \emph{(iv.b)} and \emph{(iv.c)}
to yield clauses in \emph{(iv.b)} or \emph{(iii.a)}
and \emph{(iv.c)} or \emph{(iii.b)}, respectively.
A paramodulation or superposition of a \emph{(iv.b)} clause
into a \emph{(iv.b)}, \emph{(iv.c)}, \emph{(iv.d)} or \emph{(iv.e)} clause
generates clauses also in \emph{(iv.b)}, \emph{(iv.c)}, \emph{(iv.d)} or \emph{(iv.e)},
respectively.
A superposition of a clause of kind \emph{(iv.d)} into a \emph{(iv.a)} clause
gives a clause in \emph{(iv.b)}.
A superposition between two \emph{(iv.d)} clauses adds a \emph{(iv.b)} clause.
A paramodulation of a clause of type \emph{(iv.d)} into a \emph{(iv.e)} clause
yields a clause in \emph{(iv.e)} or \emph{(iv.c)}.
\hfill$\Box$
\end{itemize}
\vskip 6pt

Given a finite signature,
only finitely many clauses of the types allowed by Lemma~\ref{lemma:caseAnalysis:ios}
can be formed. Thus, we have:

\begin{textoflemma}\label{lemma:termination:ios}
A fair ${\cal I}$-good $\SP_\succ$-strategy is guaranteed to terminate
when applied to $\mathit{Ac}(n)\cup \{(\ref{inj})\}\cup S$,
where $S$ is an ${\cal I}$-reduced set of ground flat ${\cal I}$-literals
with $\vert C_S \vert = n$.
\end{textoflemma}

\begin{textoftheorem}\label{theorem:ios}
A fair ${\cal I}$-good $\SP_\succ$-strategy is an exponential
satisfiability procedure for $\mathcal{I}$.
\end{textoftheorem}
\emph{Proof:} the main result follows from Lemma~\ref{lemma:reduction1:ios},
Corollary~\ref{corollary:reduction2:ios}
and Lemma~\ref{lemma:termination:ios}.
For the complexity, let $m$ be the number of subterms
occurring in the input set of literals $S$.
$Red_{{\cal I}}(S)$ has the same number of subterms as $S$,
since ${\cal I}$-reduction replaces literals of the form
$\p(c)\yeq b$ by literals of the form $c\yeq \s(b)$.
Flattening is $O(m)$.
The number $n$ of retained acyclicity axioms, according to
Lemma~\ref{lemma:termination:ios}, is also $O(m)$,
since in the worst case it is given by the number of occurrences of $\s$ in $S$.
By the proof of Lemma~\ref{lemma:caseAnalysis:ios},
at most $h=O(m^2)$ distinct literals
and at most $O(2^h)$ clauses can be generated.
Thus,
the size of the database of clauses during the derivation is
bound by a constant $k$ which is $O(2^h)$.
Since each inference step takes polynomial time in $k$,
the overall procedure is $O(2^{m^2})$.
\hfill$\Box$
\vskip 6pt

\begin{textofcorollary}\label{corollary:ac}
A fair $\SP_\succ$-strategy is a polynomial
satisfiability procedure for the theory presented by
the set of \emph{acyclicity axioms} $Ac$.
\end{textofcorollary}
\emph{Proof:} the proof of Lemma~\ref{lemma:caseAnalysis:ios}
shows that if the input includes only $\mathit{Ac}(n)\cup S$,
the only generated clauses are finitely many ground flat unit clauses from $S$
(inferences within \emph{(iii)}),
and finitely many equalities in the form $\s^{i-1}(d) \nyeq c$, for $1\le i\le n$,
by paramodulation of equalities $\s(c) \yeq d\in S$ into axioms
$\s^i(x) \nyeq x$ (inferences between a clause in \emph{(iii)}
and a clause in \emph{(ii)}).
It follows that
the number of clauses generated during the derivation is $O(m^2)$,
where $m$ is the number of subterms occurring in the input set of literals.
The size of the database of clauses during the derivation is
bound by a constant $k$ which is $O(m^2)$, and
since each inference step takes polynomial time in $k$,
a polynomial procedure results.
\hfill$\Box$
\vskip 6pt

\subsection{The theory of integer offsets modulo}\label{sec:iosmod}

The above treatment extends to the \emph{theory of integer offsets modulo},
which makes possible to describe data structures
with indices ranging over the integers modulo $k$,
such as circular queues.
A presentation for this theory, named ${\cal I}_k$,
is obtained from ${\cal I}$ by replacing $Ac$
with the following $k$ axioms
\begin{eqnarray}
  \forall x.\s^i(x) \nyeq x & \text{ for $1\le i\le k-1$} \label{eqn:acyc_n}\\
  \forall x.\s^k(x) \yeq x\label{eqn:cyc}
\end{eqnarray}
where $k>1$.
${\cal I}_k$ also is Horn and therefore convex.

Definition~\ref{def:reduction:ios}
and Lemma~\ref{lemma:reduction1:ios} apply also to ${\cal I}_k$,
whereas Lemma~\ref{lemma:reduction2:ios} is no longer necessary,
because ${\cal I}_k$ is finite to begin with.
Termination is guaranteed by the following lemma,
where $C(k)=\{\forall x.\s^k(x) \yeq x\}$:

\begin{textoflemma}\label{lemma:termination:iosmod}
A fair ${\cal I}$-good $\SP_\succ$-strategy is guaranteed to terminate
when applied to $\mathit{Ac}(k-1)\cup C(k)\cup \{(\ref{inj})\}\cup S$,
where $S$ is an ${\cal I}$-reduced set of ground flat ${\cal I}_k$-literals.
\end{textoflemma}
\emph{Proof:}
the proof of termination rests on the proof
of Lemma~\ref{lemma:caseAnalysis:ios}, with $n=k-1$
and the following additional cases to account for the presence of $C(k)$.
As far as inferences between axioms are concerned
(i.e., within group \emph{(ii)} in the proof of Lemma~\ref{lemma:caseAnalysis:ios}),
$C(k)$ does not introduce any,
because $\s^k(x) \yeq x$ cannot paramodulate into $\s^i(x)\nyeq x$, since $i<k$,
and cannot paramodulate into (\ref{inj}), since $k>1$.
For inferences between axioms and literals in $S$
(i.e., groups \emph{(ii)} and \emph{(iii)} in the proof
of Lemma~\ref{lemma:caseAnalysis:ios}),
the presence of $C(k)$ introduces superpositions of literals $\s(c)\yeq d\in S$
into $\s^k(x) \yeq x$,
generating $\s^{i-1}(d) \yeq c$ for $1\le i\le k$.
If we use $j$ in place of $i-1$ and $n$ in place of $k-1$,
we have $\s^j(d) \yeq c$ for $0\le j\le n$.
Excluding the cases $j=0$ and $j=1$ that are already covered
by classes \emph{(iii.a)} and \emph{(iii.c)}
of Lemma~\ref{lemma:caseAnalysis:ios},
we have an additional class of clauses,
with respect to those of Lemma~\ref{lemma:caseAnalysis:ios}:
\begin{enumerate}
\item[\emph{v)}] $\s^j(d) \yeq c$ for $2\le j\le n$.
\end{enumerate}
Thus, we only need to check the inferences induced by clauses of type \emph{(v)}.
There are only two possibilities.
Paramodulations of equalities in \emph{(v)} into inequalities in \emph{(ii.a)}
gives more clauses in \emph{(iv.e)} with $k=0$.
Paramodulations of equalities in \emph{(v)} into clauses in \emph{(iv.e)}
gives more clauses in \emph{(iv.c)} or \emph{(iv.e)}.
Since only finitely many clauses of types \emph{(i-v)}
can be formed from a finite signature, termination follows.
\hfill$\Box$
\vskip 6pt

\begin{textoftheorem}\label{theorem:iosmod}
A fair ${\cal I}$-good $\SP_\succ$-strategy is an exponential
satisfiability procedure for ${\cal I}_k$.
\end{textoftheorem}
\emph{Proof:} it follows the same pattern of the proof
of Theorem~\ref{theorem:ios},
with Lemma~\ref{lemma:termination:iosmod}
in place of Lemma~\ref{lemma:termination:ios}.
\hfill$\Box$
\vskip 6pt

Alternatively, since ${\cal I}_k$ is finite,
it is possible to omit ${\cal I}$-reduction and show termination on
the original problem format.
The advantage is that it is not necessary to include the injectivity
property (\ref{inj}), so that the resulting procedure is polynomial.
Furthermore, abandoning the framework of ${\cal I}$-reduction,
that was conceived to handle the infinite presentation of ${\cal I}$,
one can add axioms for $\p$ that are dual of (\ref{eqn:acyc_n}) and (\ref{eqn:cyc}),
resulting in the presentation ${\cal I}_k^\prime$ made of
(\ref{eqn:sp}), (\ref{eqn:ps}), (\ref{eqn:acyc_n}), (\ref{eqn:cyc}) and:
\begin{eqnarray}
  \forall x.\p^i(x) \nyeq x & \text{ for $1\le i\le k-1$} \label{eqn:acyc_n:p}\\
  \forall x.\p^k(x) \yeq x\label{eqn:cyc:p}
\end{eqnarray}
with $k > 1$. ${\cal I}_k^\prime$ is also Horn and therefore convex.

\begin{textofdefinition}\label{good:ordering:iosmod}
A CSO $\succ$ is {\em ${\cal I}_k^\prime$-good} if
$t \succ c$ for all ground compound terms $t$ and constants $c$.
\end{textofdefinition}

\begin{textoflemma}\label{lemma:termination:iosmod:poly}
A fair ${\cal I}_k^\prime$-good $\SP_\succ$-strategy is guaranteed to terminate
when applied to ${\cal I}_k^\prime\cup S$,
where $S$ is a set of ground flat ${\cal I}_k^\prime$-literals.
\end{textoflemma}
\emph{Proof:}
termination follows from the general observation that the only persistent clauses,
that can be generated by $\SP_\succ$ from ${\cal I}_k^\prime\cup S$,
are unit clauses $l\bowtie r$,
such that $l$ and $r$ are terms in the form $\s^j(u)$ or $\p^j(u)$,
where $0\le j\le k-1$ and $u$ is either a constant or a variable.
Indeed, if a term in this form with $j\ge k$ were generated,
it would be simplified by axioms
(\ref{eqn:cyc}) $\s^k(x) \yeq x$ or (\ref{eqn:cyc:p}) $\p^k(x) \yeq x$.
Similarly,
if a term where $\s$ is applied over $\p$ or vice versa were generated,
it would be simplified by axioms
(\ref{eqn:sp}) $\s(\p(x)) \yeq x$ or (\ref{eqn:ps}) $\p(\s(x)) \yeq x$.
Given a finite number of constants
and with variants removed by subsumption,
the bound on term depth represented by $k$ implies that there are only finitely
many such clauses.
\hfill$\Box$

\ignore{
Paramodulations between
$\s(\p(x)) \yeq x$ (\ref{eqn:sp}) and $\p(\s(x)) \yeq x$ (\ref{eqn:ps})
generate the trivial equations $\p(x) \yeq \p(x)$ and $\s(x) \yeq \s(x)$.

Paramodulation of $\s(\p(x)) \yeq x$ (\ref{eqn:sp})
into $\s^i(x) \nyeq x$ (\ref{eqn:acyc_n}) and further paramodulations of
$\s(\p(x)) \nyeq x$ (\ref{eqn:sp}) into the clauses thus generated generate
clauses in the class $\s^{i-j}(x) \nyeq \p^j(x)$ for $1\le j\le i$ and $1\le i\le k-1$.

Paramodulation of $\s(\p(x)) \yeq x$ (\ref{eqn:sp})
into $\s^k(x) \yeq x$ (\ref{eqn:cyc}) and further paramodulations of
$\s(\p(x)) \yeq x$ (\ref{eqn:sp}) into the clauses thus generated generate
clauses in the class $\s^{k-j}(x) \yeq \p^j(x)$ for $1\le j\le k-1$.

Paramodulation of $\p(\s(x)) \yeq x$ (\ref{eqn:ps})
into $\p^i(x) \nyeq x$ (\ref{eqn:acyc_n:p}) and further paramodulations of
$\p(\s(x)) \yeq x$ (\ref{eqn:ps}) into the clauses thus generated generate
clauses in the class $\p^{i-j}(x) \nyeq \s^j(x)$ for $1\le j\le i$ and $1\le i\le k-1$.

Paramodulation of $\p(\s(x)) \yeq x$ (\ref{eqn:ps})
into $\p^k(x) \yeq x$ (\ref{eqn:cyc:p}) and further paramodulations of
$\p(\s(x)) \yeq x$ (\ref{eqn:ps}) into the clauses thus generated generate
clauses in the class $\p^{k-j}(x) \yeq \s^j(x)$ for $1\le j\le k-1$.
}

\begin{textoftheorem}\label{theorem:iosmod:poly}
A fair ${\cal I}_k^\prime$-good $\SP_\succ$-strategy is a polynomial
satisfiability procedure for ${\cal I}_k^\prime$.
\end{textoftheorem}
\emph{Proof:} termination was established in
Lemma~\ref{lemma:termination:iosmod:poly}.
To see that the procedure is polynomial,
let $m$ be the number of subterms
in the input set of ground literals.
After flattening, we have $O(m)$ subterms,
and since ${\cal I}_k^\prime$ has $O(k)$ subterms,
the input to the $\SP_\succ$-strategy has $O(m+k)$ subterms.
By the proof of Lemma~\ref{lemma:termination:iosmod:poly},
only unit clauses are generated,
so that their number is $O((m+k)^2)$.
Since the size of the database of clauses during the derivation is
bound by a constant $h$ which is $O((m+k)^2)$,
and each inference takes polynomial time in $h$,
the overall procedure is polynomial.
\hfill$\Box$

\subsection{The theory of possibly empty lists}\label{sec:list}

Different presentations were proposed for a theory of {\em lists}.
A ``convex theory of $\cons$, $\car$ and $\cdr$,''
was studied by \cite{ShostakCC2},
and therefore it is named ${\cal L}_{Sh}$.
Its signature contains $\cons$, $\car$ and $\cdr$,
and its axioms are:
\begin{eqnarray}
  \label{ls1}
  \forall x, y.\ \car(\cons(x, y)) \yeq x\\
  \label{ls2}
  \forall x, y.\ \cdr(\cons(x, y)) \yeq y\\
  \label{lc}
  \forall y.\ \cons(\car(y),\cdr(y)) \yeq y
\end{eqnarray}

The presentation adopted by \cite{NO80},
hence called ${\cal L}_{NO}$,
adds the predicate symbol $\atom$ to the signature,
and the axioms
\begin{eqnarray}
  \label{la1}
  \forall x, y.\ \neg \atom(\cons(x, y))\\
  \label{la2}
  \forall y.\ \neg \atom(y)\imp \cons(\car(y),\cdr(y)) \yeq y
\end{eqnarray}
to axioms (\ref{ls1}) and (\ref{ls2}).

A third presentation also appeared in \cite{NO80},
but was not used in their congruence-closure-based algorithm.
Its signature features the constant symbol $\nil$,
together with $\cons$, $\car$ and $\cdr$, but not $\atom$.
This presentation, that we call ${\cal L}$,
adds to (\ref{ls1}) and (\ref{ls2}) the following four axioms:
\begin{eqnarray}
  \label{ln1}
  \forall x, y.\ \cons(x, y)\nyeq \nil\\
  \label{ln2}
  \forall y.\ y\nyeq \nil \imp \cons(\car(y),\cdr(y)) \yeq y\\
  \label{ln3}
  \car(\nil) \yeq \nil\\
  \label{ln4}
  \cdr(\nil) \yeq \nil
\end{eqnarray}
${\cal L}$ is not convex,
because $y\yeq \nil\vee \cons(\car(y),\cdr(y))\yeq y$
is in $\Th {\cal L}$, but neither disjunct is.

Unlike the presentation of records given earlier,
and that of arrays, that will be given in the next section,
these presentations of lists are unsorted,
or lists and their elements belong to the same sort.
This is desirable because it allows lists of lists.
Also, neither ${\cal L}_{Sh}$ nor ${\cal L}_{NO}$ nor ${\cal L}$
exclude cyclic lists (that is, a model of anyone of these presentations
can satisfy $\car(x) \yeq x$).
The rewriting approach was already applied
to both ${\cal L}_{Sh}$ and ${\cal L}_{NO}$ in \cite{ArRaRu2}.
The following analysis shows that it applies to ${\cal L}$ as well.

\begin{textofdefinition}\label{good:ordering:list}
A CSO $\succ$ is {\em ${\cal L}$-good} if
(1) $t \succ c$ for all ground compound terms $t$ and constants $c$,
(2) $t \succ \nil$ for all terms $t$ whose root symbol is $\cons$.
\end{textofdefinition}

It is sufficient to impose a precedence $>$,
such that function symbols are greater than constant symbols,
including $\cons > \nil$,
to make an RPO,
or a KBO with a simple weighting scheme
(e.g., weight given by arity),
${\cal L}$-good.
No ${\cal L}$-reduction is needed,
and the key result is the following:

\begin{textoflemma}\label{lemma:caseAnalysis:list}
All clauses in the limit $S_\infty$
of the derivation $S_0\der{\SP} S_1 \ldots S_i\der{\SP} \ldots$
generated by a fair ${\cal L}$-good $\SP_\succ$-strategy
from $S_0=\mathcal{L}\cup S$,
where $S$ is a set of ground flat $\mathcal{L}$-literals,
belong to one of the following classes,
where $c_i$ and $d_i$ for all $i$, $1\le i\le n$, and
$e_1,e_2,e_3$ are constants (constants include $\nil$):
\begin{enumerate}
\item[i)] the empty clause;
\item[ii)] the clauses in $\mathcal{L}$:
\begin{enumerate}
\item[ii.a)] $\car(\cons(x, y)) \yeq x$,
\item[ii.b)] $\cdr(\cons(x, y)) \yeq y$,
\item[ii.c)] $\cons(x, y)\nyeq \nil$,
\item[ii.d)] $\cons(\car(y),\cdr(y)) \yeq y \vee y\yeq \nil$,
\item[ii.e)] $\car(\nil) \yeq \nil$,
\item[ii.f)] $\cdr(\nil) \yeq \nil$;
\end{enumerate}
\item[iii)] ground flat unit clauses of the form:
\begin{enumerate}
\item[iii.a)] $c_1\yeq c_2$,
\item[iii.b)] $c_1\nyeq c_2$,
\item[iii.c)] $\car(c_1) \yeq c_2$,
\item[iii.d)] $\cdr(c_1) \yeq c_2$,
\item[iii.e)] $\cons(c_1,c_2) \yeq c_3$;
\end{enumerate}
\item[iv)] non-unit clauses of the following form:
\begin{enumerate}
\item[iv.a)] $\cons(e_1, \cdr(e_2)) \yeq e_3 \vee \bigvee_{i=1}^n c_i\bowtie d_i$,
\item[iv.b)] $\cons(\car(e_1), e_2) \yeq e_3 \vee \bigvee_{i=1}^n c_i\bowtie d_i$,
\item[iv.c)] $\cons(\car(e_1), \cdr(e_2)) \yeq e_3 \vee \bigvee_{i=1}^n c_i\bowtie d_i$,
\item[iv.d)] $\cons(e_1, e_2) \yeq e_3 \vee \bigvee_{i=1}^n c_i\bowtie d_i$,
\item[iv.e)] $\car(e_1) \yeq \car(e_2) \vee \bigvee_{i=1}^n c_i\bowtie d_i$,
\item[iv.f)] $\cdr(e_1) \yeq \cdr(e_2) \vee \bigvee_{i=1}^n c_i\bowtie d_i$,
\item[iv.g)] $\car(e_1) \yeq e_2 \vee \bigvee_{i=1}^n c_i\bowtie d_i$,
\item[iv.h)] $\cdr(e_1) \yeq e_2 \vee \bigvee_{i=1}^n c_i\bowtie d_i$,
\item[iv.i)] $\bigvee_{i=1}^n c_i\bowtie d_i$.
\end{enumerate}
\end{enumerate}
\end{textoflemma}
\emph{Proof:}
since $\succ$ is an ${\cal L}$-good CSO,
each clause in the above classes has a unique maximal literal,
which is the first one in the above listing
(up to a permutation of indices for \emph{(iv.i)}).
Furthermore,
the left side in each maximal literal is maximal
(for \emph{(iii.a)}, \emph{(iii.b)}, \emph{(iv.e)}, \emph{(iv.f)}, \emph{(iv.i)}
this can be assumed without loss of generality).
The proof is by induction on the sequence $\{{S_i}\}_i$.
For the base case, input clauses are in \emph{(ii)} or \emph{(iii)}.
For the inductive case, we consider all classes in order:
\begin{itemize}
\item \emph{Inferences within \emph{(ii)}}:
All inferences between axioms generate clauses that get deleted.
Superposition of \emph{(ii.a)} into \emph{(ii.d)} generates
$\cons(x,\cdr(\cons(x,y)))\yeq \cons(x,y)\vee\cons(x,y)\yeq \nil$,
which is simplified by \emph{(ii.b)} to
$\cons(x,y)\yeq \cons(x,y)\vee\cons(x,y)\yeq \nil$,
which is deleted.
Superposition of \emph{(ii.d)} into \emph{(ii.a)} produces
$\car(y)\yeq \car(y)\vee y\yeq \nil$
which is deleted.
Superposition of \emph{(ii.b)} into \emph{(ii.d)} yields
$\cons(\car(\cons(x,y)),y)\yeq \cons(x,y)\vee\cons(x,y)\yeq \nil$,
whose simplification by \emph{(ii.a)} gives
$\cons(x,y)\yeq \cons(x,y)\vee\cons(x,y)\yeq \nil$,
which is deleted.
Superposition of \emph{(ii.d)} into \emph{(ii.b)} generates
$\cdr(y)\yeq \cdr(y)\vee y\yeq \nil$
which gets deleted.
Paramodulation of \emph{(ii.d)} into \emph{(ii.c)} produces
the tautology $y\yeq\nil\vee y\nyeq\nil$,
which is eliminated by a step of reflection followed by one of deletion.
Superposition of \emph{(ii.e)} into \emph{(ii.d)} yields
$\cons(\nil,\cdr(\nil))\yeq\nil\vee\nil\yeq\nil$
which is deleted.
Similarly, superposition of \emph{(ii.f)} into \emph{(ii.d)} yields
$\cons(\car(\nil),\nil)\yeq\nil\vee\nil\yeq\nil$
which is also deleted,
and no other inferences apply among axioms.
\item \emph{Inferences within \emph{(iii)}}:
Inferences on the maximal terms in \emph{(iii)} can generate
only more ground flat unit clauses like those in \emph{(iii)} or the empty clause.
\item \emph{Inferences between a clause in \emph{(iii)} and
a clause in \emph{(ii)}}:
Inferences between an axiom and a ground flat unit clause generate
either more ground flat unit clauses or non-unit clauses in the classes
\emph{(iv.a)} and \emph{(iv.b)}.
Indeed, the only applicable inferences are:
superposition of a unit of kind \emph{(iii.c)} into \emph{(ii.d)},
which gives a clause in the form
$\cons(c_2,\cdr(c_1))\yeq c_1\vee c_1\yeq\nil$ of class \emph{(iv.a)};
superposition of a unit of kind \emph{(iii.d)} into \emph{(ii.d)},
which gives a clause in the form
$\cons(\car(c_1),c_2)\yeq c_1\vee c_1\yeq\nil$ of class \emph{(iv.b)};
superposition of a unit of kind \emph{(iii.e)}
into \emph{(ii.a)}, \emph{(ii.b)}, \emph{(ii.c)},
which generates unit clauses in \emph{(iii.c)}, \emph{(iii.d)} and \emph{(iii.b)},
respectively.
\item \emph{Inferences between a clause in \emph{(iv)} and a clause in \emph{(ii)}}:
We consider the clauses in \emph{(ii)} in order.
For \emph{(ii.a)}:
superposing a clause of kind \emph{(iv.a)} or \emph{(iv.d)} into \emph{(ii.a)} generates
$\car(e_3)\yeq e_1\vee \bigvee_{i=1}^n c_i\bowtie d_i$, that is in \emph{(iv.g)};
superposing a clause of kind \emph{(iv.b)} or \emph{(iv.c)} into \emph{(ii.a)} generates
$\car(e_1)\yeq \car(e_3)\vee \bigvee_{i=1}^n c_i\bowtie d_i$
that is in \emph{(iv.e)}.
For \emph{(ii.b)}:
superposing a clause of kind \emph{(iv.a)} or \emph{(iv.c)} into \emph{(ii.b)} generates
$\cdr(e_2)\yeq \cdr(e_3)\vee \bigvee_{i=1}^n c_i\bowtie d_i$
that is in \emph{(iv.f)};
superposing a clause of kind \emph{(iv.b)} or \emph{(iv.d)} into \emph{(ii.b)} generates
$\cdr(e_3)\yeq e_2\vee \bigvee_{i=1}^n c_i\bowtie d_i$, that is in \emph{(iv.h)}.
Paramodulating clauses of classes \emph{(iv.a)},
\emph{(iv.b)}, \emph{(iv.c)} and \emph{(iv.d)} into
\emph{(ii.c)} gives clauses in class \emph{(iv.i)}.
For \emph{(ii.d)}:
a superposition of \emph{(ii.d)} into \emph{(iv.c)} or \emph{(iv.c)} into \emph{(ii.d)}
yields a clause in \emph{(iv.i)};
a superposition of \emph{(iv.e)} or \emph{(iv.f)} into \emph{(ii.d)} produces
$\cons(\car(e_2),\cdr(e_1))\yeq e_1\vee e_1\yeq\nil\vee \bigvee_{i=1}^n c_i\bowtie d_i$
or $\cons(\car(e_1),\cdr(e_2))\yeq e_1\vee e_1\yeq\nil\vee \bigvee_{i=1}^n c_i\bowtie d_i$
that are in \emph{(iv.c)};
a superposition of \emph{(iv.g)} into \emph{(ii.d)} produces
$\cons(e_2,\cdr(e_1))\yeq e_1\vee e_1\yeq\nil\vee \bigvee_{i=1}^n c_i\bowtie d_i$
that is in \emph{(iv.a)};
a superposition of \emph{(iv.h)} into \emph{(ii.d)} produces
$\cons(\car(e_1),e_2)\yeq e_1\vee e_1\yeq\nil\vee \bigvee_{i=1}^n c_i\bowtie d_i$
that is in \emph{(iv.b)}.
Clause \emph{(ii.e)} can simplify clauses in \emph{(iv.b)}, \emph{(iv.c)}, \emph{(iv.e)}, \emph{(iv.g)},
to clauses in \emph{(iv.d)}, \emph{(iv.a)}, \emph{(iv.g)}, \emph{(iv.i)}, respectively.
Clause \emph{(ii.f)} can simplify clauses in \emph{(iv.a)}, \emph{(iv.c)}, \emph{(iv.f)}, \emph{(iv.h)},
to clauses in \emph{(iv.d)}, \emph{(iv.b)}, \emph{(iv.h)}, \emph{(iv.i)}, respectively.
No other inferences apply.
\item \emph{Inferences between a clause in \emph{(iv)} and a clause in \emph{(iii)}}:
The only possible expansion inference here is a paramodulation of a clause in \emph{(iv.i)}
into a clause in \emph{(iii.b)}, which generates another clause in \emph{(iv.i)}.
All other possible steps are simplifications,
where an equality of class \emph{(iii)} reduces a
clause in \emph{(iv)} to another clause in \emph{(iv)}.
\item \emph{Inferences within \emph{(iv)}}:
Reflection applied to a clause in \emph{(iv)} generates either the empty clause
or a clause in \emph{(iv)}.
Equational factoring applies only to a clause in \emph{(iv.i)},
to yield another clause of the same kind.
The only other applicable inferences are superpositions that generate more clauses in \emph{(iv)}.
Specifically,
clauses of kind \emph{(iv.a)} superpose with clauses in
\emph{(iv.a)}, \emph{(iv.f)}, \emph{(iv.h)} and \emph{(iv.i)}
to generate clauses in \emph{(iv.i)}, \emph{(iv.a)} and \emph{(iv.d)}.
Clauses of kind \emph{(iv.b)} superpose with clauses in
\emph{(iv.b)}, \emph{(iv.e)}, \emph{(iv.g)} and \emph{(iv.i)}
to generate clauses in \emph{(iv.i)}, \emph{(iv.b)} and \emph{(iv.d)}.
Clauses of kind \emph{(iv.c)} superpose with clauses in \emph{(iv.c)}, \emph{(iv.e)}, \emph{(iv.f)},
\emph{(iv.g)}, \emph{(iv.h)} and \emph{(iv.i)}
to generate clauses in \emph{(iv.i)}, \emph{(iv.c)}, \emph{(iv.a)} and \emph{(iv.b)}.
Clauses of kind \emph{(iv.d)} superpose with clauses in \emph{(iv.d)} and \emph{(iv.i)}
to generate clauses in \emph{(iv.i)} and \emph{(iv.d)}.
Clauses of kind \emph{(iv.e)} superpose with clauses in
\emph{(iv.e)}, \emph{(iv.g)} and \emph{(iv.i)}
to generate clauses in \emph{(iv.e)} and \emph{(iv.g)}.
Clauses of kind \emph{(iv.f)} superpose with clauses in
\emph{(iv.f)}, \emph{(iv.h)} and \emph{(iv.i)}
to generate clauses in \emph{(iv.f)} and \emph{(iv.h)}.
Clauses of kind \emph{(iv.g)} superpose with clauses in
\emph{(iv.g)} and \emph{(iv.i)}
to generate clauses in \emph{(iv.i)} and \emph{(iv.g)}.
Clauses of kind \emph{(iv.h)} superpose with clauses in
\emph{(iv.h)} and \emph{(iv.i)}
to generate clauses in \emph{(iv.i)} and \emph{(iv.h)}.
Clauses of kind \emph{(iv.i)} superpose with clauses in \emph{(iv.i)}
to generate clauses in \emph{(iv.i)}.
\hfill$\Box$
\end{itemize}
\vskip 6pt

It follows that the limit is finite and a fair derivation is bound to halt:

\begin{textoflemma}\label{lemma:termination:list}
A fair ${\cal L}$-good $\SP_\succ$-strategy is guaranteed to terminate
when applied to $\mathcal{L}\cup S$,
where $S$ is a set of ground flat ${\cal L}$-literals.
\end{textoflemma}

\begin{textoftheorem}\label{theorem:list}
A fair ${\cal L}$-good $\SP_\succ$-strategy is an exponential
satisfiability procedure for $\mathcal{L}$.
\end{textoftheorem}
\emph{Proof:}
let $m$ be the number of subterms
occurring in the input set of literals.
After flattening the number of subterms is $O(m)$.
The types of clauses listed in Lemma~\ref{lemma:caseAnalysis:list}
include literals of depth at most $2$
(cf. \emph{(iv.a)}, \emph{(iv.b)} and \emph{(iv.c)}).
Hence, at most $h=O(m^3)$ distinct literals
and at most $O(2^h)$ clauses can be generated.
It follows that the size of the set of clauses during the derivation
is bound by a constant $k$ which is $O(2^h)$.
Since applying an inference takes polynomial time in $k$,
the overall complexity is $O(2^{m^3})$.
\hfill$\Box$
\vskip 6pt

\noindent
Exponential complexity was expected,
because it was shown already in \cite{NO80}
that the satisfiability problem for ${\cal L}$ is NP-complete.

\subsection{The theory of arrays}\label{sec:array}

Let $\textsc{index}$, $\textsc{elem}$ and
$\textsc{array}$ be the sorts of
indices, elements and arrays, respectively.
The signature has two function symbols,
$\select:\textsc{array}\times\textsc{index}\rightarrow\textsc{elem}$,
and $\store:\textsc{array}\times\textsc{index}\times\textsc{elem}
\rightarrow\textsc{array}$, with the usual meaning.
The standard presentation, denoted $\mathcal{A}$, is made of two axioms,
where $x$ is a variable of sort $\textsc{array}$, $w$ and $z$ are variables
of sort $\textsc{index}$ and $v$ is a variable of sort $\textsc{elem}$:
\begin{eqnarray}
  \label{ss1}
  \forall x, z, v.\ \select(\store(x, z, v), z) \yeq v\\
  \label{ss2}
  \forall x, z, w, v.\ (z \nyeq w \imp 
   \select(\store(x, z, v), w) \yeq \select(x, w))
\end{eqnarray}
This theory also is not convex,
because $z \yeq w \vee \select(\store(x, z, v), w) \yeq \select(x, w))$
is valid in the theory, but neither disjunct is.
For the \emph{theory of arrays with extensionality},
the presentation, named $\mathcal{A}^e$,
includes also the extensionality axiom
\begin{eqnarray}
  \label{ext}
  \forall x, y.\ (\forall z.\select(x,z) \yeq \select(y,z) \imp x \yeq y)
\end{eqnarray}
where $x$ and $y$ are variables of sort $\textsc{array}$,
and $z$ is a variable of sort $\textsc{index}$.

\begin{textofdefinition}\label{reduction:array}
A set of ground ${\cal A}$-literals is {\em ${\cal A}$-reduced}
if it contains no literal $l\nyeq r$,
where $l$ and $r$ are terms of sort \textsc{array}.
\end{textofdefinition}

Given a set of ground ${\cal A}$-literals $S$,
{\em ${\cal A}$-reduction} consists of replacing every literal
$l\nyeq r\in S$, where $l$ and $r$ are terms of sort \textsc{array},
by $select(l,sk_{l,r})\nyeq select(r,sk_{l,r})$,
where $sk_{l,r}$ is a Skolem constant of sort $\textsc{index}$.
The resulting ${\cal A}$-reduced form of $S$,
denoted $Red_{{\cal A}}(S)$,
is related to the original problem by the following
(cf. Lemma 7.1 in \cite{ArRaRu2}):

\begin{textoflemma}\label{lemma:reduction:array}
(Armando, Ranise and Rusinowitch 2003)
Let $S$ be a set of ground $\mathcal{A}$-literals.
$\mathcal{A}^e\cup S$ is satisfiable if and only if
$\mathcal{A}\cup Red_{{\cal A}}(S)$ is.
\end{textoflemma}

\begin{textofdefinition}\label{good:ordering:array}
A CSO $\succ$ is {\em ${\cal A}$-good} if
(1) $t \succ c$ for all ground compound terms $t$ and constants $c$,
and
(2) $a \succ e \succ j$, for all constants $a$ of sort $\textsc{array}$,
$e$ of sort $\textsc{elem}$ and $j$ of sort $\textsc{index}$.
\end{textofdefinition}

\noindent
If $\succ$ is an RPO,
it is sufficient to impose a precedence $>$,
such that function symbols are greater than constant symbols,
and $a > e > j$ for all constants $a$ of sort $\textsc{array}$,
$e$ of sort $\textsc{elem}$ and $j$ of sort $\textsc{index}$,
for $\succ$ to be ${\cal A}$-good.
If it is a KBO,
the same precedence and a simple choice of weights will do.

\begin{textoflemma}\label{lemma:caseAnalysis:array}
All clauses in the limit $S_\infty$
of the derivation $S_0\der{\SP} S_1 \ldots S_i\der{\SP} \ldots$
generated by a fair $\mathcal{A}$-good $\SP_\succ$-strategy
from $S_0=\mathcal{A}\cup S$,
where $S$ is an $\mathcal{A}$-reduced set of ground flat $\mathcal{A}$-literals,
belong to one of the following classes,
where
$a,a^\prime$ are constants of sort \textsc{array},
$i$,$i_1,\ldots,i_n$,$i_1^\prime,\ldots,i_n^\prime$,
$j_1,\ldots,j_m$,$j_1^\prime,\ldots,j_m^\prime$
are constants of sort \textsc{index} ($n,m\geq 0$),
$e, e^\prime$ are constants of sort \textsc{elem},
and $c_1, c_2$ are constants of either sort \textsc{index}
or sort $\textsc{elem}$:
\begin{enumerate}
\item[i)] the empty clause; 
\item[ii)] the clauses in $\mathcal{A}$:
\begin{enumerate}
\item[ii.a)] $\select(\store(x, z, v), z) \yeq v$ and
\item[ii.b)] $\select(\store(x, z, v), w) \yeq \select(x, w) \vee z \yeq w$;
\end{enumerate}
\item[iii)] ground flat unit clauses of the form:
\begin{enumerate}
\item[iii.a)] $a \yeq a^\prime$,
\item[iii.b)] $c_1 \yeq c_2$,
\item[iii.c)] $c_1 \nyeq c_2$,
\item[iii.d)] $\store(a,i,e) \yeq a^\prime$,
\item[iii.e)] $\select(a,i) \yeq e$; and
\end{enumerate}
\item[iv)] non-unit clauses of the following form:
\begin{enumerate}
\item[iv.a)] $\select(a,x)\yeq \select(a^\prime,x)
\vee x\yeq i_1 \vee \ldots \vee x\yeq i_n
\vee j_1\bowtie j_1^\prime \vee \ldots \vee j_m\bowtie j_m^\prime$,\\
for $x\in\textsc{index}$,
\item[iv.b)] $\select(a,i)\yeq e
\vee i_1\bowtie i_1^\prime \vee \ldots \vee i_n\bowtie i_n^\prime$,
\item[iv.c)] $e \yeq e^\prime
\vee i_1\bowtie i_1^\prime \vee \ldots \vee i_n\bowtie i_n^\prime$,
\item[iv.d)] $e \nyeq e^\prime
\vee i_1\bowtie i_1^\prime \vee \ldots \vee i_n\bowtie i_n^\prime$,
\item[iv.e)] $i_1 \yeq i_1^\prime
\vee i_2\bowtie i_2^\prime \vee\ldots \vee i_n\bowtie i_n^\prime$,
\item[iv.f)] $i_1 \nyeq i_1^\prime
\vee i_2\bowtie i_2^\prime \vee\ldots \vee i_n\bowtie i_n^\prime$,
\item[iv.g)] $t \yeq a^\prime
\vee i_1\bowtie i_1^\prime \vee \ldots \vee i_n\bowtie i_n^\prime$,
where $t$ is either $a$ or $\store(a,i,e)$.
\end{enumerate}
\end{enumerate}
\end{textoflemma}
\emph{Proof:}
we recall that inequalities $a \nyeq a^\prime$
are not listed in \emph{(iii)}, because $S$ is $\mathcal{A}$-reduced.
Since $\succ$ is total on ground terms and ${\cal A}$-good,
each clause in the above classes has a unique maximal literal,
which is the first one in the above listing
(up to a permutation of indices for \emph{(iv.e)} and \emph{(iv.f)}).
Classes \emph{(iv.e)} and \emph{(iv.f)} are really one class
separated in two classes based on the sign of the maximal literal.
The proof is by induction on the sequence $\{{S_i}\}_i$.
For the base case, input clauses are in \emph{(ii)} or \emph{(iii)}.
For the inductive case, we have:
\begin{itemize}
\item \emph{Inferences within \emph{(ii)}}:
The only inference that applies to the axioms in $\mathcal{A}$
is a superposition of \emph{(ii.a)} into \emph{(ii.b)}
that generates the trivial clause
$z\yeq z \vee \select(x,z)\yeq v$, which is eliminated by deletion.
\item \emph{Inferences within \emph{(iii)}}:
Inferences between ground flat unit clauses can produce
only ground flat unit clauses in \emph{(iii)} or the empty clause.
\item \emph{Inferences between a clause in \emph{(iii)} and
a clause in \emph{(ii)}}:
Superposition of \emph{(iii.d)} $\store(a,i,e) \yeq a^\prime$
into \emph{(ii.a)} $\select(\store(x,z,v),z) \yeq v$
yields $\select(a^\prime,i) \yeq e$ which is in \emph{(iii.e)}.
Superposition of \emph{(iii.d)}
into \emph{(ii.b)} $\select(\store(x,z,v),w) \yeq \select(x,w)\vee z \yeq w$
yields $\select(a^\prime,w) \yeq \select(a,w) \vee i \yeq w$
which is in \emph{(iv.a)}.
\item \emph{Inferences between a clause in \emph{(iv)} and
a clause in \emph{(ii)}}:
Superposition of \emph{(iv.g)} $\store(a,i,e) \yeq a^\prime
\vee i_1\bowtie i_1^\prime \vee \ldots \vee i_n\bowtie i_n^\prime$
into \emph{(ii.a)}
yields $\select(a^\prime,i) \yeq e
\vee i_1\bowtie i_1^\prime \vee \ldots \vee i_n\bowtie i_n^\prime$,
which is in \emph{(iv.b)}.
Superposition of \emph{(iv.g)}
into \emph{(ii.b)}
yields $\select(a^\prime,w) \yeq \select(a,w) \vee i \yeq w
\vee i_1\bowtie i_1^\prime \vee \ldots \vee i_n\bowtie i_n^\prime$,
which is in \emph{(iv.a)}.
No other inferences apply.
\item \emph{Inferences between a clause in \emph{(iv)} and
a clause in \emph{(iii)}}:
For an inference to apply to \emph{(iii.a)} and \emph{(iv)}
it must be that $a$ (or $a^\prime$) appears in a clause in \emph{(iv)}.
Similarly, for an inference to apply to \emph{(iii.b)} and \emph{(iv)}
it must be that $c_1$ (or $c_2$) appears in a clause in \emph{(iv)}.
In either case,
simplification of the clause of class \emph{(iv)}
by the clause of class \emph{(iii)} applies.
Such a step can only generate a clause in \emph{(iv)}.
The only inference that can apply to \emph{(iii.c)} and \emph{(iv)}
is a paramodulation of a clause in \emph{(iv)} into a clause
in \emph{(iii.c)}.
If $c_1, c_2\in \textsc{elem}$,
paramodulation of \emph{(iv.c)} into \emph{(iii.c)}
generates a clause in \emph{(iv.d)}.
If $c_1, c_2\in \textsc{index}$,
paramodulation of \emph{(iv.e)} into \emph{(iii.c)}
produces a clause in \emph{(iv.f)} or \emph{(iv.e)},
depending on the sign of the maximal literal in the resulting clause.
We consider next \emph{(iii.d)} and \emph{(iv)}.
The only possible application of simplification
consists of applying \emph{(iii.d)} to reduce a clause in \emph{(iv.g)}
to a clause in the same class.
A superposition of \emph{(iv.c)} or \emph{(iv.e)} into \emph{(iii.d)}
generates a clause in \emph{(iv.g)}.
No other inferences are possible.
Last come \emph{(iii.e)} and \emph{(iv)}.
As a simplifier, \emph{(iii.e)} may apply only to \emph{(iv.b)}
to yield a clause in \emph{(iv.c)}.
All possible superpositions,
namely superposition of \emph{(iii.e)} and \emph{(iv.a)},
superposition of \emph{(iv.e)} into \emph{(iii.e)},
and
superposition of \emph{(iv.g)} into \emph{(iii.e)}
give clauses of class \emph{(iv.b)}.
\item \emph{Inferences within \emph{(iv)}}:
Equational factoring applies only to a clause
of class \emph{(iv.e)} to yield a clause in \emph{(iv.e)} or \emph{(iv.f)}.
Reflection applies to a clause in \emph{(iv.d)} or \emph{(iv.f)},
to yield a clause in one of \emph{(iii.b)}, \emph{(iii.c)}, \emph{(iv.e)} or \emph{(iv.f)}.
Then, for each kind of clause we consider all binary inferences it can have
with clauses that follows in the list.
We begin with \emph{(iv.a)}:
superposition of \emph{(iv.a)} and \emph{(iv.a)} gives \emph{(iv.a)};
superposition of \emph{(iv.a)} and \emph{(iv.b)} gives \emph{(iv.b)};
superposition of \emph{(iv.g)} into \emph{(iv.a)} gives \emph{(iv.a)}.
Second comes \emph{(iv.b)}:
superposition of \emph{(iv.b)} and \emph{(iv.b)} gives \emph{(iv.c)};
superposition of \emph{(iv.e)} into \emph{(iv.b)} gives \emph{(iv.b)};
superposition of \emph{(iv.g)} into \emph{(iv.b)} gives \emph{(iv.b)}.
Next there is \emph{(iv.c)}:
superposition of \emph{(iv.c)} and \emph{(iv.c)} gives \emph{(iv.c)};
paramodulation of \emph{(iv.c)} into \emph{(iv.d)} gives \emph{(iv.d)};
superposition of \emph{(iv.c)} into \emph{(iv.g)} gives \emph{(iv.g)}.
For \emph{(iv.e)} and \emph{(iv.f)},
we have:
superposition of \emph{(iv.e)} and \emph{(iv.e)}
gives \emph{(iv.e)} or \emph{(iv.f)};
pa\-ra\-mo\-du\-la\-tion of \emph{(iv.e)} into \emph{(iv.f)}
gives \emph{(iv.e)} or \emph{(iv.f)};
superposition of \emph{(iv.e)} into \emph{(iv.g)} gives \emph{(iv.g)}.
Last, all possible applications of
superposition within \emph{(iv.g)} give \emph{(iv.g)}.
\hfill $\Box$
\end{itemize}
\vskip 6pt

Thus, we have (cf. Lemma 7.3 and Theorem 7.2 in \cite{ArRaRu2}):

\begin{textoflemma}\label{lemma:termination:array}
(Armando, Ranise and Rusinowitch 2003)
A fair ${\cal A}$-good $\SP_\succ$-strategy is guaranteed to terminate
when applied to $\mathcal{A}\cup S$,
where $S$ is an ${\cal A}$-reduced set of ground flat ${\cal A}$-literals.
\end{textoflemma}

\begin{textoftheorem}\label{theorem:array}
(Armando, Ranise and Rusinowitch 2003)
A fair ${\cal A}$-good $\SP_\succ$-strategy is an exponential
satisfiability procedure for $\mathcal{A}$ and $\mathcal{A}^e$.
\end{textoftheorem}

\section{Rewrite-based satisfiability: combination of theories}
\label{combination}

A big-engines approach is especially well-suited for the combination of theories,
because it makes it possible to combine presentations rather than algorithms.
The inference engine is the same for all theories considered,
and studying a combination of theories amounts to studying the behavior of the
inference engine on a problem in the combination.
In a little-engines approach, on the other hand,
there is in principle a different engine for each theory,
and studying a combination of theories may require studying
the interactions among different inference engines.

In the rewrite-based methodology,
the {\em combination problem} is the problem of showing
that an $\SP_\succ$-strategy decides ${\cal T}$-satisfiability,
where ${\cal T}=\bigcup_{i=1}^n{\cal T}_i$,
knowing that it decides ${\cal T}_i$-satisfiability
for all $i$, $1\le i\le n$.
Since ${\cal T}_i$-reduction applies separately for each theory,
and flattening is harmless,
one only has to prove termination.
The main theorem in this section establishes sufficient
conditions for $\SP_\succ$ to terminate on ${\cal T}$-satisfiability problems
if it terminates on ${\cal T}_i$-satisfiability problems
for all $i$, $1\le i\le n$.
A first condition is that the ordering $\succ$ be ${\cal T}$-{\em good}:

\begin{textofdefinition}\label{good:ordering:combination}
Let ${\cal T}_1,\ldots, {\cal T}_n$ be presentations of theories.
A CSO $\succ$ is ${\cal T}$-{\em good},
where ${\cal T}=\bigcup_{i=1}^n{\cal T}_i$,
if it is ${\cal T}_i$-{\em good} for all $i$, $1\le i\le n$.
\end{textofdefinition}

The second condition will serve the purpose of excluding
pa\-ra\-mo\-du\-la\-tions from variables,
when considering inferences across theories.
This is key, since a variable may paramodulate
into any proper non-variable subterm:

\begin{textofdefinition}\label{variable:inactive:clause}
A clause $C$ is {\em variable-inactive} for $\succ$
if no maximal literal in $C$ is an equation
$t\yeq x$ where $x\not\in Var(t)$.
A set of clauses is {\em variable-inactive} for $\succ$
if all its clauses are.
\end{textofdefinition}

\begin{textofdefinition}\label{variable:inactive:presentation}
A theory presentation ${\cal T}$ is {\em variable-inactive} for $\SP_\succ$
if the limit $S_\infty$ of any fair $\SP_\succ$-derivation from $S_0=\mathcal{T}\cup S$
is variable-inactive for $\succ$.
\end{textofdefinition}

\noindent
For satisfiability problems,
$S$ is ground,
hence immaterial for variable-inactivity.
If axioms persist, as generally expected,
${\cal T}\subseteq S_\infty$,
and Definition~\ref{variable:inactive:presentation}
requires that they are variable-inactive.
If they do not persist, they are irrelevant,
because a fair strategy does not need to perform inferences
from clauses that do not persist.

The third condition is that the signatures do not share function
symbols, which excludes paramodulations from compound terms.
Sharing of constant symbols, including those introduced by flattening,
is allowed.
Thus, the only inferences across theo\-ries
are pa\-ra\-mo\-du\-la\-tions from constants
into constants, that are finitely many:

\begin{textoftheorem}\label{theorem:combination:variable:inactive}
Let ${\cal T}_1,\ldots, {\cal T}_n$ be presentations
of theories, with no shared function symbol,
and let ${\cal T}=\bigcup_{i=1}^n{\cal T}_i$.
Assume that for all $i$, $1\le i\le n$,
$S_i$ is a ${\cal T}_i$-reduced set of ground flat ${\cal T}_i$-literals.
If for all $i$, $1\le i\le n$,
a fair ${\cal T}_i$-good $\SP_\succ$-strategy is guaranteed to terminate
on $\mathcal{T}_i\cup S_i$,
and ${\cal T}_i$ is variable-inactive for $\SP_\succ$,
then a fair ${\cal T}$-good $\SP_\succ$-strategy is guaranteed to terminate
on $\mathcal{T}\cup S_1\cup \ldots \cup S_n$.
\end{textoftheorem}
\emph{Proof:}
let $S_\infty^i$ be the set of persistent clauses
generated by $\SP_\succ$ from $\mathcal{T}_i\cup S_i$.
Since $\SP_\succ$ terminates on $\mathcal{T}_i\cup S_i$,
for all $i$, $1\le i\le n$,
we are concerned only with binary expansion inferences
between a clause in $S_\infty^i$ and a clause in $S_\infty^j$,
with $1\le i\ne j\le n$.
We consider first paramodulations from variables.
Assume that a literal $t\yeq x$ occurs in a clause $C$ in $S_\infty^i$.
If $x\in Var(t)$,
it is $t\succ x$ by the subterm property of the CSO,
and therefore, there is no paramodulation from $x$.
If $x\not\in Var(t)$,
$t\yeq x$ is not maximal in $C$,
because $S_\infty^i$ is variable-inactive by hypothesis.
In other words, there is another literal $L$ in $C$ such that $L\succ t\yeq x$.
By stability of $\succ$,
$L\sigma\succ (t\yeq x)\sigma$ for all substitutions $\sigma$.
Thus, no instance $(t\yeq x)\sigma$ can be maximal,
so that, again, there is no paramodulation from $x$.
Therefore, there are no paramodulations from variables.
Since there are no shared function symbols,
no paramodulation from a compound term applies to
a clause in $S_\infty^i$ and a clause in $S_\infty^j$.
The only possible inferences
are those where a clause
$a\yeq t\vee C$ paramodulates into
a clause $l[a]\bowtie u\vee D$,
where $a$ is a constant,
$t$ is also a constant
(it cannot be a variable,
because $a\yeq t\vee C$ is variable-inactive,
and it cannot be a compound term,
because $\succ$ is stable and good),
the context $l$ may be empty,
the mgu is empty,
and $C$ and $D$ are disjunctions of literals.
Since there are only finitely many constants,
only finitely many such steps may apply.
\hfill$\Box$
\vskip 6pt

\begin{textofcorollary}\label{corollary:combination:variable:inactive}
Let ${\cal T}_1,\ldots, {\cal T}_n$ be presentations of theories,
with no shared function symbol,
and let ${\cal T}=\bigcup_{i=1}^n{\cal T}_i$.
If for all $i$, $1\le i\le n$,
a fair ${\cal T}_i$-good $\SP_\succ$-strategy
is a satisfiability procedure for $\mathcal{T}_i$,
and $\mathcal{T}_i$ is variable-inactive for $\SP_\succ$,
then a fair ${\cal T}$-good $\SP_\succ$-strategy
is a satisfiability procedure for $\mathcal{T}$.
\end{textofcorollary}

The requirement of being variable-inactive is rather natural
for equational theories:

\begin{textoftheorem}\label{theorem:nonTrivMod:equational:variable:inactive}
If ${\cal T}$ is a presentation of an equational theory
with no trivial models, then ${\cal T}$ is variable-inactive
for $\SP_\succ$ for any CSO $\succ$.
\end{textoftheorem}
\emph{Proof:}
by way of contradiction, assume that ${\cal T}$ is not variable-inactive,
that is, for some variable inactive $S_0$,
$S_\infty$ is not variable inactive.
Thus, there is an equation $t\yeq x\in S_\infty$ such that $x\not\in Var(t)$.
Since $\SP$ is sound, $S_0\models t\yeq x$.
An equation $t\yeq x$ such that $x\not\in Var(t)$
is satisfied only by a trivial model.
Thus, $S_0$ has only trivial models.
Since ${\cal T}\subset S_0$,
a model of $S_0$ is also a model of ${\cal T}$.
It follows that ${\cal T}$ has trivial models,
contrary to the hypothesis.
\hfill$\Box$
\vskip 6pt

\noindent
Given an equational presentation ${\cal T}$,
the addition of the axiom $\exists x \exists y\ x\nyeq y$
is sufficient to exclude trivial models.
Since the clausal form of this axiom
is the ground flat literal $sk_1\nyeq sk_2$,
where $sk_1$ and $sk_2$ are two Skolem constants,
this addition preserves all termination results
for $\SP_\succ$ on ${\cal T}$-satisfiability problems.

For Horn theories,
refutational completeness is preserved if $\SP_\succ$
is specialized to a {\em maximal unit strategy},
that restricts superposition to unit clauses
and paramodulates unit clauses into maximal negative literals \cite{D90:ctrs}.
Equational factoring is not needed for Horn theories.
This strategy resembles positive unit resolution in the non-equational case
and has the same character of a purely forward-reasoning strategy.
At the limit, all proofs in $S_\infty$ are {\em valley proofs},
that is, equational rewrite proofs in the form $u\jjoin t$.
It follows that all non-unit clauses are redundant in $S_\infty$:

\begin{textoftheorem}\label{theorem:nonTrivMod:Horn:variable:inactive}
If ${\cal T}$ is a presentation of a Horn equational theory
with no trivial models, then ${\cal T}$ is variable-inactive
for $\SP_\succ$ for any CSO $\succ$ and the maximal unit strategy.
\end{textoftheorem}
\emph{Proof:}
it is the same as for Theorem~\ref{theorem:nonTrivMod:equational:variable:inactive},
because $S_\infty$ only contains unit clauses.
\hfill$\Box$
\vskip 6pt

For first-order theories,
the requirement that $S_\infty$ be variable-inactive
excludes the generation of clauses in the form
$a_1\yeq x\vee\ldots\vee a_n\yeq x$,
where for all $i$, $1\le i\le n$, $a_i$ is a constant.
Such a disjunction may be generated,
but only within a clause that contains at least one greater literal,
such as one involving function symbols
(e.g., clauses of type (\emph{iv.a}) in Lemma~\ref{lemma:caseAnalysis:array}).

\begin{textoftheorem}\label{theorem:nonTrivMod:firstOrder:variable:inactive}
Let ${\cal T}$ be a presentation of a first-order theory:
if $a_1\yeq x\vee\ldots\vee a_n\yeq x\in S_\infty$,
where $S_\infty$ is the limit of any fair $\SP_\succ$-derivation from
$S_0=\mathcal{T}\cup S$, for any CSO $\succ$,
then $\Th {\cal T}$ is not stably infinite.
Furthermore, if ${\cal T}$ has no trivial models,
$\Th {\cal T}$ is also not convex.
\end{textoftheorem}
\emph{Proof:}
since $\SP$ is sound,
$a_1\yeq x\vee\ldots\vee a_n\yeq x\in S_\infty$
implies
$S_0\models \forall x\ a_1\yeq x\vee\ldots\vee a_n\yeq x$.
It follows that $S_0$ has no infinite model.
On the other hand,
$a_1\yeq x\vee\ldots\vee a_n\yeq x\in S_\infty$
implies that $S_0$ is satisfiable,
because if $S_0$ were unsatisfiable,
by the refutational completeness of $\SP$,
$S_\infty$ would contain only the empty clause.
Thus, $S_0$ has models, but has no infinite model.
Equivalently,
$S$ has ${\cal T}$-models, but has no infinite ${\cal T}$-model.
This means that $\Th {\cal T}$ is not stably infinite,
and, if it has no trivial models,
it is also not convex by Theorem~\ref{convex:stablyInfinite}.
\hfill$\Box$
\vskip 6pt

\noindent
In other words,
if ${\cal T}$ is not variable-inactive for $\SP_\succ$,
because it generates a clause in the form $a_1\yeq x\vee\ldots\vee a_n\yeq x$,
then ${\cal T}$ is not stably infinite either.

The notion of a clause in the form $a_1\yeq x\vee\ldots\vee a_n\yeq x$
was ``lifted'' in \cite{IJCAR2006csp} to those of {\em variable clause}
and {\em cardinality constraint clause}.
A {\em variable clause} is a clause containing only equations between variables
or their negations.
The {\em antecedent-mgu}\footnote{The name derives from the sequent-style notation
for clauses adopted in \cite{IJCAR2006csp}.}
({\em a-mgu}, for short) of a variable clause $C$
is the most general unifier of the unification problem
$\{x\sunif y\st x\nyeq y\in C\}$.
Then, a variable clause $C$ is a {\em cardinality constraint clause},
if $C^+\mu$ is not empty and contains no trivial equation $x\yeq x$,
where $\mu$ is the a-mgu of $C$ and $C^+$ is made of the positive literals in $C$.
This notion allows one to prove the following (cf. Lemma 5.2 in \cite{IJCAR2006csp}):

\begin{textoflemma}
(Bonacina, Ghilardi, Nicolini, Ranise and Zucchelli 2006)\label{lemma5.2:IJCAR2006csp}
If $S_0$ is a finite satisfiable set of clauses,
then $S_0$ admits no infinite models if and only if
the limit $S_\infty$ of any fair $\SP_\succ$-derivation from $S_0$
contains a cardinality constraint clause.
\end{textoflemma}

Next, we note that a cardinality constraint clause
cannot be variable-inactive,
because it must have some positive literal in the form $x\yeq y$ that is maximal.
For example, in $z\nyeq y\vee x\yeq y\vee z\yeq w$,
all three literals are maximal. Thus, it follows that:

\begin{textoftheorem}\label{varInactive:stablyInfinite}
If a first-order theory ${\cal T}$ is variable-inactive for $\SP_\succ$,
then it is stably-infinite.
\end{textoftheorem}
\emph{Proof:}
assume that ${\cal T}$ is not stably-infinite.
Then there exists a quantifier-free ${\cal T}$-formula $\varphi$,
that has a ${\cal T}$-model but no infinite ${\cal T}$-model.
Let $S_0$ be the clausal form of ${\cal T}\cup\{\varphi\}$:
$S_0$ is finite, satisfiable and admits no infinite model.
By Lemma~\ref{lemma5.2:IJCAR2006csp},
the limit $S_\infty$ of a fair $\SP_\succ$-derivation from $S_0$
contains a cardinality constraint clause.
Thus, $S_\infty$, and hence ${\cal T}$, is not variable-inactive for $\SP_\succ$.
\hfill$\Box$
\vskip 6pt

We conclude by applying Theorem~\ref{theorem:combination:variable:inactive}
to any combination of the theories studied in Section~\ref{decidabletheories}.
The goodness requirement (Definition~\ref{good:ordering:combination})
is easily satisfied:
any CSO is good for ${\cal E}$,
and it is simple to obtain an ordering that is simultaneously
${\cal R}$-good, ${\cal I}$-good, ${\cal L}$-good and ${\cal A}$-good.
The reductions of ${\cal R}^e$ to ${\cal R}$ (Lemma~\ref{lemma:reduction:record})
and ${\cal A}^e$ to ${\cal A}$ (Lemma~\ref{lemma:reduction:array})
apply also when the signature contains free function symbols
$f\colon \textsc{s}_0,\ldots,\textsc{s}_{m-1}\to \textsc{s}_m$,
$m\ge 1$, provided that none of the $\textsc{s}_i$, $1\le i\le m$,
is \textsc{rec} or \textsc{array}, respectively.
A function symbol satisfying this requirement is said to be \textit{record-safe}
or \textit{array-safe}, respectively.
Thus, we have:

\begin{textoftheorem}\label{theorem:array+list+iosmod+ios+record+euf}
A fair $\SP_\succ$-strategy
is a satisfiability procedure for
any combination of the theories of
records, with or without extensionality,
integer offsets,
integer offsets modulo,
possibly empty lists,
arrays, with or without extensionality,
and the quantifier-free theory of equality,
provided
(1) $\succ$ is ${\cal R}$-good whenever records are included,
(2) $\succ$ is ${\cal I}$-good whenever integer offsets are included,
(3) $\succ$ is ${\cal L}$-good whenever lists are included and
(4) $\succ$ is ${\cal A}$-good whenever arrays are included,
and
(5) all free function symbols are array-safe (record-safe)
whenever arrays (records) with extensionality and
the quantifier-free theory of equality are included.
\end{textoftheorem}
\emph{Proof:}
for the quantifier-free theory of equality,
$\mathcal{E}$ is vacuously variable-inactive for $\SP_\succ$.
For the other theories,
the lists of clauses in Lemma~\ref{lemma:caseAnalysis:record},
Lemma~\ref{lemma:caseAnalysis:ios},
Lemma~\ref{lemma:termination:iosmod},
Lemma~\ref{lemma:caseAnalysis:list}
and Lemma~\ref{lemma:caseAnalysis:array},
show that ${\cal R}$, $\mathcal{I}$, $\mathcal{I}_k$, ${\cal L}$ and ${\cal A}$, respectively,
are variable-inactive for $\SP_\succ$.
Therefore, the result follows from Theorem~\ref{theorem:combination:variable:inactive}.
\hfill$\Box$
\vskip 6pt

\noindent
This theorem holds if ${\cal L}$ is replaced by ${\cal L}_{Sh}$ or ${\cal L}_{NO}$,
since they are also variable-inactive for $\SP_\succ$
(cf. Lemmata 4.1 and 5.1 in \cite{ArRaRu2}).  

\section{Synthetic benchmarks}\label{problems}

This section presents six sets of synthetic benchmarks:
three in the \textit{theory of arrays with extensionality}
($\STORECOMM$, $\SWAP$ and $\STOREINV$);
one in the
\textit{combination of the theories of arrays and integer offsets} ($\IOS$);
one in the
\textit{combination of the theories of arrays, records and integer offsets}
to model queues
($\QUEUE$);
and one in the
\textit{combination of the theories of arrays,
records and integer offsets modulo}
to model circular queues
($\CIRCULARQUEUE$).
Each problem set is \textit{parametric},
that is, it is formulated as a function $Pb$,
that takes a positive integer $n$ as parameter,
and returns a set of formul\ae\ $Pb(n)$.
For all these problems,
the size of $Pb(n)$ grows {\em monotonically} with $n$.
This property makes them ideal to evaluate empirically
how a system's performance scale with input's size,
as we shall do in Section~\ref{experiments}.

\subsection{First benchmark: $\STORECOMM(n)$ and $\STORECOMMINVALID(n)$}

The problems of the $\STORECOMM$ family express the fact
that the result of storing a set of elements in different positions
within an array is not affected by the relative order of the store operations.
For instance, for $n=2$ the following valid formula belongs to
$\STORECOMM(n)$:
\begin{displaymath}
i_1\nyeq i_2\imp store(store(a,i_1,e_1),i_2,e_2)\yeq store(store(a,i_2,e_2),i_1,e_1).
\end{displaymath}
Here and in the following 
$a$ is a constant of sort $\textsc{array}$,
$i_1,\ldots,i_n$ are constants of sort $\textsc{index}$,
and $e_1,\ldots,e_n$ are constants of sort $\textsc{elem}$.

In general, let $n>0$ and $p,q$ be permutations of $\{1,\ldots,n\}$.
Let $\STORECOMM(n,p,q)$ be the formula:
\begin{displaymath}
\bigwedge_{(l,m)\in C^n_2} i_l\nyeq i_m \imp (T_n(p)\yeq T_n(q))
\end{displaymath}
where $C^n_2$ is the set of 2-combinations over $\{1,\ldots,n\}$ and
\begin{equation}
    T_k(p)=\left\{
  \begin{array}{ll}
    a & \text{if $k=0$}\\
    store(T_{k-1}(p),i_{p(k)},e_{p(k)}) & \text{if $1\leq k\leq n$.}
  \end{array}
\right.
\end{equation}
Since only the relative position of the elements of $p$ with respect
to those of $q$ is relevant, $q$ can be fixed.
For simplicity, let it be the identity permutation $\iota$.
Then
$\STORECOMM(n)= \{\STORECOMM(n,p,\iota)\st
p \text{ is a permutation of }\ \{1,\ldots,n\}\}$.

\begin{textofexample}\label{example:storecomm}
If $n=3$ and $p$ is such that $p(1)=3$, $p(2)=1$, and $p(3)=2$, then
$$T_n(p)=store(store(store(a,i_3,e_3),i_1,e_1),i_2,e_2)$$
$$T_n(\iota)=store(store(store(a,i_1,e_1),i_2,e_2),i_3,e_3))$$
and
$\STORECOMM(n,p,\iota)$ is
$$((i_1\nyeq i_2\wedge i_2\nyeq i_3\wedge i_1\nyeq i_3)\imp$$
$$store(store(store(a,i_3,e_3),i_1,e_1),i_2,e_2)\yeq 
store(store(store(a,i_1,e_1),i_2,e_2),i_3,e_3)).$$
\end{textofexample}

Each element of $\STORECOMM(n)$, once negated, reduced to clausal form,
$\mathcal{A}$-reduced and flattened,
leads to a problem whose number of clauses is in $O(n^2)$,
because it is dominated by the
${n\choose 2} = \frac{n(n-1)}{2}$ clauses in $\{i_l\nyeq i_m : (l,m)\in C^n_2\}$.
\ignore{
In fact, transformation into clausal form of the negation of
\begin{displaymath}
\bigwedge_{(l,m)\in C^n_2} i_l\nyeq i_m \imp (T_n(p)\yeq T_n(\iota))
\end{displaymath}
yields the set of clauses
$\{i_l\nyeq i_m : (l,m)\in C^n_2\}\cup\{T_n(p)\nyeq T_n(\iota)\}$.
Of these clauses,
the last one is subject to $\mathcal{A}$-reduction to yield the clause
\begin{equation}
  \label{goal_clause}
  select(T_n(p),sk(T_n(p),T_n(\iota)))\nyeq 
  select(T_n(\iota),sk(T_n(p),T_n(\iota)))
\end{equation}
Flattening clause (\ref{goal_clause}) generates $2n+4$ clauses:
the equations
$c_k\yeq T_k(p)$ and $c_{n+k}\yeq T_k(\iota)$ for $1\leq k\leq n$,
the equations
$c_{2n+1}\yeq sk(c_n,c_{2n})$, $c_{2n+2}\yeq select(c_n,c_{2n+1})$,
$c_{2n+3}\yeq select(c_{2n},c_{2n+1})$,
and the inequation $c_{2n+2}\nyeq c_{2n+3}$,
where $c_1,\ldots, c_{2n+3}$ are new constants.
}

A slight change in the definition of $\STORECOMM$ generates
sets of formul\ae\ that are not valid in ${\cal A}$:
$$\STORECOMMINVALID(n)= \{\STORECOMM(n,p,\iota^\prime)\st
\text{$p$ is a permutation of $\{1,\ldots,n\}$}\}$$
where
$\iota^\prime:\{1,\ldots,n\}\rightarrow\{1,\ldots,n\}$ is such that
$\iota^\prime(k)=k$ for $1\le k\le n-1$ and $\iota^\prime(n)=1$.

\begin{textofexample}\label{example:storecomminvalid}
For $n$, $p$ and $T_n(p)$ as in Ex.~\ref{example:storecomm},
$$T_n(\iota^\prime)=store(store(store(a,i_1,e_1),i_2,e_2),i_1,e_1))$$
and $\STORECOMMINVALID(n,p,\iota^\prime)$ is
$$((i_1\nyeq i_2\wedge i_2\nyeq i_3\wedge i_1\nyeq i_3)\imp$$
$$store(store(store(a,i_3,e_3),i_1,e_1),i_2,e_2)\yeq
store(store(store(a,i_1,e_1),i_2,e_2),i_1,e_1)).$$
\end{textofexample}

\subsection{Second benchmark: $\SWAP(n)$ and $\SWAPINVALID(n)$}

An elementary property of arrays is that swapping an
element at position $i_1$ with an element at position $i_2$ is
equivalent to swapping the element at position $i_2$ with the element
at position $i_1$.
The problems of the $\SWAP$ family are based on
generalizing this observation to any number $n$ of swap operations.
For instance, for $n=2$ the following valid fact is in
$\SWAP(n)$:
\begin{displaymath}
  swap(swap(a,i_0,i_1),i_2,i_1)\yeq swap(swap(a,i_1,i_0),i_1,i_2)
\end{displaymath}
where $swap(a,i,j)$ abbreviates the term
$store(store(a,i,select(a,j)),j,select(a,i))$.

In general, let $c_1,c_2$ be subsets of $\{1,\ldots,n\}$, and let
$p,q$ be functions $p,q:\{1,\ldots,n\}\rightarrow \{1,\ldots,n\}$.
Then, we define $\SWAP(n,c_1,c_2,p,q)$ to be the formula:
\begin{equation*}
T_n(c_1,p,q)\yeq T_n(c_2,p,q)
\end{equation*}
with $T_k(c,p,q)$ defined by
\begin{equation}\label{eq:swap}
  T_k(c,p,q)=\left\{
 \begin{array}{ll}
   a & \text{if $k=0$,}\\
   swap(T_{k-1}(c,p,q),i_{p(k)},i_{q(k)}) & \text{if $1\leq k\leq n$ and
   $k\in c$, and}\\
   swap(T_{k-1}(c,p,q),i_{q(k)},i_{p(k)}) & \text{if $1\leq k\leq n$ and
   $k\not\in c$.}
\end{array}
\right.
\end{equation}

\noindent
$T_n(c,p,q)$ is the array obtained by swapping the elements of
position $p(k)$ and $q(k)$ of the array $a$ for $1\leq k\leq n$.
The role of the subset $c$ is to determine whether the element at
position $p(k)$ has to be swapped with that at position $q(k)$ or vice
versa, and has the effect of shuffling the indices within the formula.

\begin{textofexample}\label{example:swap}
If $n=3$, $c_1=\{1\}$, $c_2=\{2,3\}$, $p$ and $q$ are
such that $p(k)=k$ and $q(k)=2$, for all $k$, $1\leq k\leq n$, then
$$T_n(c_1,p,q)=swap(swap(swap(a,i_1,i_2),i_2,i_2),i_2,i_3)$$
$$T_n(c_2,p,q)=swap(swap(swap(a,i_2,i_1),i_2,i_2),i_3,i_2).$$
\end{textofexample}

\noindent
Thus,
$\SWAP(n)=\{\SWAP(n,c_1,c_2,p,q) : c_1,c_2\subseteq\{1,\ldots,n\}\text{ and }
p,q:\{1,\ldots,n\}\rightarrow \{1,\ldots,n\} \}$.
Each formula, once negated, transformed into clausal form,
$\mathcal{A}$-reduced and flattened,
leads to a problem with $O(n)$ clauses.
\ignore{
Indeed,
flattening each side of $\SWAP(n,c_1,c_2,p,q)$ generates $4n$ equations of the form
$c_k^1\yeq select(c_{k-1}^4,j)$, $c_k^2\yeq store(c_{k-1}^4,i,c_k^1)$,
$c_k^3\yeq select(c_{k-1}^4,i)$, $c_k^4\yeq store(c_k^2,j,c_k^3)$,
where $c_k^1,c_k^2,c_k^3,c_k^4$ are new constants, for $1\leq k\leq n$
and $c_0^4 = a$.
}

A small change in the definition produces a class $\SWAPINVALID$.
With $c_1, c_2, p, q$ defined as for $\SWAP$, let
$\SWAPINVALID(n,c_1,c_2,p,q)$ be the formula:
\begin{equation*}
T_n(c_1,p,q)\yeq T_n(c_2,p,q^\prime)
\end{equation*}
where $T_k(c,p,q)$ is as in (\ref{eq:swap}),
$q^\prime:\{1,\ldots,n\}\rightarrow \{1,\ldots,n\}$ is such that
$q^\prime(1) = (q(1)+1) \mod n$,
and $q^\prime(k)=q(k)$ for all $k$, $2\le k\le n$.
Then, $\SWAPINVALID(n)=\{\SWAPINVALID(n,c_1,c_2,p,q) :
c_1,c_2\subseteq\{1,\ldots,n\},
p,q:\{1,\ldots,n\}\rightarrow \{1,\ldots,n\} \}$.

\begin{textofexample}\label{example:swapinvalid}
If $n$, $c_1$, $c_2$, $p$ and $q$ are as in Example~\ref{example:swap},
$$T_n(c_1,p,q)=swap(swap(swap(a,i_1,i_2),i_2,i_2),i_2,i_3)$$
$$T_n(c_2,p,q^\prime)=swap(swap(swap(a,i_3,i_1),i_2,i_2),i_3,i_2).$$
\end{textofexample}

\subsection{Third benchmark: $\STOREINV(n)$ and $\STOREINVINVALID(n)$}

The problems of the $\STOREINV$ family capture the following
property: if the arrays resulting from swapping elements of array $a$
with the elements of array $b$ occurring in the same positions are
equal, then $a$ and $b$ must have been equal to begin with.  For the
simple case where a single position is involved, we have:
\begin{displaymath}
  store(a,i,select(b,i))\yeq store(b,i,select(a,i)) \imp a\yeq b.
\end{displaymath}
For $n\geq 0$,
let $\STOREINV(n)=\{\multiswap(a,b,n)\imp a\yeq b\}$,
where
\begin{equation}
    \multiswap(a,b,k)=\left\{
  \begin{array}{l}
    (a\yeq b) \text{ if $k=0$,}\\
    \textbf{let}~(a^\prime \yeq b^\prime) = \multiswap(a,b,k-1)\textbf{~in}\\
    ~~~~~store(a^\prime,i_k,select(b^\prime,i_k))\yeq store(b^\prime,i_k,
         select(a^\prime,i_k))\\
    ~~~~~\text{if $k\geq 1$.}
  \end{array}\right.
\end{equation}

\begin{textofexample}\label{example:storeinv}
For $n=2$ we have
$$store(a^\prime,i_2,select(b^\prime,i_2))\yeq
store(b^\prime,i_2,select(a^\prime,i_2))\imp a\yeq b$$
where
$a^\prime=store(a,i_1,select(b,i_1))$ and
$b^\prime=store(b,i_1,select(a,i_1))$.
\end{textofexample}

\noindent
Transformation into clausal form of the negation of the formula in $\STOREINV(n)$,
followed by $\mathcal{A}$-reduction and flattening,
yields a problem with $O(n)$ clauses.
\ignore{
More precisely, $\mathcal{A}$-reduction replaces $a\nyeq b$ by
$select(a,i^*)\nyeq select(b,i^*)$,
for some new constant $i^*$ of sort $\textsc{index}$.
Flattening $\multiswap(a,b,n)$ produces $4n$ new equations,
and flattening $select(a,i^*)\nyeq select(b,i^*)$ yields $2$ equations,
$select(a,i^*)\yeq c_1$ and $select(b,i^*)\yeq c_2$,
and the inequation $c_1\nyeq c_2$, where $c_1, c_1\in\textsc{elem}$.
}

For $\STOREINVINVALID$, let
$store(t_a,i_n,select(t_b,i_n))\yeq store(t_b,i_n,select(t_a,i_n))$
be the formula returned by $\multiswap(a,b,n)$ for $n\geq 2$.
Then, we define
$$\STOREINVINVALID(n)=$$
$$\{store(t_a,i_1,select(t_b,i_n))\yeq
store(t_b,i_n, select(t_a,i_n)) \imp a\yeq b\}.$$

\subsection{Fourth benchmark: $\IOS(n)$}\label{sec:IOS}

The problems of the $\IOS(n)$ family
combine the theories of arrays and integer offsets.
Consider the following two program fragments:
\begin{verbatim}
           for(k=1;k<=n;k++)              for(k=1;k<=n;k++)
            a[i+k]=a[i]+k;                 a[i+n-k]=a[i+n]-k;
\end{verbatim}
If the execution of either fragment produces the same result
in the array \texttt{a}, then \texttt{a[i+n]==a[i]+n} must hold initially for
any value of \texttt{i}, \texttt{k}, \texttt{a}, and \texttt{n}.

\begin{textofexample}
For $n=2$, $\IOS(n)$ includes only the following valid formula:
\begin{align*}
store(store(a,i+1,select(a,i)+1),i+2,select(a,i)+2)\yeq\\
store(store(a,i+1,select(a,i+2)-1),i,select(a,i+2)-2)\\
 \imp\ select(a,i+2)\yeq select(a,i)+2.
\end{align*}
\end{textofexample}

In general, for $n\geq 0$ let 
$\IOS(n)=\{ L_n^n \yeq R_n^n \imp select(a,i+n) \yeq select(a,i)+n\}$
where
\begin{eqnarray*}
L_0^n = R_0^n = a & \\
L_k^n = store(L_{k-1}^n,i+k,select(a,i)+k) & \text{ for $k=1,\ldots,n$}\\
R_k^n = store(R_{k-1}^n,i+n-k,select(a,i+n)-k) & \text{ for $k=1,\ldots,n$.}
\end{eqnarray*}

\noindent
Each formula in $\IOS(n)$, once negated, reduced to clausal form,
flattened and $\mathcal{I}$-reduced, generates $O(n)$ clauses.
$\mathcal{A}$-reduction is not needed,
since the negation of the formula does not contain inequalities
of sort \textsc{array}.
\ignore{
By observing the actual instances one can extrapolate
that the number of equations for $\IOS(n)$ is $5n+4$.
}

\subsection{Fifth benchmark: $\QUEUE(n)$}\label{sec:queue}

The theories of arrays, records and integer offsets can be combined
to specify \emph{queues}, defined as usual in terms of the functions
$\enqueue : \textsc{elem}\times \textsc{queue} \rightarrow \textsc{queue}$,
$\dequeue : \textsc{queue}\rightarrow \textsc{queue}$,
$\first : \textsc{queue}\rightarrow \textsc{elem}$,
$\last : \textsc{queue}\rightarrow \textsc{elem}$ and
$\reset : \textsc{queue} \rightarrow \textsc{queue}$,
where $\textsc{queue}$ and $\textsc{elem}$ are the
sorts of queues and their elements, respectively.

Indeed, a queue can be implemented as a record with three fields:
\emph{items} is an array storing the elements of the queue,
\emph{head} is the index of the first element of the queue in the array,
and \emph{tail} is the index where the next element will be inserted in the queue.
Following Section~\ref{sec:records},
the signature features function symbols
$\rstore_{items}$, $\rstore_{head}$, $\rstore_{tail}$,
$\rselect_{items}$, $\rselect_{head}$ and $\rselect_{tail}$,
abbreviated as $\rstore_i$, $\rstore_h$, $\rstore_t$, $\rselect_i$, $\rselect_h$,
$\rselect_t$, respectively.
Then, the above mentioned functions on $\textsc{queue}$
are defined as follows:
\begin{displaymath}
  \begin{array}{rcl}
    \enqueue(v,x) &=&
    \rstore_t(\!\!\!\begin{array}[t]{l}
      \rstore_i(x,
      \store(\rselect_i(x),\rselect_t(x),v)),
      \s(\rselect_t(x)))
    \end{array}\\
    \dequeue(x) &=& \rstore_h(x,\s(\rselect_h(x)))\\
    \first(x) &=& \select(\rselect_i(x),\rselect_h(x))\\
    \last(x) &=& \select(\rselect_i(x),\p(\rselect_t(x)))\\
    \reset(x) &=& \rstore_h(x,\rselect_t(x))
  \end{array}
\end{displaymath}
where $x$ and $v$ are variables of sort $\textsc{queue}$
and $\textsc{elem}$, respectively, $\store$ and $\select$ are the
function symbols from the signature of $\mathcal{A}$,
$\p$ and $\s$ are the function symbols for predecessor and successor
from the signature of $\mathcal{I}$.

A basic property of queues is the following:
assume that $q_0$ is a properly initialized queue
and $q$ is obtained from $q_0$
by performing $n+1$ enqueue operations ($n>0$),
that insert $n+1$ elements $e_0, e_1, \ldots, e_n$,
and $m$ dequeue operations ($0\le m\le n$),
that remove $m$ elements $e_0, e_1, \ldots, e_{m-1}$;
then $\first(q)=e_m$.
Dequeue operations can be interleaved with enqueue operations in any order,
provided the number of dequeue operations is always strictly smaller
than the number of preceding enqueue operations.
For instance, if $q=\enqueue(e_2,\dequeue(\enqueue(e_1,\linebreak[0]
\enqueue(e_0,\reset(q_0)))))$, then $\first(q)=e_1$.
Problems in the $\QUEUE(n)$ family express an instance of
this property, where the dequeue operator is applied once
every $3$ applications of the enqueue operator.
Thus, the number of dequeue operations will be $m=\lfloor (n+1)/3 \rfloor$.

Given a term $t$ where the function symbols $\reset$, $\enqueue$,
$\dequeue$, $\first$, $\last$ and $\reset$ may occur,
let $t\unfold$ denote the term obtained from $t$
by unfolding the above function definitions.
Then $\QUEUE(n)$, for $n>0$, is defined as follows:
\begin{displaymath}
  \QUEUE(n)=\{
  (q_0 \yeq \reset(q)\unfold \wedge
  \bigwedge_{i=0}^{n-1} q_{i+1} \yeq f_{i+1}(e_i,q_i))
  \imp \first(q_n)\unfold\ \yeq e_m\},
\end{displaymath}
where
\begin{displaymath}
f_i(e,q) = \left\{
\begin{array}{ll}
\dequeue(\enqueue(e,q))\unfold & \text{if $~i\ mod\ 3 = 0$,}\\
\enqueue(e,q)\unfold & \text{otherwise,}
\end{array}
\right.
\end{displaymath}
and $m=\lfloor (n+1)/3 \rfloor$.

\begin{textofexample}
If $n=1$, then $\QUEUE(n)$ is the formula:
\begin{displaymath}
  \begin{array}{c}
  \left(
    \begin{array}{c}
    q_0 \yeq \rstore_h(q,\rselect_t(q))~~~~~~~~~~~\wedge\\
    q_1 \yeq \rstore_t(
      \begin{array}[t]{l}
      \rstore_i(q_0,
      \store(\rselect_i(q_0),\rselect_t(q_0),e_0)),
      \s(\rselect_t(q_0)))
      \end{array}
  \end{array}\right)
\imp\\
    \select(\rselect_i(q_1),\rselect_h(q_1)) \yeq e_0.
  \end{array}
\end{displaymath}
\end{textofexample}

\noindent
Each formula in $\QUEUE(n)$,
once negated, reduced to clausal form,
flattened and $\mathcal{I}$-reduced,
generates $O(n)$ clauses.
$\mathcal{A}$-reduction and $\mathcal{R}$-reduction are not needed,
since the negation of the formula does not contain inequalities
of sort \textsc{array} or \textsc{rec}.

\subsection{Sixth benchmark: $\CIRCULARQUEUE(n,k)$}\label{sec:cqueue}

It is sufficient to replace the theory of integer offsets $\mathcal{I}$
by $\mathcal{I}_k$,
to work with indices modulo $k$ and
extend the approach of the previous section to model
{\em circular queues of length $k$}.
The problems of the $\CIRCULARQUEUE(n,k)$ family say
that if $q_{n+1}$ is obtained from
a properly initialized circular queue $q_0$ by inserting $n+1$
elements $e_0, e_1, \ldots, e_n$, for $n>0$, and $n\!\!\mod k=0$,
then $\first(q_{n+1})\yeq\last(q_{n+1})$ holds,
because the last element inserted overwrites the one in the first position
(e.g., picture inserting $4$ elements in a circular queue of length $3$).
This is formally expressed by
$$\CIRCULARQUEUE(n,k)=$$
\begin{displaymath}
    \{
    (q_0\yeq\reset(q)\unfold \wedge
    \bigwedge_{i=0}^{n} q_{i+1}\yeq\enqueue(e_i,q_i)\unfold)
    \imp \first(q_{n+1})\unfold\yeq\last(q_{n+1})\unfold\},
\end{displaymath}
for $n>0$ such that $n\!\!\mod k=0$.

\begin{textofexample}
If $k=n=1$, then $\CIRCULARQUEUE(1,1)$ is the formula:
\begin{displaymath}
  \begin{array}{c}
  \left(
    \begin{array}{c}
    q_0 \yeq \rstore_h(q,\rselect_t(q))~\wedge\\
    q_1 \yeq \rstore_t(
      \rstore_i(q_0,
      \store(\rselect_i(q_0),\rselect_t(q_0),e_0)),
      \s(\rselect_t(q_0)))~\wedge\\
    q_2 \yeq \rstore_t(
      \rstore_i(q_1,
      \store(\rselect_i(q_1),\rselect_t(q_1),e_1)),
      \s(\rselect_t(q_1)))
  \end{array}\right)\\
\imp
    \select(\rselect_i(q_2),\rselect_h(q_2)) \yeq
    \select(\rselect_i(q_2),\p(\rselect_t(q_2))).
  \end{array}
\end{displaymath}
\end{textofexample}

\noindent
Each formula in $\CIRCULARQUEUE(n,k)$,
once negated, reduced to clausal form,
flattened and $\mathcal{I}$-reduced,
generates $O(n)$ clauses.

\section{Experiments}\label{experiments}

The synthetic benchmarks of Section~\ref{problems} were submitted
to three systems:
\eprover{}~0.82, CVC~1.0a and CVC~Lite~1.1.0.
The prover \eprover{} implements (a variant of)
$\SP$ with a large choice of fair search plans,
based on the ``given-clause'' algorithm \cite{JE,E081},
that ensure that the empty clause will be found,
if the input is unsatisfiable,
and a finite satisfiable saturated set will be generated,
if the input set is satisfiable and admits one.
CVC \cite{CVC} and CVC~Lite \cite{CVCLite}
combine several theory decision procedures following the Nelson-Oppen method,
including that of \cite{StuBaDiLe} for arrays with extensionality,
and integrate them with a SAT engine \cite{BDS02b}.
CVC is no longer supported; it was superseded by CVC~Lite,
a more modular and programmable system.
While CVC~Lite has many advantages,
at the time of these experiments
CVC was reported to be still faster on many problems.
CVC and CVC~Lite feature a choice of SAT solvers:
a built-in solver or Chaff~\cite{Chaff} for CVC,
a \emph{``fast''} or a \emph{``simple''} solver for CVC~Lite.
In our experiments, CVC and CVC~Lite performed
consistently better with their built-in and \emph{``fast''} solver,
respectively, and therefore only those results are reported.

We wrote a generator of pseudo-random instances of the synthetic benchmarks,
producing either TPTP\footnote{TPTP, or ``Thousands of
Problems for Theorem Provers'' is a {\em de facto} standard
for testing general-purpose first-order theorem provers:
see \url{http://www.tptp.org/}.} or CVC syntax,
and a set of scripts to run the solvers on all benchmarks.
The generator creates either $T$-reduced, flattened input files or plain
input files.
Flattening times were not included in the reported
run times, because flattening is a one-time linear time operation,
and the time spent on flattening was insignificant.
In the following, \emph{native input} means flattened,
$T$-reduced files for \eprover{},
and plain, unflattened files for CVC and CVC~Lite.

The experiments were performed on a 3.00GHz Pentium 4 PC with 512MB RAM.
Time and memory were limited to 150 sec and 256~MB per instance.
If a system ran out of either time or memory under these limits,
the result was recorded as a ``failure.''
When $Pb(n)$ is not a singleton
(cf. $\STORECOMM(n)$, $\STORECOMMINVALID(n)$,
$\SWAP(n)$ and $\SWAPINVALID(n)$),
the \emph{median} run time over all tested instances
is reported.\footnote{Reported figures refer to runs with 9 instances
for every value of $n$.
Different numbers of instances (e.g., 5, 20) were also tried,
but the impact on the plots was negligible.}
For the purpose of computing the median,
a failure is considered to be larger than all successful run times.
The median was chosen in place of the average,
precisely because it is well-defined even in cases
where a system fails on some, but not all instances of a given size,
a situation that occurred for all systems.

The results for \eprover{} refer to two variants of a simple strategy,
termed \emph{E(good-lpo)} and \emph{E(std-kbo)}, for reasons
that will be clear shortly.
This strategy adopts a \emph{single priority queue for clause selection},
where \emph{E(good-lpo)} gives the same priority to all clauses,
while \emph{E(std-kbo)} privileges ground clauses.
Additionally, \emph{E(good-lpo)} ensures that
all input clauses are selected before the generated ones,
whereas \emph{E(std-kbo)} does not.
Both variants employ a very simple \emph{clause evaluation heuristic}
to rank clauses of equal priority:
it weights clauses by counting symbols,
giving weight $2$ to function and predicate symbols and
weight $1$ to variable symbols.
Since these are the default term weights that \eprover{}
uses for a variety of operations,
they are pre-computed and cached, so that this heuristic
is very fast, compared to more complex schemes.
\ignore{
\begin{figure}[htb]
  \begin{tabular}{p{0.49\textwidth}p{0.49\textwidth}}
\emph{E(good-lpo)} & \emph{E(std-kbo)}\\[-2.0ex]
\begin{verbatim}
--expert-heuristic=
  '(1*Defaultweight(ConstPrio))'
--prefer-initial-clauses
-tLPO4
-Garrayopt
--memory-limit=256
\end{verbatim}&
\begin{verbatim}
--expert-heuristic=
  '(1*Defaultweight(PreferGround))'
-tKBO 
-Garity 
-warity 
--memory-limit=256
\end{verbatim}\\
  \end{tabular}
  \caption{{\small E command line options relevant to the search process.}}
\label{fig:e_options}
\end{figure}
}

\emph{E(std-kbo)} features a Knuth-Bendix ordering (KBO),
where the weight of symbols is given by their arity,
and the precedence sorts symbols by arity first,
and by inverse input frequency second (that is, rarer symbols are greater),
with ties broken by order of appearance in the input.
This KBO is ${\cal R}$-good, ${\cal I}$-good,
and it satisfies Condition (1), but not Condition (2),
in Definition~\ref{good:ordering:array},
so that it is not ${\cal A}$-good.
It was included because it is a typical ordering for first-order theorem proving,
and therefore \emph{E(std-kbo)} can be considered representative
of the behavior of a plain, standard, theorem-proving strategy.
\emph{E(good-lpo)} has a lexicographic (recursive) path ordering (LPO),
whose precedence extends that of \emph{E(std-kbo)},
in such a way that constants are ordered by sort.
Thus, also Condition (2) in Definition~\ref{good:ordering:array}
is satisfied and the resulting LPO is ${\cal R}$-good, ${\cal I}$-good and ${\cal A}$-good.
In both precedences, constants introduced by flattening are smaller
than those in the original signature.
It is worth emphasizing that contemporary provers, such as \eprover{},
can generate precedences and weighting schemes automatically.
The only human intervention was a minor modification
in the code to enable the prover
to recognize the sort of constants and satisfy
Condition (2) in Definition~\ref{good:ordering:array}.

\subsection{Experiments with $\STORECOMM$ and $\STORECOMMINVALID$}

Many problems involve \emph{distinct objects},
that is, constants which name elements that are known to be distinct
in all models of the theory.
\eprover{} features a complete variant of \SP{}
that builds knowledge of the existence of distinct objects
into the inference rules \cite{SBonObjects}.
Under this refinement,
the prover treats strings in double quotes and positive integers
as distinct objects.
This aspect is relevant to the $\STORECOMM$ problems,
because they include the inequalities in $\{i_l\nyeq i_m : (l,m)\in C^n_2\}$,
stating that all indices are distinct.
Thus, \eprover{} was applied to these problems in two ways:
with $\{i_l\nyeq i_m : (l,m)\in C^n_2\}$ included in the input
(\emph{axiomatized indices}),
and with array indices in double quotes
(\emph{built-in index type}).

\begin{figure}[htb] 
  \includegraphics[width=0.5\textwidth]{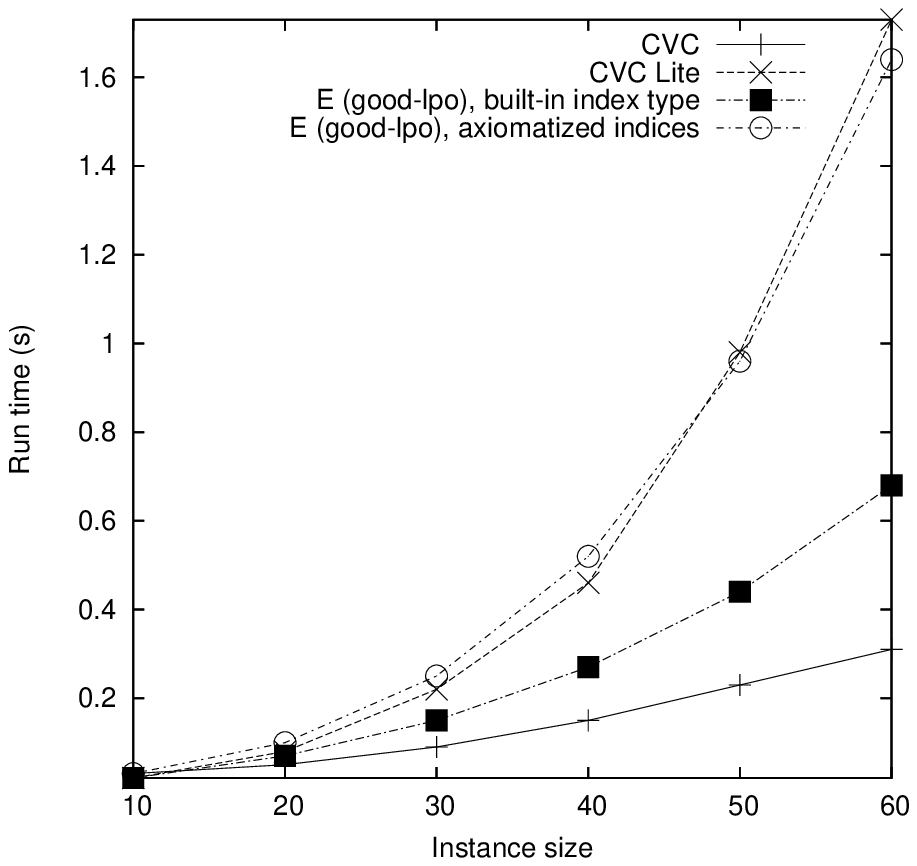}
  \includegraphics[width=0.5\textwidth]{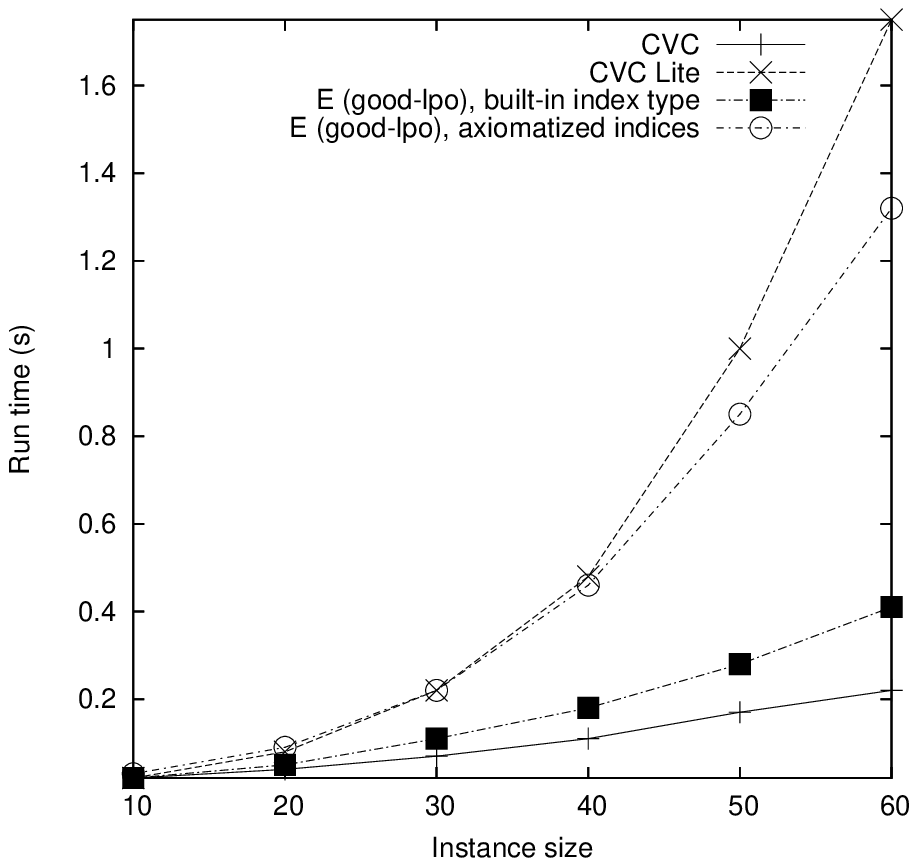}
  \caption{{\small Performance on valid (left) and invalid  (right) $\STORECOMM$
  instances with native input.}}
  \label{fig:storecomm_t1}
\end{figure}

Figure~\ref{fig:storecomm_t1} shows
that all systems solved the problems comfortably and scaled smoothly.
On valid instances, \emph{E(good-lpo)} with axiomatized indices
and CVC~Lite show nearly the same performance,
with \eprover{} apparently slightly ahead in the limit.
\emph{E(good-lpo)} with built-in indices outperformed CVC~Lite
by a factor of about 2.5.
CVC performed best improving by another factor of 2.
It is somewhat surprising that \eprover{}, a theorem prover
optimized for showing \emph{unsatisfiability},
performed comparatively even better on invalid (that is, satisfiable) instances,
where it was faster than CVC~Lite,
and \emph{E(good-lpo)} with built-in indices came closer to CVC.
The shared characteristics of all the plots strongly suggest that for
$\STORECOMM$ the most important feature is sheer processing speed.
Although there is no deep reasoning or search involved,
the general-purpose prover can hold its own against the
specialized solvers, and even edge out CVC~Lite.

\begin{figure}[htb] 
  \includegraphics[width=0.5\textwidth]{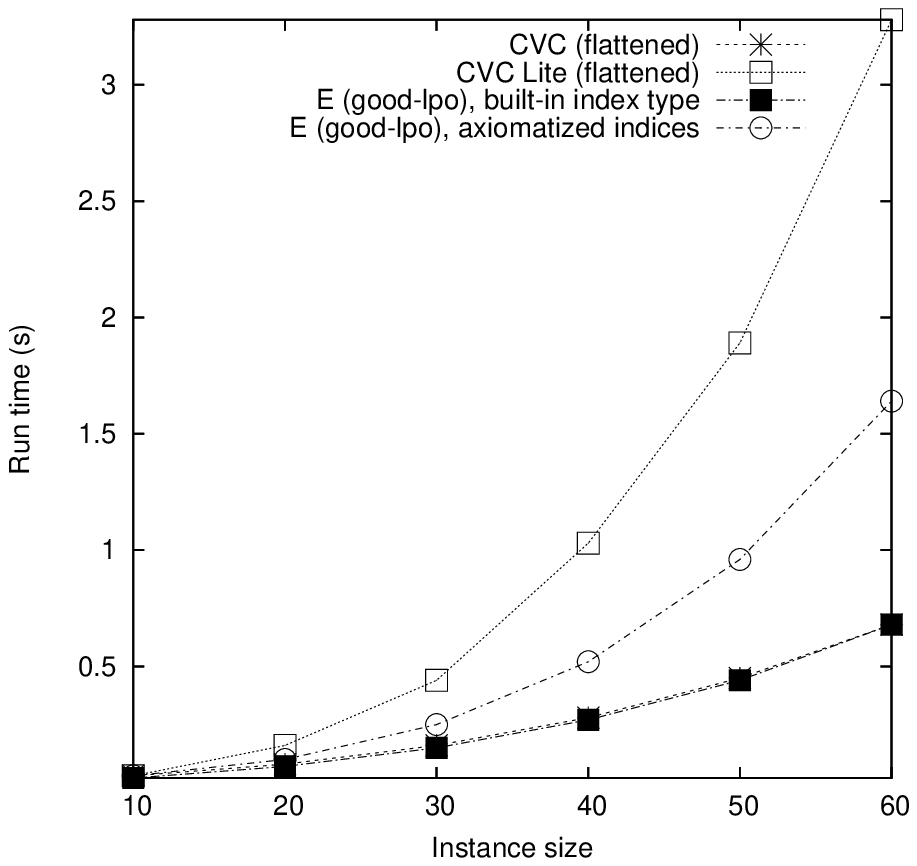}
  \includegraphics[width=0.5\textwidth]{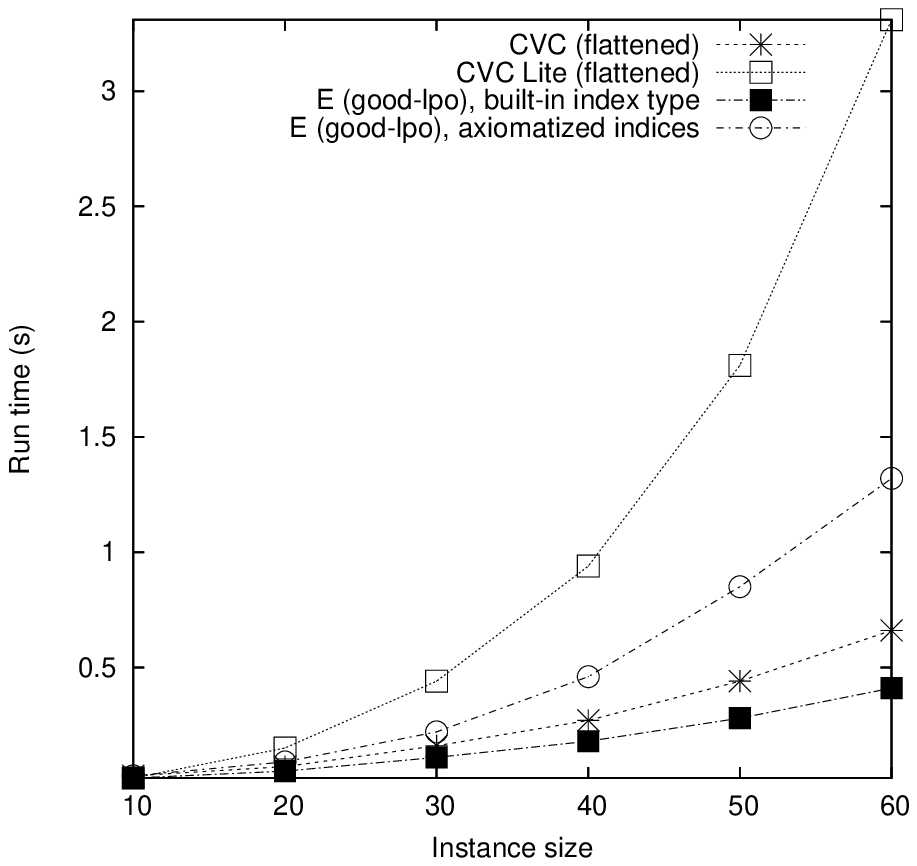}
  \caption{{\small Performance on valid (left) and invalid (right) $\STORECOMM$
           instances with flat input for all.}}
  \label{fig:storecomm_t1_flat}
\end{figure}

When \emph{all} systems ran on flattened input (Figure~\ref{fig:storecomm_t1_flat}),
both CVC and CVC~Lite exhibited run times approximately two times
higher than with native format, and CVC~Lite turned out to be the slowest system.
CVC and \eprover{} with built-in indices were the fastest:
on valid instances, their performances are so close,
that the plots coincide,
but \eprover{} is faster on invalid instances.
It is not universally true that flattening hurt CVC and CVC~Lite:
on the $\SWAP$ problems CVC~Lite performed much better on flattened input.
This suggests that specialized decision procedures are not insensitive to input format.

Although CVC was overall the fastest system on $\STORECOMM$, \eprover{}
was faster than CVC~Lite, and did better than CVC on invalid instances
when they were given the same input.
As CVC may be considered a paradigmatic representative
of optimized systems with built-in theories,
it is remarkable that the general-purpose prover
could match CVC and outperform CVC~Lite.

\subsection{Experiments with $\SWAP$ and $\SWAPINVALID$}\label{lemmaAdded}

\begin{figure}[htb]
  \includegraphics[width=0.5\textwidth]{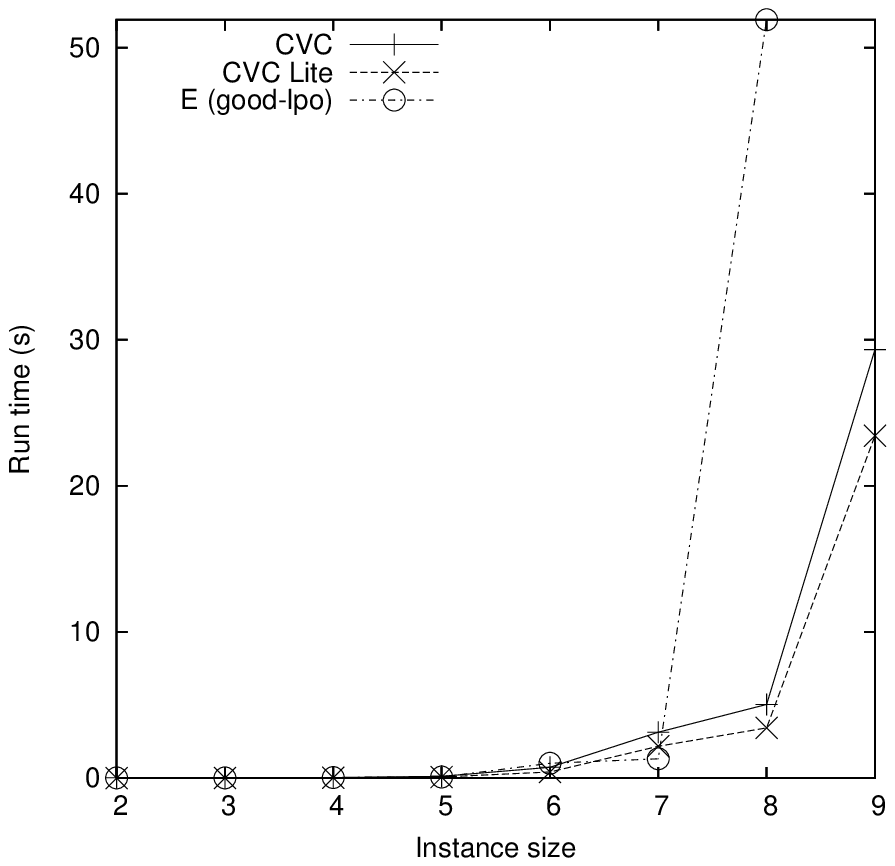}
  \includegraphics[width=0.5\textwidth]{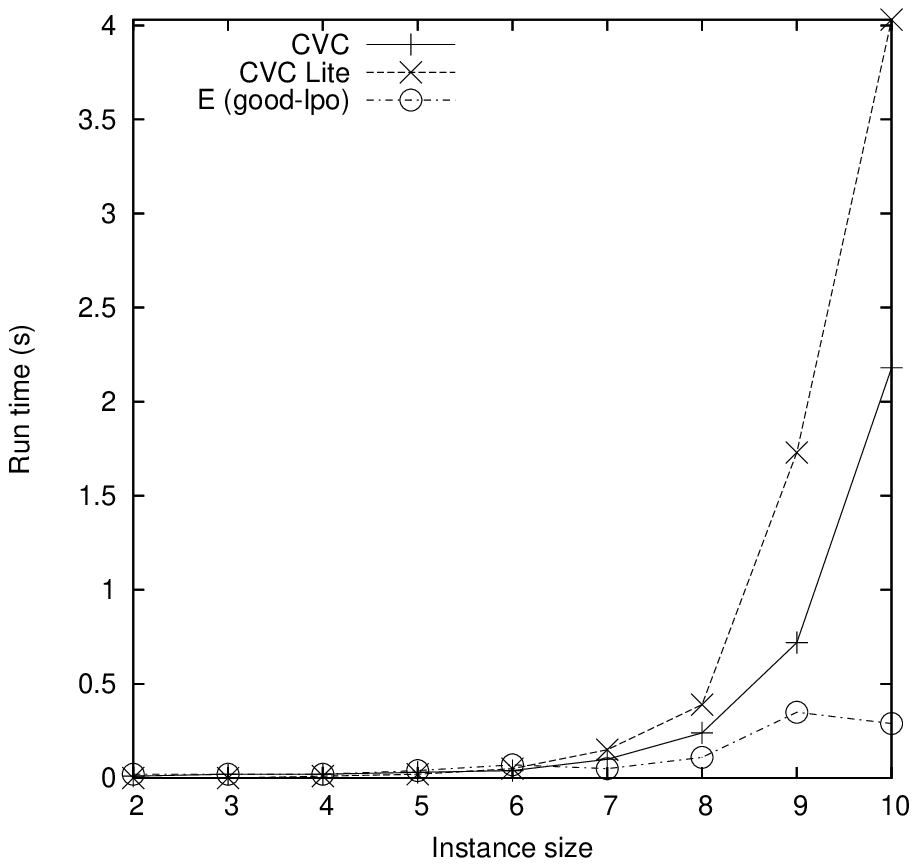}
  \caption{{\small Performance on valid (left) and invalid (right)
    $\SWAP$ instances, native input.}} 
  \label{fig:swap_t1}
\end{figure}

Rather mixed results arose for $\SWAP$, as shown in Figure.~\ref{fig:swap_t1}.
Up to instance size 5, the systems are very close.
Beyond this point, on valid instances, \eprover{} leads up to size 7,
but then is overtaken by CVC and CVC~Lite.
\eprover{} could solve instances of size 8,
but was much slower than CVC and CVC~Lite, which solved instances up to size 9.
No system could solve instances of size 10.
For invalid instances, E solved easily instances up to size 10 in less than 0.5 sec.
CVC and CVC~Lite were much slower there, taking 2 sec and 4 sec, respectively.
Their asymptotic behaviour seems to be clearly worse.

\begin{figure}[htb]
  \begin{center}
  \includegraphics[width=0.5\textwidth]{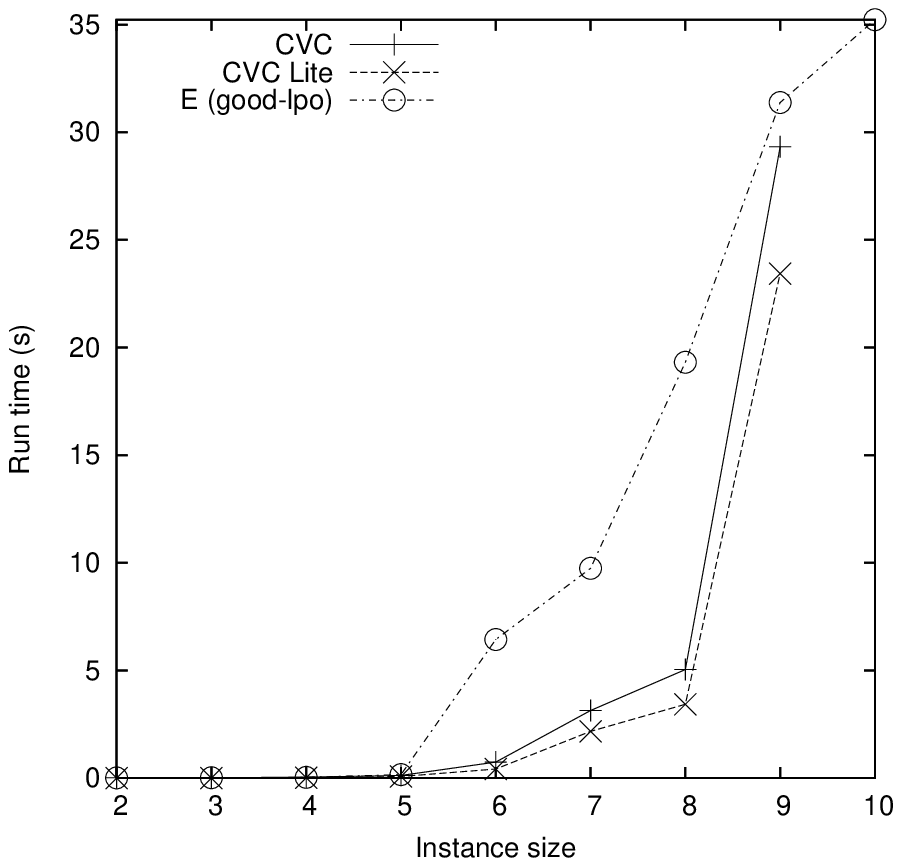}
  \end{center}
  \caption{{\small Performance on valid $\SWAP$ instances
         with added lemma for \eprover{}.}}
  \label{fig:swap_t3}
\end{figure}

Consider the lemma
$$store(store(x,z,select(x,w)),w,select(x,z)) \yeq$$
$$store(store(x,w,select(x,z)),z,select(x,w))$$
that expresses ``commutativity'' of $store$.
Figure~\ref{fig:swap_t3} displays the systems' performance on valid instances,
when the input for \eprover{} includes this lemma.
Although this addition means that the theorem prover is no longer
a decision procedure,\footnote{Lemma~\ref{lemma:caseAnalysis:array}
and therefore Theorem~\ref{theorem:array} do not hold,
if this lemma is added to presentation $\mathcal{A}$.}
\eprover{} terminated also on instances of size 9 and 10,
and its plot suggests a better asymptotic behavior.
While no system emerged as a clear winner,
this experiment shows how a prover that takes a theory presentation in input
offers an additional degree of freedom,
because useful lemmata may be added.

\subsection{Experiments with $\STOREINV$ and $\STOREINVINVALID$}

\begin{figure}[htb]
  \includegraphics[width=0.5\textwidth]{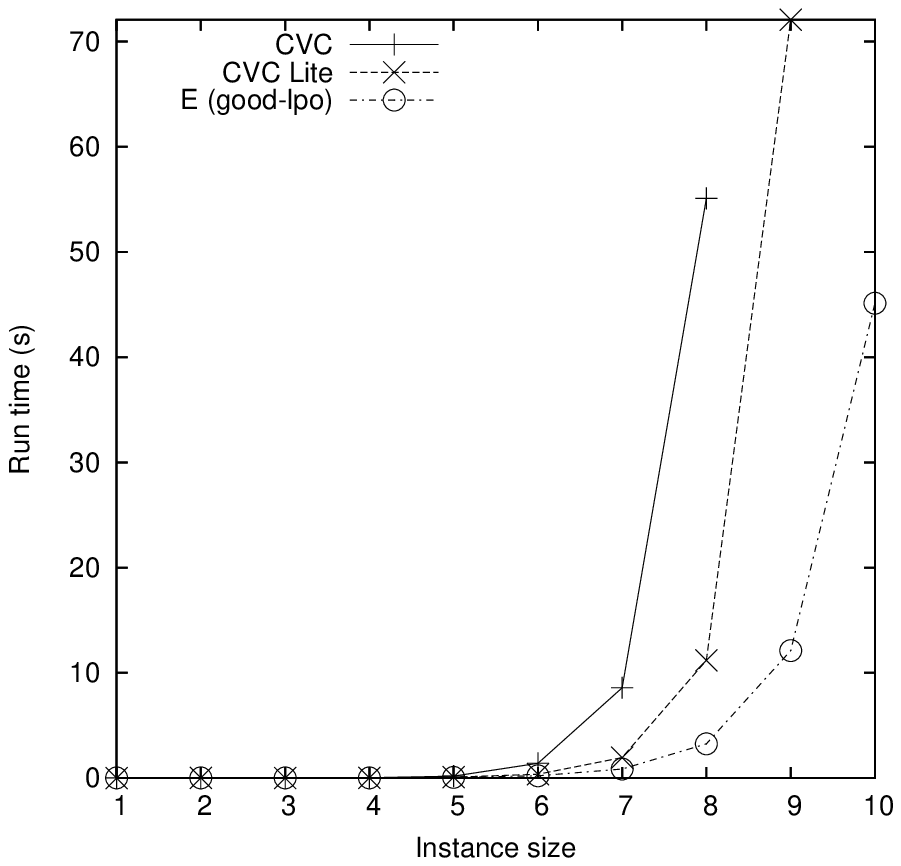}
  \includegraphics[width=0.5\textwidth]{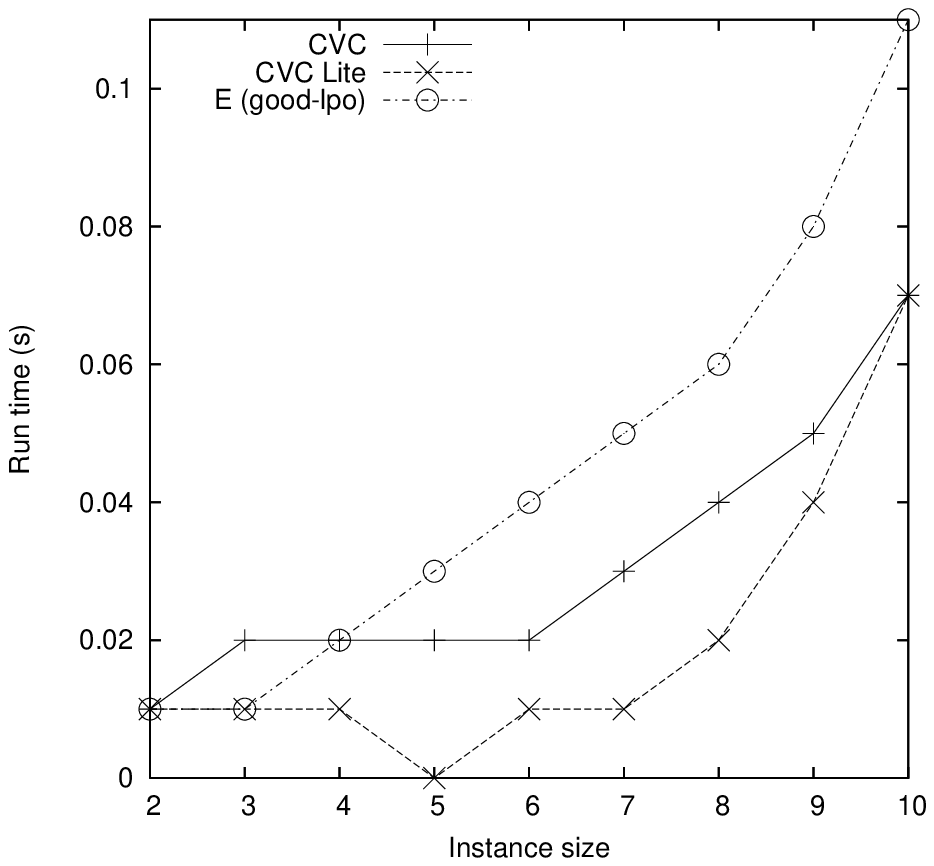}
  \caption{{\small Performance on valid (left) and invalid
  (right) $\STOREINV$ instances with native input.}}
  \label{fig:storeinv_t1}
\end{figure}

The comparison becomes even more favorable for the prover on the
$\STOREINV$ problems, reported in Figure~\ref{fig:storeinv_t1}.
CVC solved valid instances up to size 8 within the given resource limit.
CVC~Lite went up to size 9,
but \eprover{} solved instances of size 10, the largest generated.
A comparison of absolute run times at size 8,
the largest solved by all systems,
gives 3.4 sec for \eprover{}, 11 sec for CVC~Lite, and 70 sec for CVC.
Furthermore, \emph{E(std-kbo)} (not shown in the figure) solved valid
instances in \emph{nearly constant time}, taking less than 0.3 sec
for the hardest problem.
Altogether, \eprover{} with a suitable
ordering was clearly qualitatively superior than the dedicated systems.
For invalid instances, \eprover{} did not do as well,
but the run times there were minimal, with the largest
run time for instances of size 10 only about 0.1 sec.

\subsection{Experiments with $\IOS$}\label{sec:ios_benchmark}

\begin{figure}[htb]
  \includegraphics[width=0.5\textwidth]{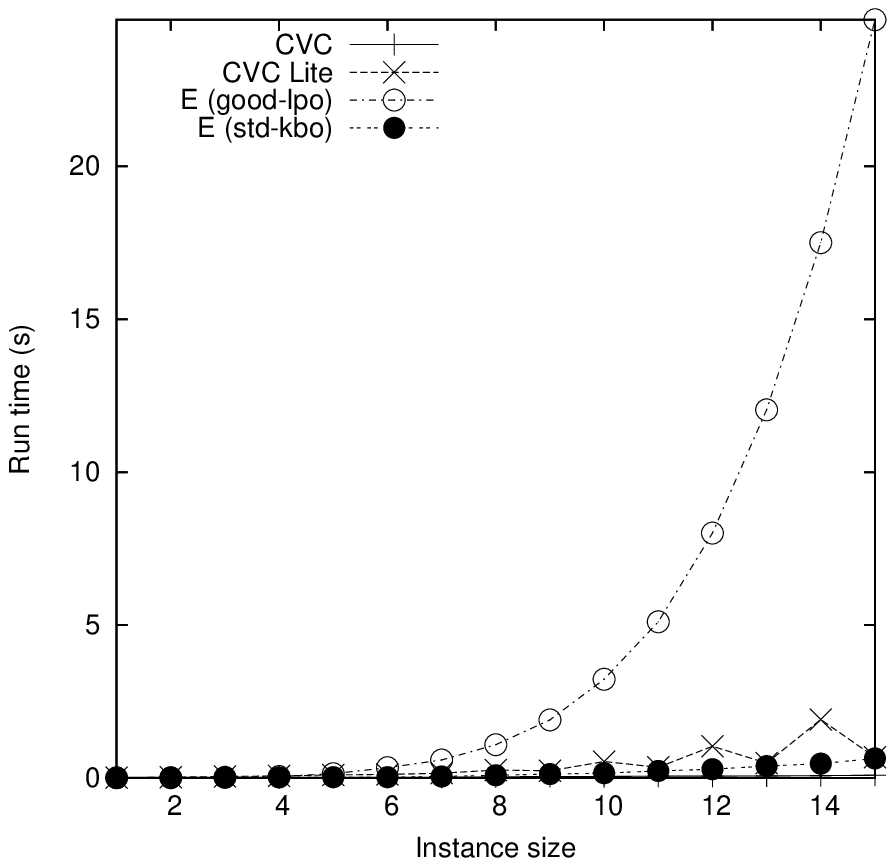}
  \includegraphics[width=0.5\textwidth]{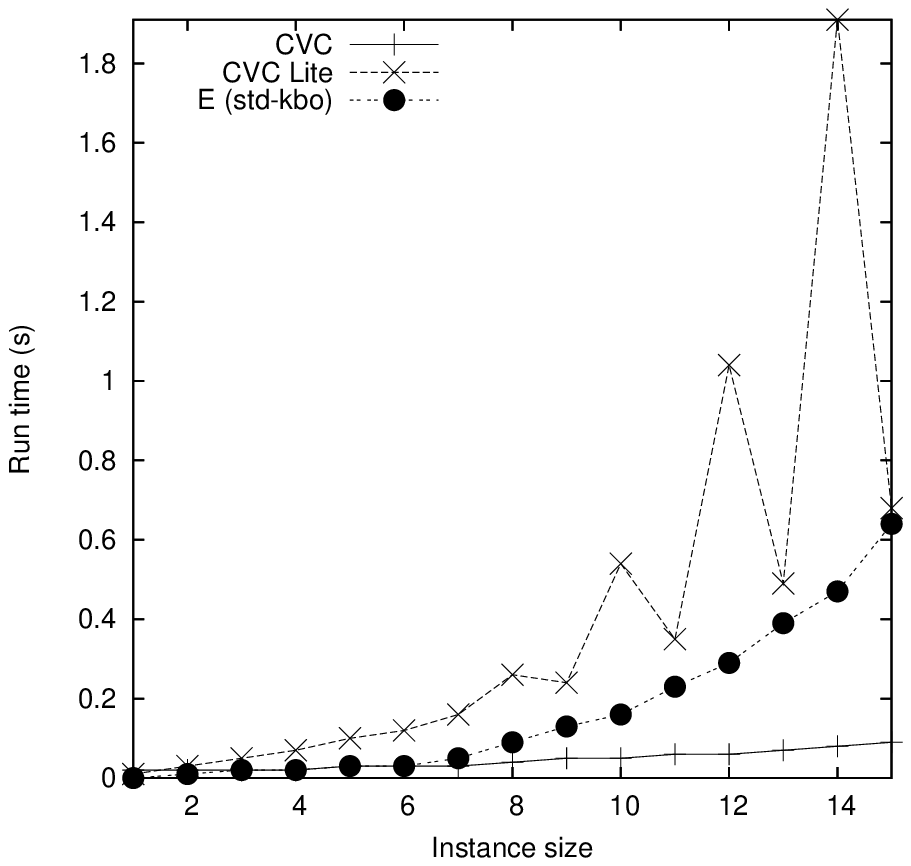}
  \caption{{\small Performance on the $\IOS$ instances: since in the graph
on the left the curve for CVC is barely visible, the graph on the right
shows a rescaled version of the same data, including only the three fastest systems.}}
  \label{fig:ios}
\end{figure}

The $\IOS$ problems were encoded for CVC and CVC~Lite
by using their built-in linear arithmetic,
on the reals for CVC and on the integers for CVC~Lite.
We tried to use inductive types in CVC,
but it performed badly and even reported incorrect results.\footnote{This is a known bug,
that will not be fixed since CVC is no longer supported \cite{ASToAA}.
CVC~Lite~1.1.0 does not support inductive types.}
In terms of performance (Figure~\ref{fig:ios}, left),
CVC was clearly the best system,
as expected from a tool with built-in arithmetic.
\emph{E(good-lpo)} was no match,
although it still solved all tried instances (Figure~\ref{fig:ios}, left).
On the other hand, \emph{E(std-kbo)} proved to be competitive
with the systems with built-in arithmetic.
At least two reasons explain why \emph{E(std-kbo)} behaved much better
than \emph{E(good-lpo)}:
first, KBO turned out to be more suitable than LPO for these benchmarks;
second, by not preferring initial clauses,
the search plan of \emph{E(std-kbo)} did not consider
the acyclicity and array axioms early in the search,
a choice that turned out to be good.
More remarkably,
\emph{E(std-kbo)} did better than CVC~Lite (Figure~\ref{fig:ios}, right):
its curve scales smoothly,
while CVC~Lite displays oscillating run times,
showing worse performance for even instance sizes than for odd ones.

\subsection{Experiments with $\QUEUE$ and $\CIRCULARQUEUE$}
\label{sec:exper-with-queue}

\begin{figure}[htb]
  \includegraphics[width=0.5\textwidth]{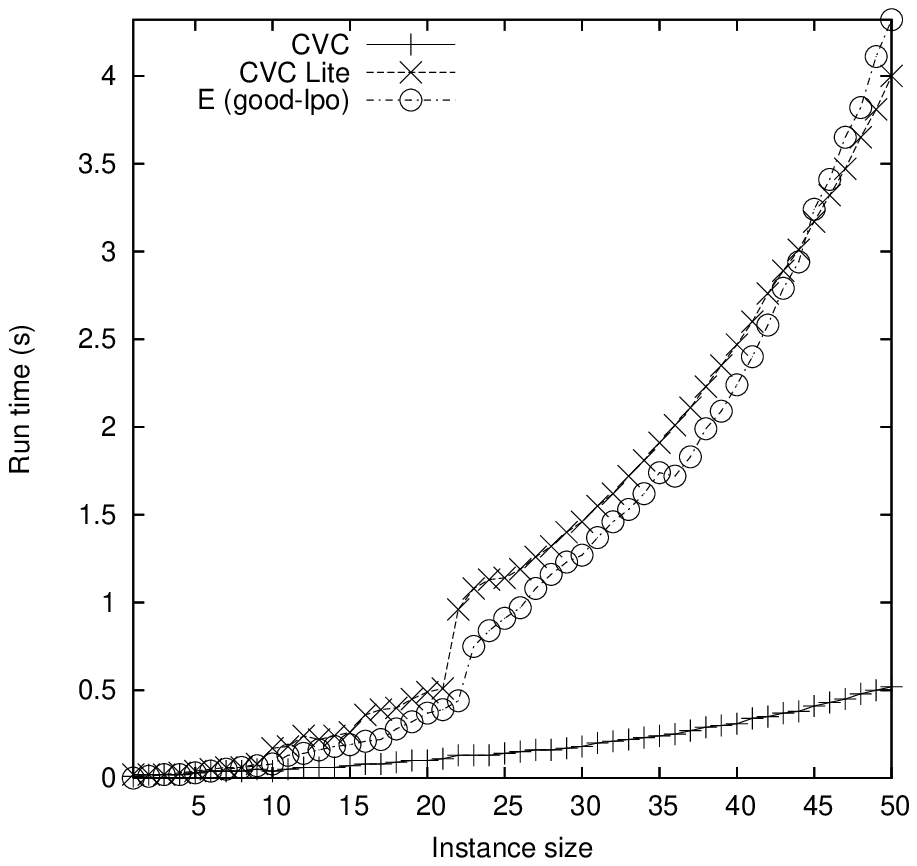}
  \includegraphics[width=0.5\textwidth]{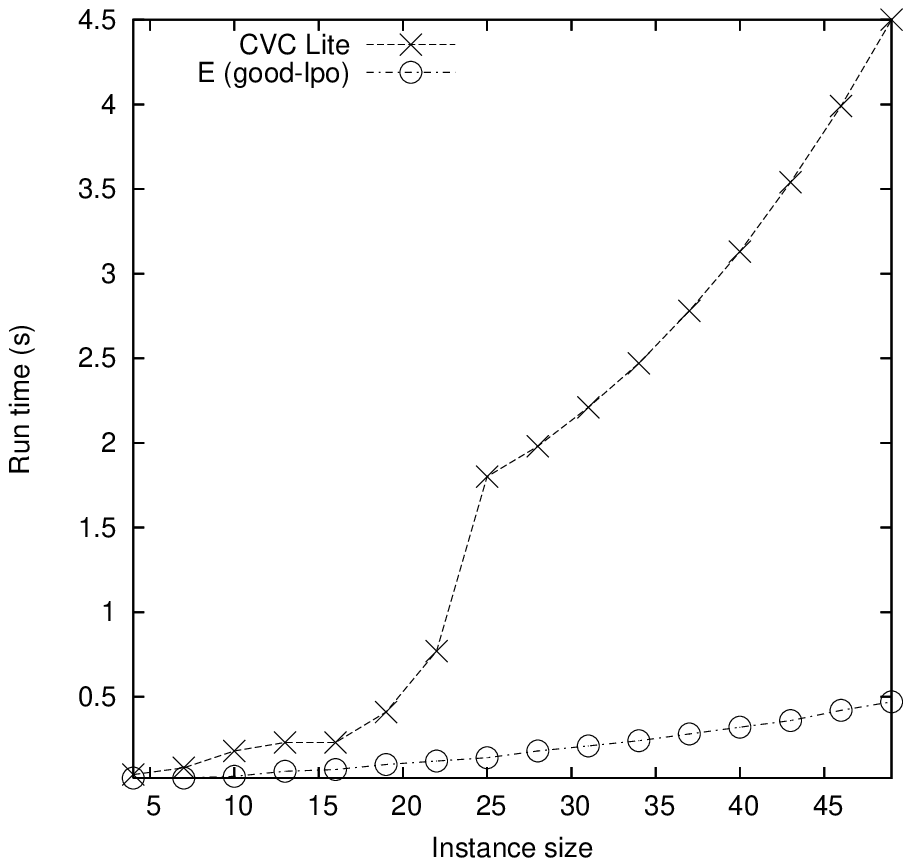}
  \caption{{\small Performance of E, CVC and CVC~Light on $\QUEUE$ (left) and
           $\CIRCULARQUEUE$ (right) with $k=3$.}}
  \label{fig:queues}
\end{figure}

Similar to the $\IOS$ tests,
CVC and CVC~Light were expected to enjoy a great
advantage over \eprover{} on the $\QUEUE$ and $\CIRCULARQUEUE$ problems,
because both CVC and CVC~Light build
all theories involved in these benchmarks,
namely arrays, records and linear arithmetic,
into their decision procedures.
The diagram on the left of Figure~\ref{fig:queues} confirms this expectation,
showing that CVC was the fastest system on the $\QUEUE(n)$ problems.
However, \emph{E(good-lpo)} was a good match for CVC~Light,
and \emph{E(std-kbo)} (not reported in the figure) also solved all tried instances.
The plots on the right of Figure~\ref{fig:queues} refers to the
experiments with $\CIRCULARQUEUE(n,k)$, where $k=3$.
It does not include CVC,
because CVC cannot handle the \emph{modulo-k} integer
arithmetic required for circular queues.
Between CVC~Lite and \eprover{},
the latter demonstrated a clear superiority:
\emph{E(good-lpo)} exhibited nearly linear performance,
and proved the largest instance in less than 0.5 sec,
nine times faster than CVC~Lite.
\emph{E(std-kbo)} behaved similarly.

\subsection{Experiments with ``real-world'' problems}

While synthetic benchmarks test scalability,
``real-world'' problems such as those from the
UCLID suite \cite{SS:CAV-2004} test performance
on huge sets of literals.
UCLID is a system that reduces all problems to propositional form
without theory reasoning.
Thus, in order to get problems relevant to our study,
we used \harvey{} \cite{haRVey-sefm03} to extract ${\cal T}$-satisfiability
problems from various UCLID inputs.
This resulted in 55,108 proof tasks in the
{\em combination of the theory of integer offsets and
the quantifier-free theory of equality},
so that ${\cal I}$-reduction was applied next.
We ran \eprover{} on all of them,
using a cluster of 3 PC's with 2.4GHz Pentium-4 processors.
All other parameters were the same as for the synthetic benchmarks.

\begin{figure}[htb]
  \includegraphics[width=0.5\textwidth]{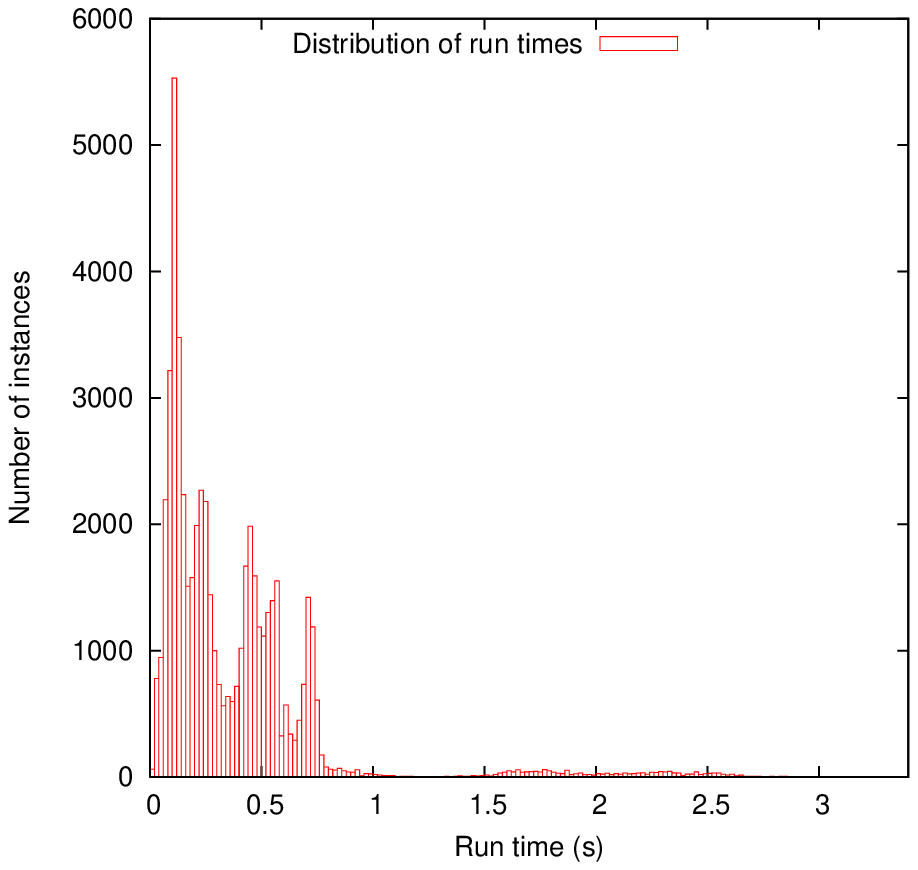}
  \includegraphics[width=0.5\textwidth]{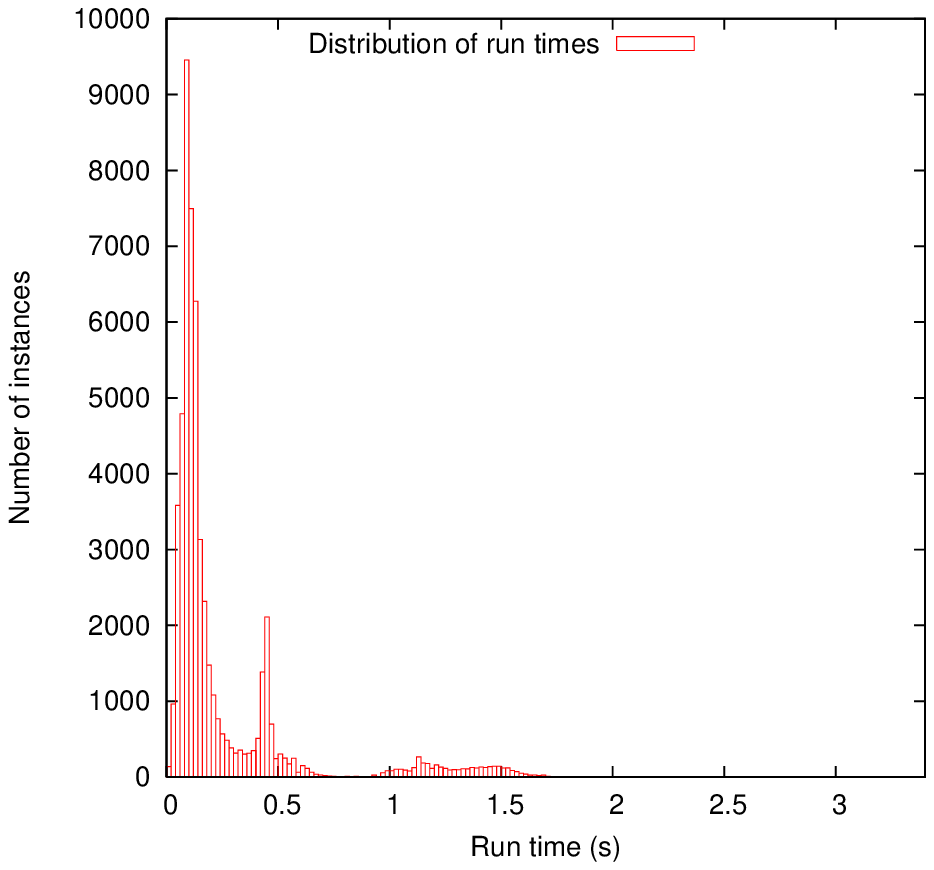}
  \caption{{\small Distribution of run times for \eprover{} in automatic mode (left)
    and using an optimized strategy (right) for the UCLID test set.}}
  \label{fig:e_on_uclid}
\end{figure}

These problems (all valid) turned out to be easy for \eprover{}
in automatic mode, where the prover chooses automatically ordering and search plan.
It could solve \emph{all} problems, taking less than 4 sec on the hardest one,
with average 0.372 sec and median 0.25 sec.
Figure~\ref{fig:e_on_uclid} shows a histogram of run times:
the vast majority of problems was solved in less than 1 sec and
very few needed between 1.5 and 3 sec.
An optimized search plan was found by testing on a random sample
of 500 problems, or less than 1\% of the full set.
With this search plan, very similar to \emph{E(std-kbo)},
the performance improved by about 40\% (Figure~\ref{fig:e_on_uclid}, right):
the average is 0.249 sec, the median 0.12 sec,
the longest time 2.77 sec,
and most problems were solved in less than 0.5 sec.

\section{Discussion}\label{disc}

The application of automated reasoning to verification
has long shown the importance of decision procedures
for satisfiability in decidable theories.
The most common approach to these procedures,
popularized as the {\em ``little'' proof engines} paradigm \cite{Shankar},
works by building each theory ${\cal T}$ into a dedicated inference engine.
For ${\cal T}$-satisfiability procedures,
that decide conjunctions of ground ${\cal T}$-li\-te\-rals,
the mainstay is the congruence closure algorithm,
enhanced by building ${\cal T}$ into the algorithm
(e.g., \cite{NO80,StuBaDiLe,LaMu,Musuvathi,DM06,NieuOli06}).
Such procedures are combined according to the scheme of \cite{NO79}
or its variants
(e.g., \cite{TinelliHarandi,BDS02a,Ga02,Ghilardi04,BaaGhi,ZaRaRi}).
Recent systematic treatments appeared in
\cite{KrsCon:CADE-2002,MZ03,ConKrs:TACAS2003,RaRiTran:ICTAC04,GanRuSha,Ghilardi05}.

For ${\cal T}$-decision procedures,
that decide arbitrary quantifier-free ${\cal T}$-formul\ae\@,
the so-called ``eager'' approaches seek efficient reductions of the problems to SAT
and submit them to SAT solvers
(e.g., \cite{issta00,velev01,uclid,SSB:DAC-2003,OMOS}).
The so-called ``lazy'' approaches (e.g.,
\cite{ICSrelated,BDS02b,FJOS03,haRVey-sefm03,CVCLite,dpllt,simplify,mathsat1,mathsat2,NieuOli06})
integrate ${\cal T}$-satisfiability procedures
based on congruence closure with SAT solvers,
usually based on the Davis-Putnam-Logemann-Loveland procedure (e.g., \cite{ChangLee,Chaff}).
The resulting systems are called SMT solvers.

By symmetry with {\em little proof engines},
we used {\em ``big'' proof engines} (e.g., \cite{Stickel}),
for theorem-proving strategies for full first-order logic with equality,
as implemented in state-of-the-art general-purpose theorem provers
(e.g., \cite{otter3,WABCEKTT,RV:AICOM-2002,JE}).
There has always been a continuum
between big and little engines of proof,
as testified by the research on {\em reasoning modulo a theory} in big engines.
The {\em rewriting approach to ${\cal T}$-satisfiability} aims
at a cross-fertilization where big engines work \emph{as} little engines.
The general idea is to explore how the technology of big engines
(orderings, inference rules, search plans, algorithms, data structures,
implementation techniques) may be applied selectively and efficiently
``in the small,'' that is, to decide specific theories.

This exploration finds its historical roots in the relationship
between congruence closure and Knuth-Bendix completion \cite{KnuthBendix70}
in the ground case:
the application of ground completion
to compute congruence closure was discovered as early as \cite{L75:ut};
the usage of congruence closure to generate canonical rewrite systems
from sets of ground equations was investigated further in
\cite{JFAA,PlaistedASK};
more recently, a comparison of ground completion and congruence
closure algorithms was given in \cite{BacTiwVig};
and ground completion and congruence closure were included in an abstract
framework for canonical inference in \cite{TCLcanonicity}.
The central component of the rewriting approach is the inference system $\SP$,
that is an offspring of a long series of studies
on completion for first-order logic with equality.
These systems were called by various authors
{\em rewrite-based}, {\em completion-based}, {\em superposition-based},
{\em paramodulation-based}, {\em contraction-based}, {\em saturation-based}
or {\em ordering-based}, to emphasize one aspect or the other.
Relevant surveys include \cite{Plaistedeqsurvey,taxonomy,NRHB,DPHB}.

The gist of the rewriting approach to ${\cal T}$-satisfiability
is to show that a sound and refutationally complete ``big'' engine,
such as $\SP$, is guaranteed to generate finitely many clauses
from ${\cal T}$-satisfiability problems.
By adding termination to soundness and completeness,
one gets a decision procedure: an $\SP$-strategy that combines
$\SP$ with a fair search plan is a ${\cal T}$-satisfiability procedure.
Depending on the theory, termination may require some problem transformation,
termed {\em ${\cal T}$-reduction},
which is fully mechanizable for all the theories we have studied.
We emphasize that the inference system is {\em not} adapted to the theories.
The only requirement is on the ordering:
$\SP$ is parametric with respect to a CSO,
and the termination proof for ${\cal T}$ may require
that this ordering satisfies some property named {\em ${\cal T}$-goodness}.
For all considered theories,
${\cal T}$-goodness is very simple and easily satisfied by common
orderings such as RPO's and KBO's.

We proved termination of $\SP$ on several new theories,
including one with infinite axiomatization\footnote{Our termination
result for the theory of integer offsets was generalized in \cite{rds}
to the theories of {\em recursive data structures} as defined in \cite{Oppen}.}.
We gave a {\em general modularity theorem} for the {\em combination of theories},
and carried out an experimental evaluation,
to test the pratical feasibility of the rewriting approach.
The modularity theorem states sufficient conditions
(no shared function symbols, variable-inactive theories)
for $\SP$ to terminate on a combination of theories
if it terminates on each theory separately.
The ``no shared function symbols'' hypothesis is common for combination results.
Variable-inactivity is satisfied by all equational theories with no trivial models.
First-order theories that fail to be variable-inactive
in an intuitive way are not stably infinite,
and therefore cannot be combined by the Nelson-Oppen scheme either.
On the other hand, it follows from work in \cite{IJCAR2006csp}
that variable-inactive theories are stably infinite.
The quantifier-free theories of \emph{equality}, \emph{lists},
\emph{arrays} with or without extensionality,
\emph{records} with or without extensionality,
\emph{integer offsets} and \emph{integer offsets modulo},
all satisfy these requirements,
so that any fair $\SP$-strategy is a satisfiability procedure
for any of their combinations.
The theories of arrays and possibly empty lists are {\em not convex},
and therefore cannot be combined by the Nelson-Oppen scheme without case analysis.

A different approach to big engines in the context of
theory reasoning, and especially combination of theories,
appeared in \cite{GanSofWal}: the combination or extension of theories
is conceived as mixing total and partial functions,
and a new inference system with
{\em partial superposition} is introduced to handle them.
Because the underlying notion of validity is modified
to accomodate partial functions,
the emphasis of \cite{GanSofWal} is on defining the new inference system
and proving its completeness.
The realization of such an approach requires to implement the new inference system.
The essence of our methodology, on the other hand, is to leave the
inference system and its completeness proof unchanged,
and prove termination to get decision procedures.
This allows us to take existing theorem provers ``off the shelf.''

For the experimental evaluation,
we designed six sets of synthetic benchmarks on the theory
of arrays with extensionality and combinations
of the theories of arrays, records and integer offsets or integer offsets modulo.
For ``real-world problems,''
we considered satisfiability benchmarks extracted from the UCLID suite.
Our experimental comparison between the $\SP$-based \eprover{} prover
and the validity checkers CVC and CVC~Lite is a first of its kind,
and offers many elements for reflection and suggestions for future research.

The analysis of the traces of the theorem prover
showed that these satisfiability problems behave very differently
compared to more typical theorem-proving problems.
Classical proof tasks involve a fairly large set of axioms,
a rich signature, many universally quantified variables,
many unit clauses usable as rewrite rules and
many mixed positive/negative literal clauses.
The search space is typically infinite,
and only a very small part of it gets explored.
In ${\cal T}$-satisfiability problems,
input presentations are usually very small
and there is a large number of ground rewrite rules generated by flattening.
The search space is finite,
but nearly all of it has to be explored,
before unsatisfiability (validity) can be shown.
Table~\ref{tab:search_features} compares the behavior of \eprover{}
on some medium-difficulty unsatisfiable array problems and some
representative TPTP problems of similar difficulty for the prover.

\begin{table}[tbh]
  \centering
  \begin{tabular}{lrrrrr}
    Problem & Initial & Generated &  Processed  & Remaining & Unnecessary \\
    Name    & clauses & clauses   &  clauses    & clauses   & inferences\\
    \hline
    \hline
    $\STORECOMM(60)/1$ & 1896 & 2840  & 4323  & 7  & 26.4\% \\ 
    $\STOREINV(5)$     & 27   & 22590 & 7480  & 31 & 95.5\% \\ 
    $\SWAP(8)/3$       & 62   & 73069 & 21743 & 56 & 98.2\% \\ 
    \hline
    \texttt{SET015-4}  & 15   & 39847 & 7504  & 16219 & 99.90\% \\ 
    \texttt{FLD032-1}  & 31   & 44192 & 3964  & 31642 & 99.96\% \\ 
    \texttt{RNG004-1}  & 20   & 50551 & 4098  & 26451 & 99.90\% \\ 
    \hline
  \end{tabular}

  \medskip
  {\small The data are for \eprover{} in automatic mode.
    $\STORECOMM(60)/1$ is one of the problems in $\STORECOMM(n)$ for $n=60$,
    $\STOREINV(5)$ is $\STOREINV(n)$ for $n=5$ and
    $\SWAP(8)/3$ is one of the problems in $\SWAP(n)$ for $n=8$.
    The others are representative problems from TPTP~3.0.0.
    The sum of processed
    and remaining clauses may be smaller than the sum of initial and
    generated clauses, because \eprover{} removes newly
    generated trivial clauses, as well as unprocessed clauses whose
    parents become redundant, without counting them as processed.
    The final column shows the percentage of \emph{all} inferences
    (expansion and contraction) that did \emph{not} contribute to the
    final proof.}
  \caption{{\small Performance characteristics of array and TPTP problems.}}
  \label{tab:search_features}
\end{table}

Most search plans and features of first-order provers
were designed assuming the search space characteristics
of typical first-order problems.
Thus, the theorem prover turned out to be competitive
with the little-engines systems,
although it was optimized for different search problems.
This means that
not only is using a theorem prover already a viable option in practice,
but there is a clear potential to improve both performance and usability,
by studying implementation techniques
of first-order inferences that target ${\cal T}$-satisfiability,
by designing {\em theory-specific search plans},
and by equipping the prover with the ability to recognize which
theories appear in the input set.
The prover also terminated in
many cases beyond the known termination results
(cf. Figure~\ref{fig:swap_t3} and the runs with \emph{E(std-kbo)},
whose ordering is not ${\cal A}$-good).
Thus, theorem provers are not as brittle as one may fear
with respect to termination, and still offer the flexibility
of adding useful lemmata to the presentation,
as shown in Section~\ref{lemmaAdded}.

The above remarks suggest that stronger termination results may be sought.
The complexity of the rewrite-based procedures may be improved
by adopting {\em theory-specific search plans}.
Methods to extract models from saturated sets
can be investigated to complement proof finding
with {\em model generation},
which is important for applications.
For instance, in verification,
a model represents a counter-example to a conjecture of correctness of a system.
The ability to generate models marks the difference between being able
to tell that there are errors (by reporting ``satisfiable''),
and being able to give some information on the errors
(by reporting ``satisfiable'' and a model).
Since we do not expect the rewrite-based approach
to work for \emph{full linear arithmetic},
another quest is how to integrate it with methods for arithmetic \cite{RueSha}
or other theories such as \emph{bit-vectors} \cite{CyrMoeRue}.
Research on integration with the latter theory began in \cite{HKRaRiTran:ICTAC05}.

Most verification problems involve
arbitrary quantifier-free ${\cal T}$-formul\ae,
or, equivalently, sets of ground ${\cal T}$-clauses.
Thus, a major open issue is how to apply big engines
towards solving general ${\cal T}$-decision problems.
Since $\SP$ is an inference system for general first-order clauses,
a set of ground ${\cal T}$-clauses may be submitted to an $\SP$-based prover.
Showing that an $\SP$-strategy is a ${\cal T}$-decision procedure
requires extending the termination results
from sets of ground literals to sets of ground clauses.
Sufficient conditions for termination of $\SP$ on ${\cal T}$-decision problems
were given recently in \cite{gendp}.
However, verification problems of practical interest typically yield
large sets with huge non-unit clauses,
and first-order provers are not designed to deal with large disjunctions
as efficiently as SAT solvers.

In practice,
a more plausible approach could be to follow
a simple ``lazy'' scheme (e.g., \cite{BDS02b,FJOS03}),
and integrate a rewrite-based ${\cal T}$-satisfiability procedure
with a SAT solver that generates assignments.
The rewrite-based ${\cal T}$-solver produces a proof,
whenever it detects unsatisfiability,
and can be made incremental to interact with the SAT solver
according to the ``lazy'' scheme.
However, the state of the art in SMT solvers indicates that
a tight integration of the two solvers is required to achieve high performances.
SAT solvers are based on case analysis by backtracking,
whereas rewrite-based inference engines are \emph{proof-confluent},
which means they need no backtracking.
While proof confluence is an advantage in first-order theorem proving,
this dissimilarity means that a tight integration of SAT solver and
rewrite-based ${\cal T}$-solver requires to address the issues
posed by the interplay of two very different kinds of control.

In current work we are taking a diffrent route:
we are exploring ways to decompose ${\cal T}$-decision problems,
in such a way that the big engine acts as a pre-processor for an SMT solver,
doing as much theory reasoning as possible in the pre-processing phase.
In this way we hope to combine the strength of a prover, such as
\eprover{}, in equational reasoning with that of an SMT solver in case analysis.
Even more general problems require to reason with universally quantified variables,
that SMT solvers handle only by heuristics,
following the historical lead of \cite{simplify}.
In summary, big engines are strong at reasoning with equalities,
universally quantified variables and Horn clauses.
Little engines are strong at reasoning with propositional logic,
non-Horn clauses and arithmetic.
The reasoning environments of the future will have to harmonize their forces.

\begin{acks}
We are indebted to Mnacho Echenim for several suggestions
to improve preliminary versions of this paper,
and especially for helping us correcting the proofs of Section~\ref{sec:ios}.
\end{acks}

\bibliographystyle{acmtrans}
\bibliography{references}

\begin{received}
Received April 2006\\
Revised February 2007\\
Accepted June 2007
\end{received}

\end{document}